\documentclass[preprint,3p,times]{elsarticle}
\RequirePackage{multirow,booktabs,subfigure,color,array,hhline,makecell}
\usepackage{amssymb}
\usepackage{amsmath}
\usepackage{graphicx}
\usepackage{subfigure}
\usepackage{amsthm}
\usepackage{mathrsfs}
\usepackage{indentfirst}
\usepackage[colorlinks,citecolor=blue,urlcolor=blue]{hyperref}
\allowdisplaybreaks
\usepackage[table]{xcolor}
\usepackage{pgf}
\usepackage{tikz}
\usetikzlibrary{arrows, decorations.pathmorphing, backgrounds, positioning, fit, petri, automata}
\usetikzlibrary{shadows,arrows,positioning}
\RequirePackage{algorithm,algpseudocode}
\allowdisplaybreaks
\usepackage{pdfpages}
\usepackage{changes}
\usepackage{caption}
\usepackage{setspace}
\newtheorem{thm}{Theorem}
\newtheorem{defin}{Definition}
\newtheorem{lem}{Lemma}
\newtheorem{assum}{Assumption}
\newtheorem{rem}{Remark}
\newtheorem{cor}{Corollary}

\newtheorem{prop}{Proposition}

\allowdisplaybreaks[4]
\AtBeginDocument{%
	\providecommand\BibTeX{{%
			\normalfont B\kern-0.5em{\scshape i\kern-0.25em b}\kern-0.8em\TeX}}}

\journal{Information Sciences}
\begin{document}
\captionsetup[figure]{labelfont={bf},labelformat={default},labelsep=period,name={Fig.}}

\begin{frontmatter}
\title{Directed mixed membership  stochastic blockmodel}

\author[label1]{Huan Qing}
\ead{qinghuan@u.nus.edu}
\author[label2]{Jingli Wang\corref{cor1}}
\ead{jlwang@nankai.edu.cn}
\cortext[cor1]{Corresponding author.}
\address[label1]{School of Economics and Finance, Chongqing University of Technology, Chongqing, 400054, China}
\address[label2]{School of Statistics and Data Science, KLMDASR, LEBPS, and LPMC, Nankai University, Tianjin, 300071, Tianjin, China}
\begin{abstract}
Mixed membership modeling for undirected networks has been extensively explored in network science over the past few years. Despite the substantial progress made for undirected cases, handling mixed membership structures in directed networks continues to pose substantial difficulties. To address this gap, we introduce the Directed Mixed Membership Stochastic Blockmodel (DiMMSB), a novel framework tailored for directed networks with overlapping communities. A key feature of DiMMSB is its ability to treat the row and column nodes of the adjacency matrix as distinct entities, each potentially following its own community organization. Building on this model, we develop DiSP, an efficient spectral procedure to estimate mixed memberships for both sets of nodes. Through delicate analysis, we derive node-specific error bounds of DiSP under mild sparsity conditions. Simulation results support the theoretical results, demonstrating that DiSP achieves lower error rates and faster computation than its competitor. Moreover, applications to real data highlight DiSP's effectiveness in uncovering asymmetric structural patterns.
\end{abstract}
\begin{keyword}
Community detection\sep ideal simplex\sep overlapping directed network\sep sparsity\sep spectral clustering\sep SVD
\end{keyword}
\end{frontmatter}
\section{Introduction}\label{sec1}
In the era of big data, networks with significant structural patterns are found everywhere. Illustrative examples include social networks generated by platforms such as Facebook, Twitter, WeChat, and WhatsApp, where links capture user interactions or friendships \cite{pizzuti2008ga}; and protein-protein interaction networks, which document associations between proteins \cite{Notebaart2006}; and citation networks, through which authors’ research preferences are reflected \cite{DSCORE}. To facilitate mathematical analysis, networks are typically represented as graphs, where nodes denote subjects/individuals, while edges capture the relationships between them -- including their presence, directionality, and associated weights. Community detection serves as a key tool for extracting structural information from such data.

Many community detection methods focus on undirected networks for simplicity \cite{SBM,lancichinetti2009community, RSC,SCORE, chen2018convexified}. A classic and popular model for generating undirected networks is the Stochastic Blockmodel (SBM) \cite{SBM}. Under SBM, community membership is exclusive for each node, and the probability of observing an edge between two nodes is determined purely by the pair of communities they occupy. Additionally, SBM imposes that all nodes within the same community share identical expected degrees. In real-world scenarios, however, nodes often participate in multiple communities with varying degrees -- called mixed membership (overlapping) networks.  To accommodate such cases,  \cite{MMSB} generalized SBM to handle mixed memberships, introducing the Mixed Membership Stochastic Blockmodel (MMSB). 

Real-world networks are often directed, as seen in citation networks, protein-protein interaction networks, and hyperlink networks. The added complexity of directed networks arises from the need to distinguish between two types of nodes: those that send edges and those that receive edges. In a citation network, for example, a paper that cites multiple others acts as a “sending node,” while the papers it cites are “receiving nodes.” A number of methods have been proposed to analyze such directed networks. Among them, the Stochastic co-Blockmodel (ScBM) \cite{DISIM}  is designed for directed (asymmetric) relationships, assuming that each node belongs to exactly one community (i.e., no mixed memberships).   A degree-corrected version of ScBM is designed by \cite{ji2016coauthorship}, and a related community detection algorithm for directed networks and theoretical guarantees can be found in \cite{DSCORE}.  Using non‑negative matrix factorization (NMF), \cite{lim2018codinmf} develops a flexible, noise‑resilient graph clustering method that handles community detection for both undirected and directed graphs. A method enabling simultaneous network embedding and community detection in directed networks was introduced by \cite{Zhang2022Directed}. However, these methods primarily assume that each node belongs to a single community, limiting their applicability to directed networks with mixed memberships, which are prevalent in practice.

For the directed network with mixed memberships, \cite{airoldi2013multi} proposed a multi-way stochastic blockmodel with a Dirichlet distribution, which is an extension of the MMSB model, and applied the nonparametric methods, collapsed Gibbs sampling, and variational Expectation-Maximization to make inference. This paper focuses on directed mixed‑membership networks and aims to build a provably consistent spectral algorithm for membership estimation.

The key contributions of this paper are highlighted below. 
\begin{itemize}
   \item \textbf{Statistical Model:} We develop a generative model for directed mixed-membership networks, called the Directed Mixed Membership Stochastic Blockmodel (DiMMSB). DiMMSB allows nodes to belong to multiple communities and permits senders (rows) and receivers (columns) to be distinct, i.e., the adjacency matrix can be non‑square. Identifiability of DiMMSB is confirmed under standard mixed‑membership constraints.
   \item \textbf{Recovery Algorithm:} We construct a fast spectral algorithm, DiSP. The algorithm is founded on the insight that a Row Ideal Simplex structure resides in the left singular vectors of the population adjacency matrix, and a Column Ideal Simplex structure resides in its right singular vectors. 
   \item \textbf{Node-Wise Error Bound:} To quantify the sparsity of a directed mixed membership network, we introduce the sparsity parameter. Using the row-wise singular vector deviation \cite{chen2021spectral} in DiSP, we derive upper bounds on the estimation errors for individual row and column nodes. Under mild sparsity assumptions on the network, our approach is shown to yield asymptotically consistent parameter estimates. To our knowledge, this is the first work to establish a consistent estimation for an estimation algorithm for directed mixed membership (overlapping) network models. 
   \item \textbf{Empirical Validation:} Numerical results on substantial simulated directed mixed membership networks show that DiSP is useful and fast in estimating mixed memberships, and experiments on real-world datasets highlight the benefits of DiSP for analyzing asymmetric network structures and finding highly mixed nodes in a directed network.
\end{itemize}

The structure of this paper is as follows: In Section \ref{sec2}, the DiMMSB model is proposed, with its identifiability and sparsity scaling verified in Subsections \ref{sec21} and \ref{sec22}. Section \ref{sec3} introduces the proposed method DiSP. Theoretical results for DiSP are presented in Section \ref{sec4}. Simulation results and real data analysis are provided in Sections \ref{sec5} and \ref{sec6}, respectively. Finally, Section \ref{sec7} provides a discussion of the proposed model and method.

\textbf{\textit{Notations.}}
We take the following general notations throughout. For a vector $x$, $\|x\|_{q}$ is its $l_{q}$-norm. $M'$ denotes the transpose matrix,  $\|M\|$ the spectral norm,  $\|M\|_{F}$ the Frobenius norm, $\|M\|_{2\rightarrow\infty}$ the maximum maximum row-wise $l_{2}$ norm.  $\sigma_{i}(M)$ and $\lambda_{i}(M)$ are the $i$-th llargest singular value and the corresponding eigenvalue (ordered by magnitude). $M(i,:)$ and $M(:,j)$ arethe $i$-th row and the $j$-th column of matrix $M$. $M(S_{r},:)$ and $M(:,S_{c})$ denote the rows and columns in the index sets $S_{r}$ and $S_{c}$. For any matrix $M$, let $Y=\mathrm{max}(0, M)$ denote the matrix whose $(i,j)$-th entry equals $\max(0, M_{ij})$, i.e., the positive part of $M_{ij}$.

\section{The model}\label{sec2}
This section provides a detailed description of the proposed model. First, denote $A\in \{0,1\}^{n_{r}\times n_{c}}$ be a bi-adjacency matrix, where $n_r$ and $n_c$ are numbers of rows and columns, respectively. For any entry $(i,j)$, the entry $A(i,j)$ is taken to be $1$ if a directed edge originates from row node $i$ and points to column node $j$; otherwise, $A(i,j)=0$.  (The following content is also annotated with similar symbols.). Hence, row $i$ of $A$ reflects the edge-sending pattern of row node $i$, and column $j$ reflects the edge-receiving pattern of column node $j$. Let $S_{r}$ be the set of row nodes (indices $1$ through $n_{r}$) and $S_{c}$ the set of column nodes (indices $1$ through $n_{c}$). In this paper, we assume that the row (sending) nodes can be different from the column (receiving) nodes, and the number of row nodes and the number of columns are not necessarily equal.   Our model assumes that the row nodes of $A$ fall into $K$ identifiable clusters, termed row communities  (or occasionally as sending clusters).
\begin{align}\label{DefinSC}
\mathcal{C}^{(1)}_{r},\mathcal{C}^{(2)}_{r},\ldots,\mathcal{C}^{(K)}_{r},
\end{align}
and the column nodes of $A$  fall into $K$ identifiable clusters, termed either column communities or receiving clusters. \begin{align}\label{DefinRC}
\mathcal{C}^{(1)}_{c},\mathcal{C}^{(2)}_{c},\ldots,\mathcal{C}^{(K)}_{c}.
\end{align}

Let $\Pi_{r}\in \mathbb{R}^{n_{r}\times K}$ and $\Pi_{c}\in \mathbb{R}^{n_{c}\times K}$ be row and  column nodes membership matrices, respectively. For each row  node $i$
$\Pi_{r}(i,:)$ forms a $1\times K$ Probability Mass Function (PMF),  similarly,  $\Pi_{c}(j,:)$ forms a $1\times K$ PMF for column node $j$, and
\begin{align}\label{DefineSPMF}
&\Pi_{r}(i,k) \; \text{gives the membership proportion of row node } i \text{ assigned to row community } \mathcal{C}^{(k)}_{r},\quad 1\leq k\leq K,\\
&\Pi_{c}(j,k) \; \text{gives the membership proportion of column node } j \text{ assigned to column community } \mathcal{C}^{(k)}_{c},\quad 1\leq k\leq K.
\end{align}

A row node $i$ is termed “pure” whenever $\Pi_{r}(i,:)$ is degenerate (one entry equals 1, all others 0), and “mixed” otherwise. Column nodes follow the same convention.

We denote by $P\in\mathbb{R}^{K\times K}$ a \textit{probability matrix}, the elements of which are bounded as follows:
\begin{align}\label{ConB}
P(k,l) \in [0,1], \qquad \text{for all } k,l = 1,\dots,K.
\end{align}

Observe that, because the present work focuses on directed networks with mixed memberships, the probability matrix $P$ is not required to be symmetric. For every pair $(i,j)$ with $1\le i\le n_r$ and $1\le j\le n_c$, the entries $A(i,j)$ are modeled as independent Bernoulli variables whose success probabilities are given by a bilinear form:
\begin{align}\label{DefinP}
\mathbb{P}(A(i,j)=1) = \sum_{k=1}^{K}\sum_{l=1}^{K} \Pi_r(i,k),\Pi_c(j,l),P(k,l).
\end{align}

\begin{defin}
The model specified by Equations (\ref{DefinSC})–(\ref{DefinP}) is referred to as the Directed Mixed Membership Stochastic Blockmodel, abbreviated as DiMMSB, and is denoted by $\mathrm{DiMMSB}(n_r, n_c, K, P, \Pi_r, \Pi_c)$.
\end{defin}

DiMMSB can be seen as a unified framework that encompasses and extends multiple previous models.
\begin{itemize}
\item DiMMSB is reduced to the ScBM with $K$ row and $K$ column clusters under the condition that all row and column nodes are pure.  \cite{DISIM}.
\item When $\Pi_{r}(i,:)$ and $ \Pi_{c}(j,:)$ follow Dirichlet distribution for $1\leq i\leq n_{r}$ and $1\leq j\leq n_{c}$, DiMMSB reduces to the two-way stochastic blockmodels with Bernoulli distribution \cite{airoldi2013multi}.
\item When $\Pi_{r}=\Pi_{c}$ and $ P=P'$, $\Pi_{r}(i,:)$ follow Dirichlet distribution for $1\leq i\leq n_{r}$, and all row nodes and column nodes are the same, DiMMSB reduces to MMSB \cite{MMSB}.
\item When $\Pi_{r}=\Pi_{c}$ and $ P=P'$, the row and column node sets are identical, and every node is pure, DiMMSB reduces to SBM \cite{SBM}.
\end{itemize}

\begin{figure}
	\centering
	\subfigure[]{\includegraphics[width=0.512\textwidth]{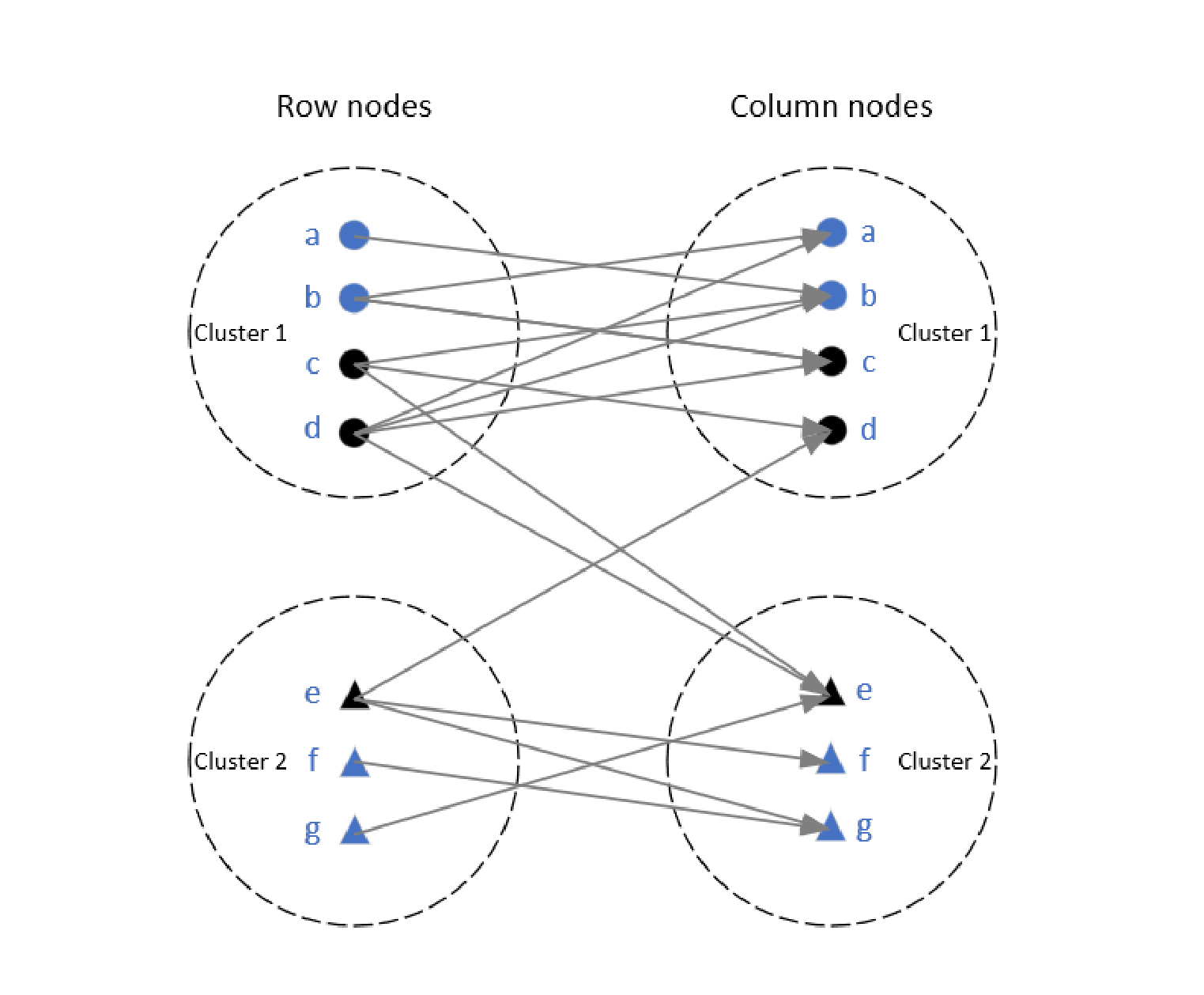}}
	\subfigure[]{\includegraphics[width=0.48\textwidth]{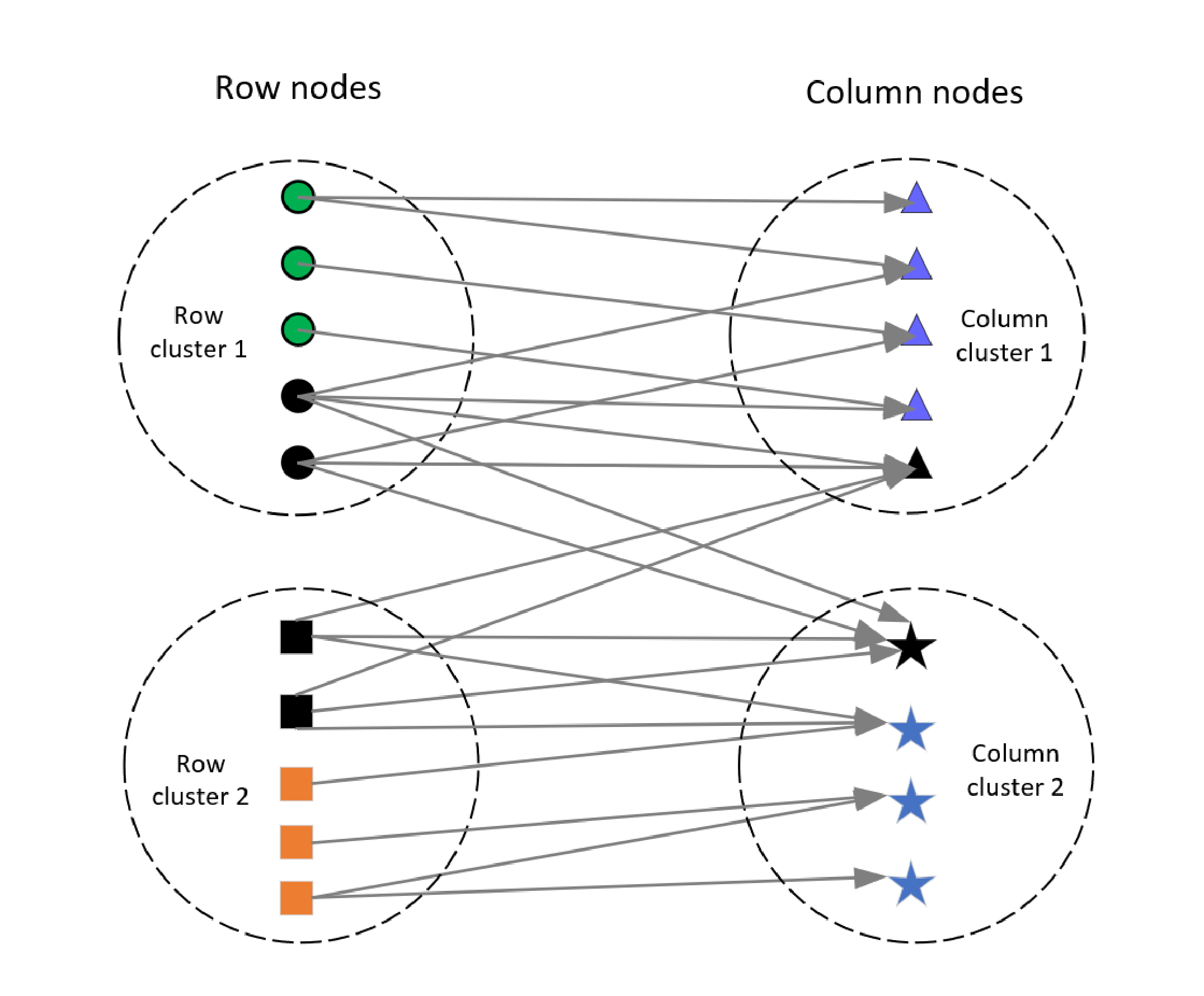}}
	\caption{Two schematic diagrams for DiMMSB.}
	\label{SchematicDigramsDiMMSB}
\end{figure}
Fig.~\ref{SchematicDigramsDiMMSB} provides two schematic diagrams that concisely demonstrate the generality of DiMMSB.  In the figure, an arrow demonstrates a directed edge from one node to another. Dashed circles enclose nodes within the same cluster, and nodes with mixed memberships are colored black. In panel (a) of Fig.~\ref{SchematicDigramsDiMMSB} illustrates a scenario where row and column nodes are identical. This network contains 7 nodes, so $A\in\mathbb{R}^{7\times 7}$. Within this network, we observe four nodes--$a$, $b$, $c$, and $d$--each affiliated with both the first row community and the first column community. The remaining three nodes, $e$, $f$, and $g$, are associated exclusively with the second communities (both row and column). Turning to the link patterns: $c$ and $d$ each direct an edge toward $e$, while $e$ sends an edge back to $d$. These reciprocal-like yet asymmetric connections cause $c$, $d$, and $e$ to be regarded as nodes with overlapping community affiliations (i.e., mixed memberships). Panel (b) of this figure, in contrast, illustrates a completely different configuration in which the ensemble of row nodes differs from that of column nodes. On the row side, ten nodes appear: those drawn as filled circles fall into row cluster~1, whereas those drawn as filled squares fall into row cluster 2. On the column side, there are nine nodes: filled triangles denote membership in column cluster 1, and filled stars denote membership in column cluster 2. As a direct consequence, the bi-adjacency matrix $A$ shown in this panel has dimension $10$ (rows) by $9$ (columns). Among these row nodes, the black circles and black squares both have outgoing edges to the black triangle node and the black star node; therefore, these row nodes are considered mixed. Conversely, since the black triangle and black star nodes receive edges from these mixed row nodes, they are classified as mixed column nodes. 
\begin{rem}
It is important to note that the definition of the DiMMSB model, as presented in Equation (\ref{DefinP}), necessitates the existence of a clearly defined block structure, which is captured by the $K\times K$ probability matrix $P$. This matrix $P$ enhances the adaptability and versatility of our DiMMSB model. To illustrate, consider the scenario where $P$ is a $K\times K$ identity matrix. In this case, Equation (\ref{DefinP}) simplifies to $\mathbb{P}(A(i,j)=1)=\sum_{k=1}^{K}\sum_{l=1}^{K}\Pi_{r}(i,k)\Pi_{c}(j,l)$, a formulation that is considerably less flexible compared to the original equation. Furthermore, it is worth mentioning that other statistical models discussed herein, including the ScBM, two-way stochastic blockmodels, MMSB, and SBM, also rely on the existence of $P$ to augment their flexibility and applicability.
\end{rem}
\subsection{Identifiability}\label{sec21}
To ensure that the DiMMSB model is identifiable, certain constraints must be imposed on its parameters. For community-based models, identifiability means that the model parameters can be uniquely determined up to a relabeling of the communities [\cite{MixedSCORE,mao2020estimating}]. For the proposed DiMMSB framework, the following two conditions are sufficient to guarantee identifiability:
\begin{itemize}
\item \textbf{(I1)} The connectivity matrix $P$ has full rank, i.e., $\operatorname{rank}(P)=K$.
\item \textbf{(I2)} Each of the $K$ row communities and each of the $K$ column communities contains at least one pure node.
\end{itemize}

The full rank condition (I1) for the connectivity matrix $P$ and pure nodes condition (I2) are popular conditions for models modeling networks with mixed memberships, see \cite{MixedSCORE,mao2020estimating}. In practical social networks, the full rank condition (I1) is a reasonable assumption, as communities tend to be sufficiently distinct and well-connected within themselves. This condition ensures that the model is identifiable and capable of distinguishing between different communities. Notably, the pure nodes condition (I2) is also very mild, as it simply requires that each community has at least one representative node. In real-world social networks, it is common for communities to have core members who are more closely associated with their respective communities than with others.

We begin by separating the observed adjacency matrix $A$ into two additive components: a signal'' term and anoise'' term,
\begin{align}
A = \Omega + W,
\end{align}
where $\Omega$ (an $n_r \times n_c$ matrix) represents the expected value of $A$, and $W$ captures the random fluctuations. Under the DiMMSB specification, this signal matrix takes the following low‑rank form:
\begin{align}\label{Omega}
\Omega = \Pi_r P \Pi_c'.
\end{align}

The matrix $\Omega$  is termed the population adjacency matrix. By basic algebra, we know $\Omega$ is of rank $K$, which is smaller than $\min\{n_r, n_c\}$. This low-rank property is precisely what enables spectral clustering methods to be effective for DiMMSB.

The following proposition establishes that DiMMSB is identifiable whenever conditions (I1) and (I2) are satisfied.
\begin{prop}\label{id}
Under conditions (I1) and (I2), the DiMMSB model enjoys identifiability: any representation of $\Omega$ as $\Pi_r P \Pi_c'$ yields parameters that are unique except for a permutation of the community labels.
\end{prop}
Henceforth, conditions (I1) and (I2) are adopted as default unless otherwise specified.

\subsection{Sparsity scaling}\label{sec22}
In practice, many large-scale networks exhibit sparsity: the degree of a node (i.e., its number of connections) is often small compared to the overall number of nodes. Under these sparse conditions, recovering community structure is generally difficult. Therefore, a key metric for assessing any community recovery algorithm is how well it performs across varying degrees of sparsity. To quantify this property for directed networks with mixed memberships, we introduce a sparsity parameter denoted by $\rho$, where
\begin{align*}
P=\rho\tilde{P}\mathrm{~where~}\mathrm{max}_{1\leq k,l\leq K}\tilde{P}(k,l)=1.
\end{align*}

Under $DiMMSB(n_{r}, n_{c}, K, P, \Pi_{r}, \Pi_{c})$, a smaller $\rho$ leads to a smaller probability of generating an edge from the row node $i$ to the column node $j$, i.e., the sparsity parameter $\rho$ captures the sparsity behavior to generate a directed mixed membership network. It is common to control network sparsity when deriving theoretical studies on the estimation consistency of spectral clustering methods \cite{SCORE,DISIM,lei2015consistency}. Especially, when DiMMSB degenerates to SBM, Assumption \ref{a1} matches the sparsity requirement in Theorem 3.1 in \cite{lei2015consistency}, and this guarantees the optimality of our sparsity condition.  Meanwhile, as mentioned in \cite{MixedSCORE,mao2020estimating}, the separation between communities is captured by $\sigma_{K}(\tilde{P})$, with larger values corresponding to more sharply distinguished communities.  This paper also aims to study the effect of $\rho$ and $\sigma_{K}(\tilde{P})$ on the performance of spectral clustering by allowing them to be contained in the error bound. Therefore, our theoretical results allow the model parameters $K,\rho,\sigma_{K}(\tilde{P})$ to vary with $n_{r}$ and $n_{c}$.

\section{A spectral algorithm}\label{sec3}
Our algorithm primarily targets the estimation of the row and column membership matrices, $\Pi_{r}$  and $\Pi_{c}$ 
 from the observed data $A$ when $K$ is known. Considering computational scalability, we focus on the idea of spectral clustering in this paper.

We next explain the rationale behind our proposed algorithm. Under conditions (I1) and (I2), we can obtain $\mathrm{rank}(\Omega)=K$,  and $K\ll \mathrm{min}\{n_{r}, n_{c}\}$. Write the compact singular value decomposition (SVD) of $\Omega$ as $\Omega=U\Lambda V'$, with $U\in\mathbb{R}^{n_{r}\times K}, \Lambda\in\mathbb{R}^{K\times K}, V\in\mathbb{R}^{n_{c}\times K}$, $U'U=I_{K}, V'V=I_{K}$ (where $I_{K}$ denotes the $K\times K$ identity matrix). For each community index $k$ ($1\leq k\leq K$), define the sets $\mathcal{I}^{(k)}_{r}=\{i\in\{1,2,\ldots, n_{r}\}: \Pi_{r}(i,k)=1\}$ and $\mathcal{I}^{(k)}_{c}=\{j\in \{1,2,\ldots, n_{c}\}: \Pi_{c}(j,k)=1\}$. Condition (I2) guarantees that both $\mathcal{I}^{(k)}_{r}$ and $\mathcal{I}^{(k)}_{c}$ are non-empty for  $k=1, \dots, K$. We then pick exactly one row node from each $\mathcal{I}^{(k)}_{r}$ to form an index set $\mathcal{I}_{r}$, thus $\mathcal{I}_{r}$ contains the indices of $K$ pure row nodes, one per community. The column index set  $\mathcal{I}_{c}$ is constructed analogously. Without loss of generality, we arrange the ordering so that $\Pi_{r}(\mathcal{I}_{r},:)=I_{K}$ and $\Pi_{c}(\mathcal{I}_{c},:)=I_{K}$ (Lemma 2.1 in \cite{mao2020estimating} also has similar setting to design their spectral algorithms under MMSB.). The following lemma confirms the presence of a Row Ideal Simplex (RIS) structure in $U$ and a Column Ideal Simplex (CIS) structure in $V$.
\begin{lem}\label{RISCIS}
(Row and Column Ideal Simplices). Under the $DiMMSB(n_{r}, n_{c}, K, P,\Pi_{r}, \Pi_{c})$, there exist a unique $K\times K$ matrix $B_{r}$ and a unique $K\times K$ matrix $B_{c}$ satisfying
\begin{itemize}
\item $U=\Pi_{r}B_{r}$ where $B_{r}=U(\mathcal{I}_{r},:)$. Meanwhile, $U(i,:)=U(\bar{i},:)$, if $\Pi_{r}(i,:)=\Pi_{r}(\bar{i},:)$ for $1\leq i,\bar{i}\leq n_{r}$.
  \item $V=\Pi_{c}B_{c}$ where $B_{c}=V(\mathcal{I}_{c},:)$. Meanwhile, $V(j,:)=V(\bar{j},:)$, if $\Pi_{c}(j,:)=\Pi_{c}(\bar{j},:)$ for $1\leq j,\bar{j}\leq n_{c}$.
\end{itemize}
\end{lem}
Lemma \ref{RISCIS} says that the rows of $U$ form a $K$-simplex in $\mathbb{R}^{K}$ which we call the Row Ideal Simplex (RIS), with the $K$ rows of $B_{r}$ being the vertices. Similarly,  rows of $V$ form a $K$-simplex in $\mathbb{R}^{K}$ which we call the Column Ideal Simplex (CIS), with the $K$ rows of $B_{c}$ being the vertices. Meanwhile, $U(i,:)$ is a convex linear combination of $B_{r}(1,:), B_{r}(2,:), \ldots, B_{r}(K,:)$ for $1\leq i\leq n_{r}$.  If row node $i$ is pure,  $U(i,:)$ lies exactly on one of the vertices of the RIS. If row node $i$ is mixed, $U(i,:)$ is in the interior or face of the RIS, but not on any of the vertices. Similar conclusions hold for column nodes.

Since $B_{r}$ and $B_{c}$ are full-rank matrices, if $U, V, B_{r}$ and $B_{c}$ are known in advance ideally, we can exactly obtain  $\Pi_{r}$ and $\Pi_{c}$ by setting $\Pi_{r}=UB_{r}'(B_{r}B_{r}')^{-1}$ and  $\Pi_{c}=VB_{c}'(B_{c}B_{c}')^{-1}$.

While in practice, the estimation of $UB_{r}'(B_{r}B_{r}')^{-1}$ and $VB_{c}'(B_{c}B_{c}')^{-1}$ may not have unit row norm, thus we need to make the following transformation:
Set $Y_{r}=UB_{r}'(B_{r}B_{r}')^{-1},$ and $ Y_{c}=VB_{c}'(B_{c}B_{c}')^{-1}$. 
Then the membership matrices can be estimated by
\begin{align*}
\Pi_{r}(i,:)=\frac{Y_{r}(i,:)}{\|Y_{r}(i,:)\|_{1}},\Pi_{c}(j,:)=\frac{Y_{c}(j,:)}{\|Y_{c}(j,:)\|_{1}}, 1\leq i\leq n_{r},1\leq j\leq n_{c}.
\end{align*}

The RIS structure implies $U=\Pi_{r}B_{r}\equiv \Pi_{r}U(\mathcal{I}_{r},:)$. Hence, exact recovery of $\Pi_{r}$ can be achieved once the row corner matrix $U(\mathcal{I}_{r},:)$ (i.e., $B_{r}$) is obtained. As pointed out in \cite{MixedSCORE} and \cite{mao2020estimating}, applying the successive projection (SP) algorithm \cite{gillis2015semidefinite} to the left singular vectors $U$ with the prescribed number $K$ of row communities yields $B_{r}$; the specifics of this algorithm are given in Algorithm \ref{alg:SP}. This reasoning outlines the ideal procedure for recovering $\Pi_{r}$ given  $K$ and $\Omega$ under DiMMSB. Similarly, $\Pi_{c}$ can be recovered exactly by applying SP on all rows of $V$ with $K$ column communities.

The foregoing analysis leads us to propose a three-stage algorithm. We refer to this method as Ideal DiSP. 

Input $\Omega$ and $K$. Output: $\Pi_{r}$ and $\Pi_{c}$.
\begin{itemize}
  \item \texttt{PCA step.} Take the compact SVD of $\Omega$ to be $\Omega = U\Lambda V'$, in which $U$ and $V$ are column‑orthogonal matrices ($U'U = I$, $V'V = I$) of sizes $n_r\times K$ and $n_c\times K$, respectively, and $\Lambda$ is a $K\times K$ diagonal matrix.
  \item \texttt{Vertex Hunting (VH) step.} Run SP algorithm on all rows of $U$  (and $V$) assuming there are $K$ row (column) communities to obtain $B_{r}$ (and $B_{c}$).
  \item \texttt{Membership Reconstruction (MR) step.} Set  $Y_{r}=UB_{r}'(B_{r}B_{r}')^{-1}$ and\\ $Y_{c}=UB_{c}'(B_{c}B_{c}')^{-1}$.  Recover $\Pi_{r}$ and $\Pi_{c}$ by setting $\Pi_{r}(i,:)=\frac{Y_{r}(i,:)}{\|Y_{r}(i,:)\|_{1}}$ for $1\leq i\leq n_{r}$, and $\Pi_{c}(j,:)=\frac{Y_{c}(j,:)}{\|Y_{c}(j,:)\|_{1}}$ for $1\leq j\leq n_{c}$.
\end{itemize}

Exact recovery of node memberships by Ideal DiSP is guaranteed by the following theorem, which in turn serves as a verification of DiMMSB's identifiability.
\begin{thm}\label{IdealDiSP}
(Ideal DiSP). Under $DiMMSB(n_{r}, n_{c}, K, P,\Pi_{r}, \Pi_{c})$, the ideal version of DiSP perfectly recovers $\Pi_{r}$ and $\Pi_{c}$.
\end{thm}

Moving from the idealized setting to real‑data applications, we consider the truncated SVD of $A$ to the top $K$ dimensions: $\tilde{A} = \hat{U}\hat{\Lambda}\hat{V}'$. In this decomposition, $\hat{U}$ and $\hat{V}$ are $n_r\times K$ and $n_c\times K$ column‑orthogonal matrices ($\hat{U}'\hat{U}=I_K$, $\hat{V}'\hat{V}=I_K$), and $\hat{\Lambda}$ is a $K\times K$ diagonal matrix holding the $K$ largest singular values of $A$. For the empirical case, the estimates $\hat{B}_r$, $\hat{B}_c$, $\hat{Y}_r$, $\hat{Y}_c$, $\hat{\Pi}_r$, $\hat{\Pi}_c$ (obtained via Algorithm~\ref{alg:DiSP}) serve as proxies for the true quantities $B_r$, $B_c$, $Y_r$, $Y_c$, $\Pi_r$, $\Pi_c$. Algorithm \ref{alg:DiSP}, termed DiSP, is a straightforward adaptation of Ideal DiSP to real-world data.
\begin{algorithm}
\caption{\textbf{DiSP}}
\label{alg:DiSP}
\begin{algorithmic}[1]
\Require $A\in \mathbb{R}^{n_{r}\times n_{c}}$ and $K$.
\Ensure The estimated row and column membership matrix $\hat{\Pi}_{r}$ and $\hat{\Pi}_{c}$.
\State \texttt{PCA step.} Compute the left and right singular vectors $\hat{U}\in\mathbb{R}^{n_{r}\times K}$ and $\hat{V}\in \mathbb{R}^{n_{c}\times K}$ of $A$.
\State \texttt{VH step.} Apply SP algorithm on the rows of $\hat{U}$ and $\hat{V}$  respectively,  to get the near-corners matrix $\hat{B}_{r}\equiv \hat{U}(\mathcal{\hat{I}}_{r},:)\in\mathbb{R}^{K\times K}$ and $\hat{B}_{c}\equiv \hat{V}(\mathcal{\hat{I}}_{c},:)\in\mathbb{R}^{K\times K}$, where $\mathcal{\hat{I}}_{r}$ and $\mathcal{\hat{I}}_{c}$ are SP's estimated index sets. 
\State \texttt{MR step.} Compute $\hat{Y}_{r}=\hat{U}\hat{B}_{r}'(\hat{B}_{r}\hat{B}_{r}')^{-1}$ and $\hat{Y}_{c}=\hat{V}\hat{B}_{c}'(\hat{B}_{c}\hat{B}_{c}')^{-1}$. Set $\hat{Y}_{r}=\mathrm{max}(0, \hat{Y}_{r})$ and $\hat{Y}_{c}=\mathrm{max}(0, \hat{Y}_{c})$.  Finally, obtain $\hat{\Pi}_{r}(i,:)=\hat{Y}_{r}(i,:)/\|\hat{Y}_{r}(i,:)\|_{1}, 1\leq i\leq n_{r}$ and $\hat{\Pi}_{c}(j,:)=\hat{Y}_{c}(j,:)/\|\hat{Y}_{c}(j,:)\|_{1}, 1\leq j\leq n_{c}$.
\end{algorithmic}
\end{algorithm}

Since row node weights are inherently nonnegative, we apply a truncation operation in the MR step by setting $\hat{Y}_{r}=\mathrm{max}(0, \hat{Y}_{r})$. Such correction is needed because the matrix product $\hat{U}\hat{B}_{r}'(\hat{B}_{r}\hat{B}_{r}')^{-1}$ can produce negative values. Additionally, the invertibility of $\hat{B}_{r}\hat{B}_{r}'$ is practically guaranteed, as $\hat{B}_{r}$ contains $K$ distinct rows and $n_{r}\gg K$. Same for columns.

\begin{figure}
	\centering \subfigure[$U$]{\includegraphics[width=0.4\textwidth]{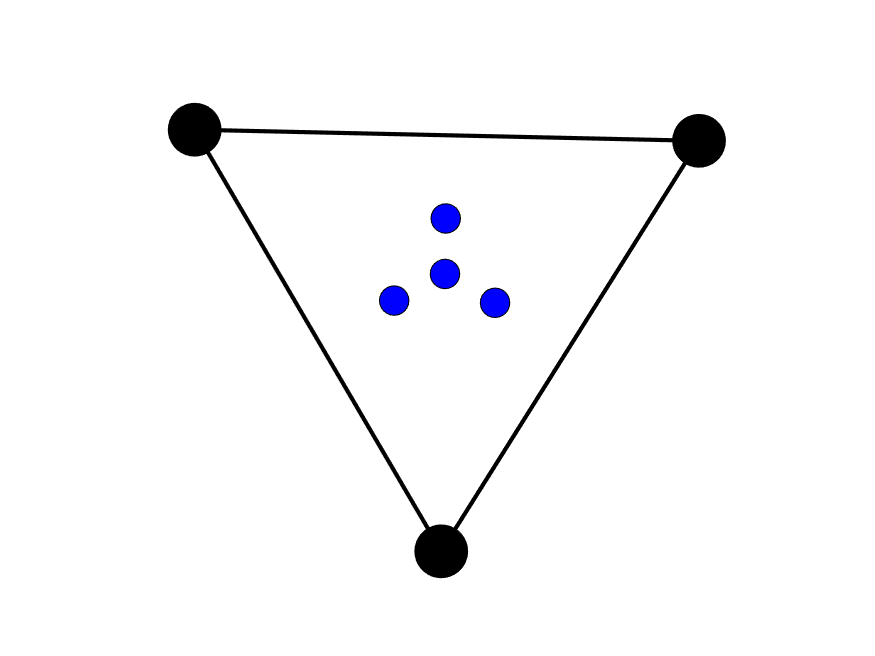}} \subfigure[$V$]{\includegraphics[width=0.4\textwidth]{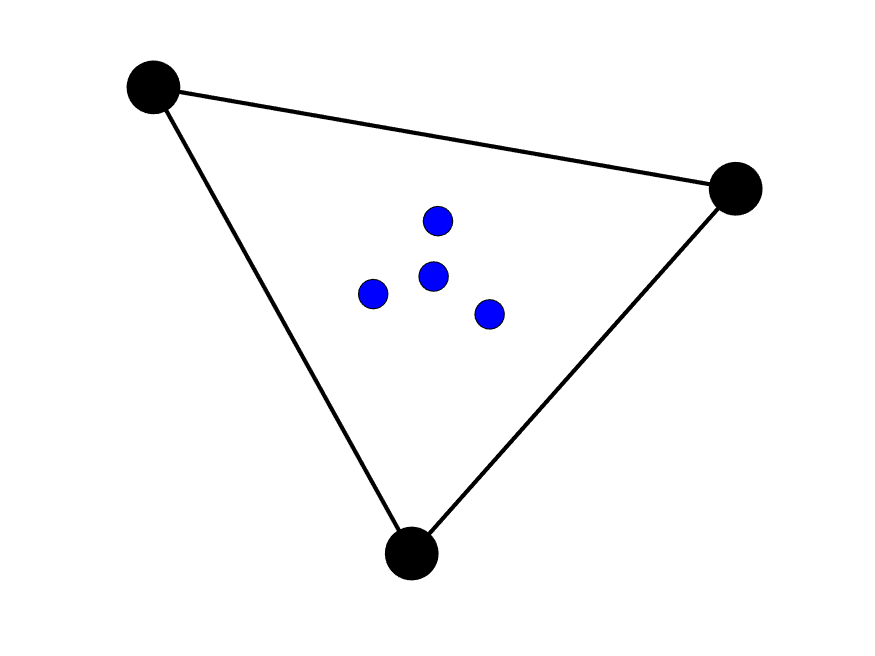}}
	\subfigure[$\hat{U}$ for $n_{r,0}=60$]{\includegraphics[width=0.32\textwidth]{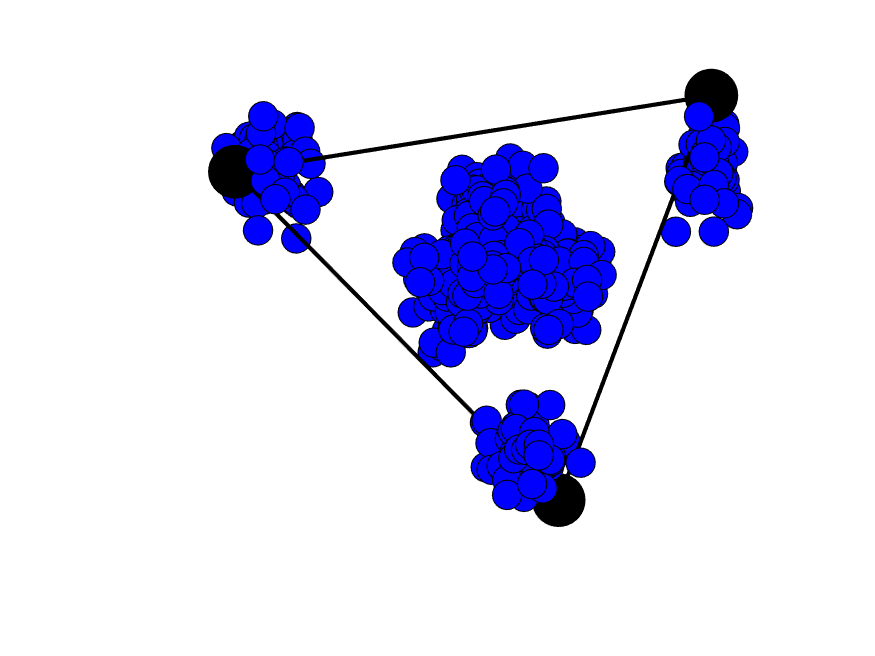}}
	\subfigure[$\hat{U}$ for $n_{r,0}=120$]{\includegraphics[width=0.32\textwidth]{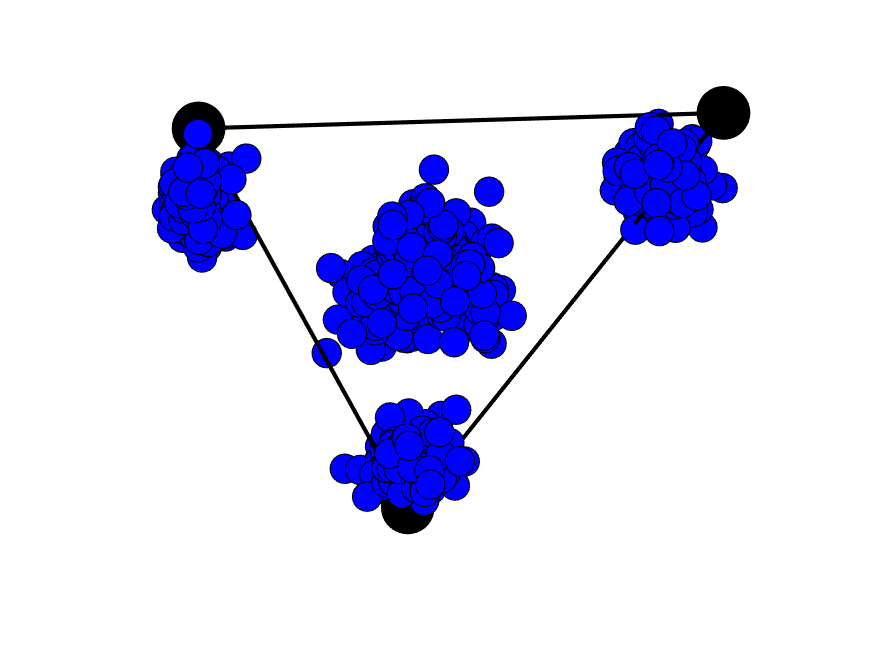}}
	\subfigure[$\hat{U}$ for $n_{r,0}=180$]{\includegraphics[width=0.32\textwidth]{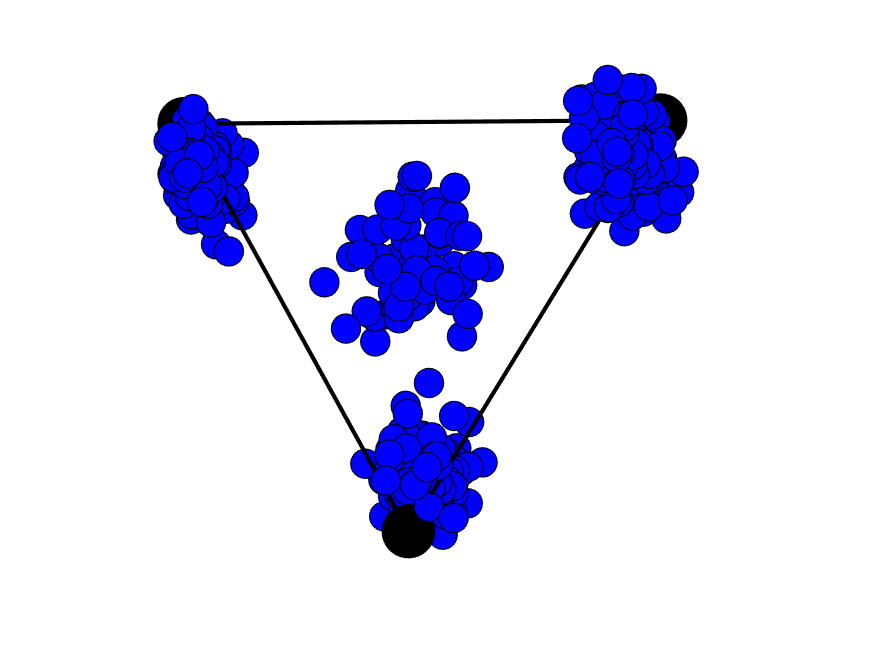}}
	\subfigure[$\hat{V}$ for $n_{c,0}=40$]{\includegraphics[width=0.32\textwidth]{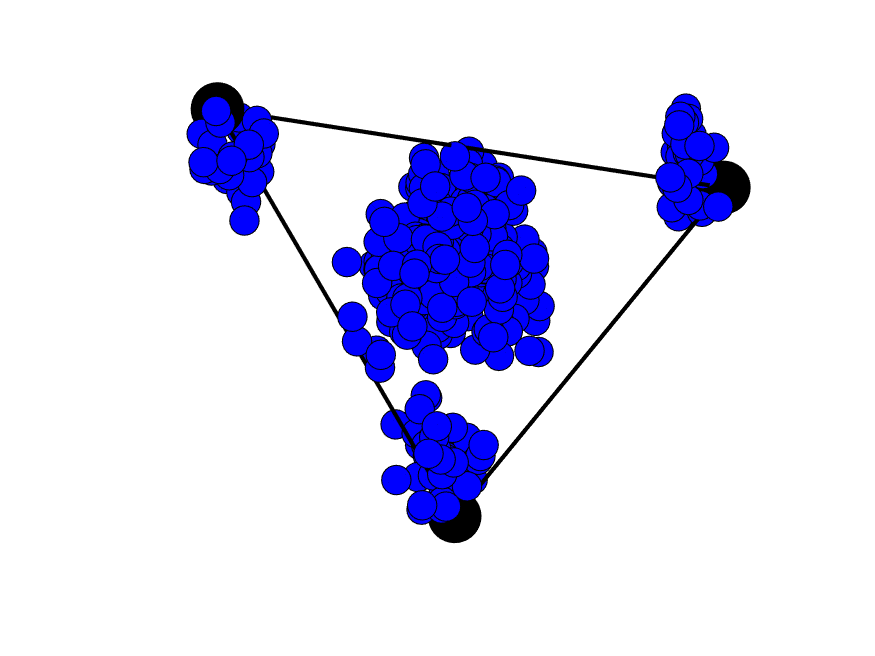}}
	\subfigure[$\hat{V}$ for $n_{c,0}=100$]{\includegraphics[width=0.32\textwidth]{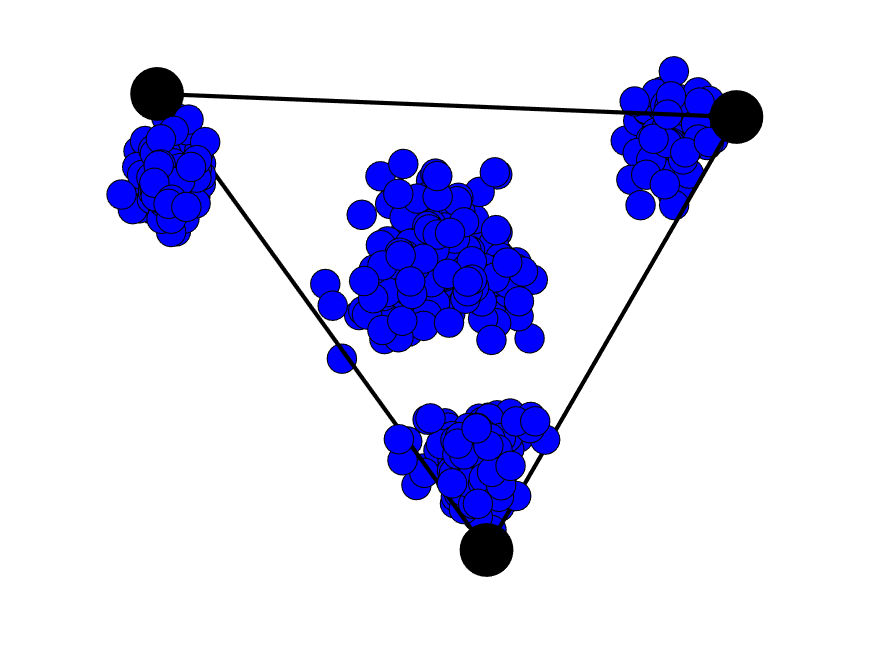}}
	\subfigure[$\hat{V}$ for $n_{c,0}=160$]{\includegraphics[width=0.32\textwidth]{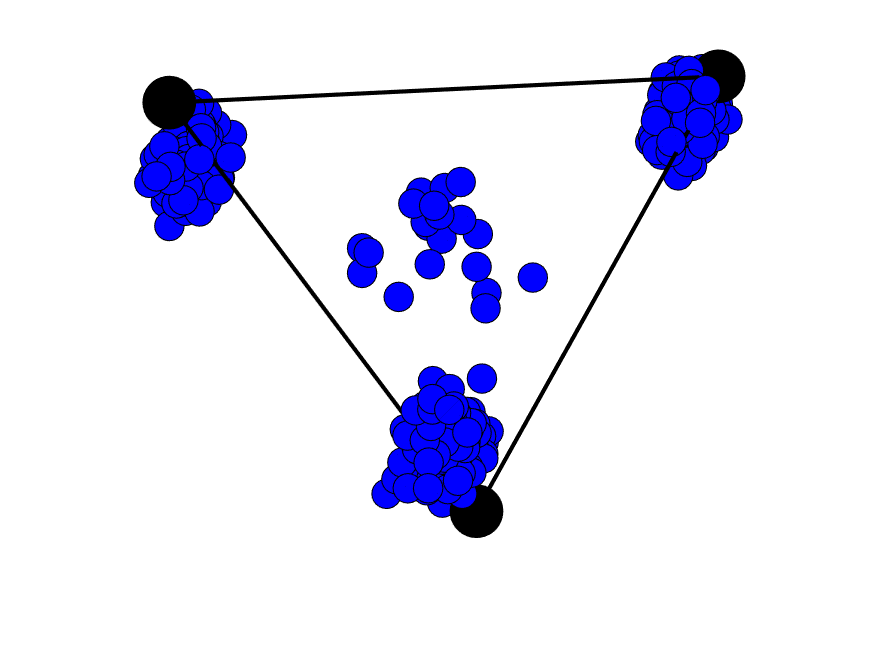}}
	\caption{Panel (a) shows the RIS in Experiment 4. Here, pure‑node counts are $n_{r,0}=n_{c,0}=120$ ($n_{r,0}$: count of pure row nodes, $n_{c,0}$: count of pure column nodes). Color encoding: black for pure rows, blue for mixed rows. Each dot corresponds to one row of $U$; coincident rows share a dot. All mixed nodes are divided into four equal groups with probability mass vectors $(0.4,0.4,0.2)$, $(0.4,0.2,0.4)$, $(0.2,0.4,0.4)$, and $(1/3,1/3,1/3)$. Panel (b) shows the CIS under identical settings. Panels (c)–(h) plot empirical $\hat{U}$ and $\hat{V}$ for varying pure counts, where the black dot is SP's estimated pure nodes. Because $K=3$, original $\mathbb{R}^3$ points are projected and rotated to $\mathbb{R}^2$ for visualization.}
\label{PlotUV}
\end{figure}

To demonstrate RIS and CIS, we drew Fig.~\ref{PlotUV}. Panel (a) demonstrates the following: for a pure row node $i$, the vector $U(i,:)$ is located at a vertex of the RIS; if the row node is mixed, $U(i,:)$ lies strictly inside the RIS. The same behavior applies to $V$. Panels (c)-(h) plot $\hat{U}$ and $\hat{V}$ under different row/column pure node counts, using DiMMSB data from Experiment 4. Additionally, panels (c)-(h) of Fig.~\ref{PlotUV} display the $\hat{B}_{r}$ and $\hat{B}_{c}$. From panels (c)-(e), one can see that points in $\hat{U}$ belonging to the same row community form significantly tighter clusters than those from different row communities. Moreover, as $n_{r,0}$ grows,  the count of points lying inside the triangular region (i.e., interior of the RIS) diminishes. Similar arguments hold for $\hat{V}$.

\section{Theoretical  results for DiSP}\label{sec4}
We now prove the consistency of our proposed method. Specifically, we demonstrate that the empirical estimates $\hat{\Pi}_{r}$ and $\hat{\Pi}_{c}$ are concentrated near the true membership matrices $\Pi_{r}$ and $\Pi_{c}$. Here and subsequently, we take $K$ (the total number of row/column communities) as a given positive constant.

We first bound $\|A-\Omega\|$ using the rectangular version of the Bernstein inequality \citep{tropp2012user}. his approach handles rectangular random matrices and is key to DiSP's performance when $n_{r}\neq n_{c}$. We assume that
\begin{assum}\label{a1}
$\rho \mathrm{max}(n_{r},n_{c})\geq \mathrm{log}(n_{r}+n_{c}).$
\end{assum}

Assumption \ref{a1} ensures that the network cannot be too sparse, allowing the spectral clustering method to operate effectively. This assumption is notably mild, as it only sets a very small lower bound for the sparsity parameter $\rho$ by requiring $\rho\geq\frac{\log(n_{r}+n_{c})}{\max(n_{r},n_{c})}$. For instance, considering a bipartite network with $n_{r}=2000$ and $n_{c}=4000$, we obtain $\rho\geq\frac{\log(n_{r}+n_{c})}{\max(n_{r},n_{c})}\approx0.0022$, demonstrating that even a very small value of $\rho$ can satisfy this condition. Furthermore, our Assumption 1 aligns with the sparsity requirements found in various existing works, including the assumption in Theorem 3.1 of \citep{lei2015consistency}, Assumption 3.1 of \citep{mao2020estimating}, Assumption 1 of \cite{qing2023community}, Assumption (A1) of \citep{qing2023regularized}, and Assumption (C2) of \citep{guo2023randomized}. In many real-world social networks, despite their overall low density, there are typically sufficient connections within communities to uphold this assumption. When this assumption holds, we have the following lemma.
\begin{lem}\label{BoundAOmega}
Given a $DiMMSB(n_{r}, n_{c}, K, P,\Pi_{r}, \Pi_{c})$ network fulfilling Assumption~\ref{a1}, it holds for every $\alpha>0$ that, with probability at least $1 - o((n_{r}+n_{c})^{-\alpha})$,
\begin{align*}
\|A-\Omega\| = O\!\left(\sqrt{\rho \,\max(n_{r}, n_{c})\, \log(n_{r} + n_{c})}\right).
\end{align*}
\end{lem}

Then we can obtain the row-wise deviation bound for the singular eigenvectors of $\Omega$.
\begin{lem}\label{rowwiseerror}
	(Row-wise singular eigenvector error) Under DiMMSB, suppose that $\sigma_{K}(\Omega)\geq C\sqrt{\rho (n_{r}+n_{c})\mathrm{log}(n_{r}+n_{c})}$ and Assumption \ref{a1} is satisfied, with high probability, we have
\begin{align*}
&\mathrm{max}(\|\hat{U}\hat{U}'-UU'\|_{2\rightarrow\infty}, \|\hat{V}\hat{V}'-VV'\|_{2\rightarrow\infty})=O(\frac{\sqrt{K}(\kappa(\Omega)\sqrt{\frac{\mathrm{max}(n_{r},n_{c})\mu}{\mathrm{min}(n_{r},n_{c})}}+\sqrt{\mathrm{log}(n_{r}+n_{c})})}{\sqrt{\rho}\sigma_{K}(\tilde{P})\sigma_{K}(\Pi_{r})\sigma_{K}(\Pi_{c})}),
\end{align*}
where $\mu=\mathrm{max}(\frac{n_{r}\|U\|^{2}_{2\rightarrow\infty}}{K},\frac{n_{c}\|V\|^{2}_{2\rightarrow\infty}}{K})$.
\end{lem}

To simplify the notation, set $\varpi=\mathrm{max}(\|\hat{U}\hat{U}'-UU'\|_{2\rightarrow\infty}, \|\hat{V}\hat{V}'-VV'\|_{2\rightarrow\infty})$. When $n_{r}=n_{c}=n$ and $ \Pi_{r}=\Pi_{c}=\Pi$, DiMMSB degenerates to MMSB. Note that when $\lambda_{K}(\Pi'\Pi)=O(\frac{n}{K})$ and $K=O(1)$, the theoretical result in Lemma \ref{rowwiseerror} can be simplified as $O(\frac{1}{\sigma_{K}(\tilde{P})}\frac{1}{\sqrt{n}}\sqrt{\frac{\mathrm{log}(n)}{\rho n}})$, which is consistent with Lemma 2.1 in \cite{MixedSCORE}. In detail, by setting $\Theta$ in \cite{MixedSCORE} as $\sqrt{\rho}I$ to degenerate their DCMM to MMSB, and translating their assumptions to $\lambda_{K}(\Pi'\Pi)=O(\frac{n}{K})$, when the number of communities is a constant, the 4th theoretical bound displayed in Lemma 2.1 in \cite{MixedSCORE} is the same as our reduced bound. Then if we further assume that $\sigma_{K}(\tilde{P})=O(1)$, the bound is of order $\frac{1}{\sqrt{n}}\sqrt{\frac{\mathrm{log}(n)}{\rho n}}$, aligning with \cite{lei2019unified}'s result shown in their Table 2.

Next, we bound the vertex centers matrix obtained by the SP algorithm.
\begin{lem}\label{boundC}
	Under $DiMMSB(n_{r}, n_{c}, K, P,\Pi_{r}, \Pi_{c})$, when conditions in Lemma \ref{rowwiseerror} hold, there exist two permutation matrices $\mathcal{P}_{r}$ and $\mathcal{P}_{c}$ in $\mathbb{R}^{K\times K}$ such that with high probability, 
\begin{align*}
&\mathrm{max}_{1\leq k\leq K}\|e'_{k}(\hat{U}_{2}(\mathcal{\hat{I}}_{r},:)-\mathcal{P}'_{r}U_{2}(\mathcal{I}_{r},:))\|_{F}=O(\varpi\kappa(\Pi'_{r}\Pi_{r})),\\
&\mathrm{max}_{1\leq k\leq K}\|e'_{k}(\hat{V}_{2}(\mathcal{\hat{I}}_{c},:)-\mathcal{P}'_{c}V_{2}(\mathcal{I}_{c},:))\|_{F}=O(\varpi\kappa(\Pi'_{c}\Pi_{c})).
\end{align*}
\end{lem}
\begin{lem}\label{boundY}
Under $DiMMSB(n_{r}, n_{c}, K, P,\Pi_{r}, \Pi_{c})$, when conditions in Lemma \ref{rowwiseerror} hold,, with high probability, for all $i, j$, 
\begin{align*}
&\|e'_{i}(\hat{Y}_{r}-Y_{r}\mathcal{P}_{r})\|_{F}=O(\varpi\kappa(\Pi'_{r}\Pi_{r})\sqrt{K\lambda_{1}(\Pi'_{r}\Pi_{r})}),\|e'_{j}(\hat{Y}_{c}-Y_{c}\mathcal{P}_{c})\|_{F}=O(\varpi\kappa(\Pi'_{c}\Pi_{c})\sqrt{K\lambda_{1}(\Pi'_{c}\Pi_{c})}).
\end{align*}
\end{lem}

The following theorem presents the theoretical error bounds for the estimated memberships of both row and column nodes, serving as the central theoretical finding of our DiSP approach.
\begin{thm}\label{Main}
Under $DiMMSB(n_{r}, n_{c}, K, P,\Pi_{r}, \Pi_{c})$, when conditions in Lemma \ref{rowwiseerror} are satisfied, with high probability,
\begin{align*}
&\|e'_{i}(\hat{\Pi}_{r}-\Pi_{r}\mathcal{P}_{r})\|_{1}=O(\varpi\kappa(\Pi'_{r}\Pi_{r})K\sqrt{\lambda_{1}(\Pi'_{r}\Pi_{r})}),\\
&\|e'_{j}(\hat{\Pi}_{c}-\Pi_{c}\mathcal{P}_{c})\|_{1}=O(\varpi\kappa(\Pi'_{c}\Pi_{c})K\sqrt{\lambda_{1}(\Pi'_{c}\Pi_{c})}).
\end{align*}
\end{thm}
Without Assumption \ref{a1}, which specifies a lower bound on the sparsity parameter $\rho$ in terms of the network size, the theoretical bounds of DiSP's error rates as presented in Theorem \ref{Main} become less informative. Specifically, without this assumption, we are unable to obtain any upper bounds for the quantity $\varpi$. Without such upper bounds, we lack insights into how DiSP performs under the proposed DiMMSB model. Consequently, Assumption \ref{a1} is crucial for studying the theoretical properties of DiSP. Furthermore, similar to Corollary 3.1 in \cite{mao2020estimating}, by considering more conditions, we have the following corollary.
\begin{cor}\label{AddConditions}
Under $DiMMSB(n_{r}, n_{c}, K, P,\Pi_{r}, \Pi_{c})$, when conditions in Lemma \ref{rowwiseerror} hold, suppose $\lambda_{K}(\Pi'_{r}\Pi_{r})=O(\frac{n_{r}}{K})$ and $\lambda_{K}(\Pi'_{c}\Pi_{c})=O(\frac{n_{c}}{K})$, with high probability,
\begin{align*}
&\|e'_{i}(\hat{\Pi}_{r}-\Pi_{r}\mathcal{P}_{r})\|_{1}=O(\frac{K^{2}(\sqrt{C\frac{\mathrm{max}(n_{r},n_{c})}{\mathrm{min}(n_{r},n_{c})}}+\sqrt{\mathrm{log}(n_{r}+n_{c})})}{\sigma_{K}(\tilde{P})\sqrt{\rho
n_{c}}}),\\
&\|e'_{j}(\hat{\Pi}_{c}-\Pi_{c}\mathcal{P}_{c})\|_{1}=O(\frac{K^{2}(\sqrt{C\frac{\mathrm{max}(n_{r},n_{c})}{\mathrm{min}(n_{r},n_{c})}}+\sqrt{\mathrm{log}(n_{r}+n_{c})})}{\sigma_{K}(\tilde{P})\sqrt{\rho n_{r}}}),
\end{align*}
where $C$ is a positive constant. Meanwhile,
\begin{itemize}
\item when
    $C\frac{\mathrm{max}(n_{r},n_{c})}{\mathrm{min}(n_{r},n_{c})}\leq\mathrm{log}(n_{r}+n_{c})$,
    we have
\begin{align*}
\|e'_{i}(\hat{\Pi}_{r}-\Pi_{r}\mathcal{P}_{r})\|_{1}=O(\frac{K^{2}\sqrt{\mathrm{log}(n_{r}+n_{c})}}{\sigma_{K}(\tilde{P})\sqrt{\rho
n_{c}}}),\|e'_{j}(\hat{\Pi}_{c}-\Pi_{c}\mathcal{P}_{c})\|_{1}=O(\frac{K^{2}\sqrt{\mathrm{log}(n_{r}+n_{c})}}{\sigma_{K}(\tilde{P})\sqrt{\rho n_{r}}}).
\end{align*}
\item when $n_{r}=O(n), n_{c}=O(n)$ (i.e.,
    $\frac{n_{r}}{n_{c}}=O(1)$), we have
\begin{align*}
&\|e'_{i}(\hat{\Pi}_{r}-\Pi_{r}\mathcal{P}_{r})\|_{1}=O(\frac{K^{2}}{\sigma_{K}(\tilde{P})}\sqrt{\frac{\mathrm{log}(n)}{\rho n}}),\|e'_{j}(\hat{\Pi}_{c}-\Pi_{c}\mathcal{P}_{c})\|_{1}=O(\frac{K^{2}}{\sigma_{K}(\tilde{P})}\sqrt{\frac{\mathrm{log}(n)}{\rho n}}).
\end{align*}
\end{itemize}
\end{cor}

Under the settings of Corollary \ref{AddConditions},  when $K=O(1)$, to ensure the consistency of estimation, for the case $C\frac{\mathrm{max}(n_{r},n_{c})}{\mathrm{min}(n_{r},n_{c})}\leq\mathrm{log}(n_{r}+n_{c})$, $\sigma_{K}(\tilde{P})$ should shrink slower than $\sqrt{\frac{\mathrm{log}(n_{r}+n_{c})}{\rho \mathrm{min}(n_{r}+n_{c})}}$; Similarly,  for the case $\frac{n_{r}}{n_{c}}=O(1)$, $\sigma_{K}(\tilde{P})$ should shrink slower than $\sqrt{\frac{\mathrm{log}(n)}{\rho n}}$.
\begin{rem}
By Lemma \ref{P4}, we know $\sigma_{K}(\Omega)\geq\rho\sigma_{K}(\tilde{P})\sigma_{K}(\Pi_{r})\sigma_{K}(\Pi_{c})$. To ensure the condition $\sigma_{K}(\Omega)\geq C(\rho (n_{r}+n_{c})\mathrm{log}(n_{r}+n_{c}))^{1/2}$ in lemma \ref{rowwiseerror} hold, we need $\rho\sigma_{K}(\tilde{P})\sigma_{K}(\Pi_{r})\sigma_{K}(\Pi_{c})\geq C(\rho (n_{r}+n_{c})\mathrm{log}(n_{r}+n_{c}))^{1/2}$. Thus
\begin{align}\label{sigmaKPlowerbound}
\sigma_{K}(\tilde{P})\geq C\big(\frac{(n_{r}+n_{c})\mathrm{log}(n_{r}+n_{c})}{\rho \lambda_{K}(\Pi'_{r}\Pi_{r})\lambda_{K}(\Pi'_{c}\Pi_{c})}\big)^{1/2}.
\end{align}

When $K=O(1), \lambda_{K}(\Pi'_{r}\Pi_{r})=O(\frac{n_{r}}{K}),$ and $ \lambda_{K}(\Pi'_{c}\Pi_{c})=O(\frac{n_{c}}{K})$, Eq (\ref{sigmaKPlowerbound}) gives that $\sigma_{K}(\tilde{P})$ should grow faster than $\mathrm{log}^{1/2}(n_{r}+n_{c})/(\rho \mathrm{min}(n_{r},n_{c}))^{1/2}$, which matches with the consistency requirement on $\sigma_{K}(\tilde{P})$ obtained from Corollary \ref{AddConditions}.
\end{rem}
\begin{rem}
When DiMMSB degenerates to MMSB, following similar analysis after Lemma \ref{rowwiseerror}, we see that the error bound in Theorem 2.2 in \cite{MixedSCORE} is consistent with ours. Meanwhile, by adopting the separation condition and the sharp threshold criterion established in \citep{qing2022useful}, we can affirm that the separation condition and sharp threshold derived from our theoretical findings align with the classical results presented in \citep{erdos1960evolution, mcsherry2001spectral, lei2015consistency, abbe2018community}. The above analysis guarantees the optimality of our theoretical results.
\end{rem}

Similarly, under the settings of Corollary \ref{AddConditions}, for the case $C\frac{\mathrm{max}(n_{r},n_{c})}{\mathrm{min}(n_{r},n_{c})}\leq\mathrm{log}(n_{r}+n_{c})$, when $\sigma_{K}(\tilde{P})$ is a constant, the theoretical error bounds are $O(K^{2}\sqrt{\frac{\mathrm{log}(n_{r}+n_{c})}{\rho\mathrm{min}(n_{r},n_{c})}})$. Therefore, for consistent estimation of DiSP, $K$ should grow slower than $(\frac{\rho \mathrm{min}(n_{r}+n_{c})}{\mathrm{log}(n_{r}+n_{c})})^{1/4}$. Similarly, under the settings of Corollary \ref{AddConditions}, for the case $\frac{n_{r}}{n_{c}}=O(1)$, when $\sigma_{K}(\tilde{P})$ is a constant, the upper bounds of the error rates are $O(K^{2}\sqrt{\frac{\mathrm{log}(n)}{\rho n}})$. For a consistent estimation, $K$ should grow slower than $(\frac{\rho n}{\mathrm{log}(n)})^{1/4}$.

Consider the balanced directed mixed membership network (i.e., $\lambda_{K}(\Pi'_{r}\Pi_{r})=O(\frac{n_{r}}{K}), \lambda_{K}(\Pi'_{c}\Pi_{c})=O(\frac{n_{c}}{K})$ and $n_{r}=O(n), n_{c}=O(n)$) in Corollary \ref{AddConditions}, we further assume that $\tilde{P}=\beta I_{K}+(1-\beta)\textbf{1}_{K}\textbf{1}'_{K}$ (where $\textbf{1}_{K}$ is a $K\times 1$ vector with all entries being ones.) for $0<\beta<1$ when $K=O(1)$ and call such directed network as standard directed mixed membership network. To obtain a consistent estimate, $\beta$ must reduce slower than $\sqrt{\frac{\mathrm{log}(n)}{\rho n}}$ since $\sigma_{K}(\tilde{P})=\beta$. Let $P_{\mathrm{max}}=\max_{k,l}P(k,l), P_{\mathrm{min}}=\mathrm{min}_{k,l}P(k,l)$. Since $P=\rho \tilde{P}$, we have $P_{\mathrm{max}}-P_{\mathrm{min}}=\rho\beta$ (the probability gap) must reduce slower than $\sqrt{\frac{\rho \mathrm{log}(n)}{n}}$ and $\frac{P_{\mathrm{max}}-P_{\mathrm{min}}}{\sqrt{P_{\mathrm{max}}}}=\beta\sqrt{\rho}$ (the relative edge probability gap) must reduce slower than $\sqrt{\frac{\mathrm{log}(n)}{n}}$. Especially, when we consider the sparest network $\rho n=\mathrm{log}(n)$ satisfying Assumption (\ref{a1}), the probability gap should shrink slower than $\frac{\mathrm{log}(n)}{n}$.

\section{Simulations}\label{sec5}
This section presents simulation studies to evaluate the effectiveness of DiSP. To quantify performance, we employ three metrics defined as follows:
\begin{itemize}
  \item DiMHamm$=\frac{\mathrm{min}_{\mathcal{P}\in S}\|\hat{\Pi}_{r}\mathcal{P}-\Pi_{r}\|_{1}+\mathrm{min}_{\mathcal{P}\in S}\|\hat{\Pi}_{c}\mathcal{P}-\Pi_{c}\|_{1}}{n_{r}+n_{c}}$,
  \item row-MHamm$=\frac{\mathrm{min}_{\mathcal{P}\in S}\|\hat{\Pi}_{r}\mathcal{P}-\Pi_{r}\|_{1}}{n_{r}}$,
  \item column-MHamm$=\frac{\mathrm{min}_{\mathcal{P}\in S}\|\hat{\Pi}_{c}\mathcal{P}-\Pi_{c}\|_{1}}{n_{c}}$,
\end{itemize}
where $S$ represents the set that contains all $K\times K$ permutation matrices. We additionally account for label permutations, because the error metric ought to be invariant to the specific labeling of groups. DiMHamm is applied for evaluating DiSP's performance on both row nodes and column nodes. Similar explanations hold for row-MHamm and column-MHamm. Meanwhile, in the following experiments 1-3, we compare DiSP with the variational expectation-maximization (vEM for short) algorithm \cite{airoldi2013multi} for their two-way stochastic blockmodels with Bernoulli distribution. By Table 1 in \cite{airoldi2013multi}, we see that vEM under the two input Dirichlet parameters $\alpha=0.05, \beta=0.05$ (by \cite{airoldi2013multi}'s notation) generally performs better than that under $\alpha=\beta=0.2$. Therefore, in our simulations, we set the two Dirichlet parameters $\alpha$ and $\beta$ of vEM as $0.05$. We also compare DiSP with the directed versions of Mixed-SCORE \citep{MixedSCORE}, GeoNMF \citep{GeoNMF}, SVM-cone-DCMMSB (SVM-cD) \citep{MaoSVM}. These three methods are originally studied in undirected networks. To adapt them to directed networks, we make the following modifications: using $AA'$ and $A'A$ to replace their original input adjacency matrix $A$ to estimate $\Pi_{r}$ and $\Pi_{c}$, respectively.

For the first three experiments, unless specified, all model parameters are set as follows. For row nodes, $n_{r}=60$ and $K=3$. Let every row community have $n_{r,0}$ pure nodes. We let the first $Kn_{r,0}$ row nodes $\{1,2, \ldots, Kn_{r,0}\}$ be pure and all rest row nodes be mixed. Unless specified, consider four different mixed memberships $(1/3,1/3,1/3), (0.4, 0.4, 0.2),\\ (0.4, 0.2, 0.4)$, and $(0.2, 0.4, 0.4)$, each with $\frac{n_{r}-Kn_{r,0}}{4}$ number of nodes when $K=3$. For column nodes, set $n_{c}=80$. Each column block is assumed to have  $n_{c,0}$ pure nodes. Let the top $Kn_{c,0}$ column nodes $\{1,2, \ldots, Kn_{c,0}\}$ be pure, while the remaining columns $\{Kn_{c,0}+1, Kn_{c,0}+2,\ldots, n_{c}\}$ are treated as mixed. Column mixed membership configurations mirror those of their row counterparts. Whenever $n_{r,0}$ equals $n_{c,0}$, we write $n_{0}$ for this common value (i.e., $n_{0}=n_{r,0}=n_{c,0}$) to simplify notation. The probability matrix $P$ is set independently for each experiment.

Once $P$, $\Pi_r$, and $\Pi_c$ are in hand, each simulation experiment follows the following steps.

(a) Construct the population signal matrix $\Omega = \Pi_r P \Pi_c'$.

(b) Form the raw observed matrix $\tilde{A} = \Omega + W$, where $W$ is a noise matrix of dimensions $n_r \times n_c$ with each entry $W(i,j)$ being an independent centered Bernoulli random variable with success probability $\Omega(i,j)$. 

(c) Identify row nodes with zero total outgoing edges: $\tilde{S}_r = \{ i \mid \sum_{j=1}^{n_c} \tilde{A}(i,j) = 0 \}$. Likewise, identify column nodes with zero total incoming edges: $\tilde{S}_c = \{ j \mid \sum_{i=1}^{n_r} \tilde{A}(i,j) = 0 \}$. The final adjacency matrix $A$ is obtained by deleting from $\tilde{A}$ all rows indexed by $\tilde{S}_r$ and all columns indexed by $\tilde{S}_c$. The membership matrices are updated accordingly: remove the rows in $\tilde{S}_r$ from $\Pi_r$, and remove the columns in $\tilde{S}_c$ from $\Pi_c$.

(d) Run each competing method on $A$, and record each metric as well as the computational time.

(e) Repeat steps (b) through (d) a total of 50 repstitions. For each repetition, compute the error metrics and runtime; then report the averages (i.e., mean DiMHamm, mean row-MHamm, mean column-MHamm, and mean runtime) over the 50 repetitions.

In our experiments, the number of rows of $A$ and the number of columns of $A$ typically differ only slightly from $n_r$ and $n_c$, respectively. Hence, we omit the precise row and column counts of $A$ in our reporting.
\begin{figure}
\centering
\subfigure[Changing $n_{0}$:
DiMHamm]{\includegraphics[width=0.32\textwidth]{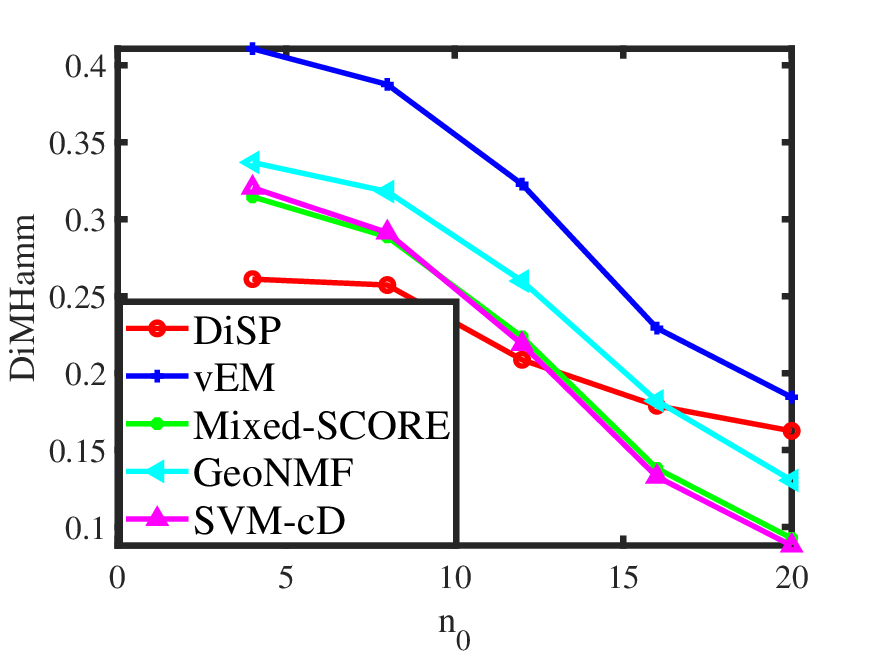}}
\subfigure[Changing $n_{0}$:
row-MHamm]{\includegraphics[width=0.32\textwidth]{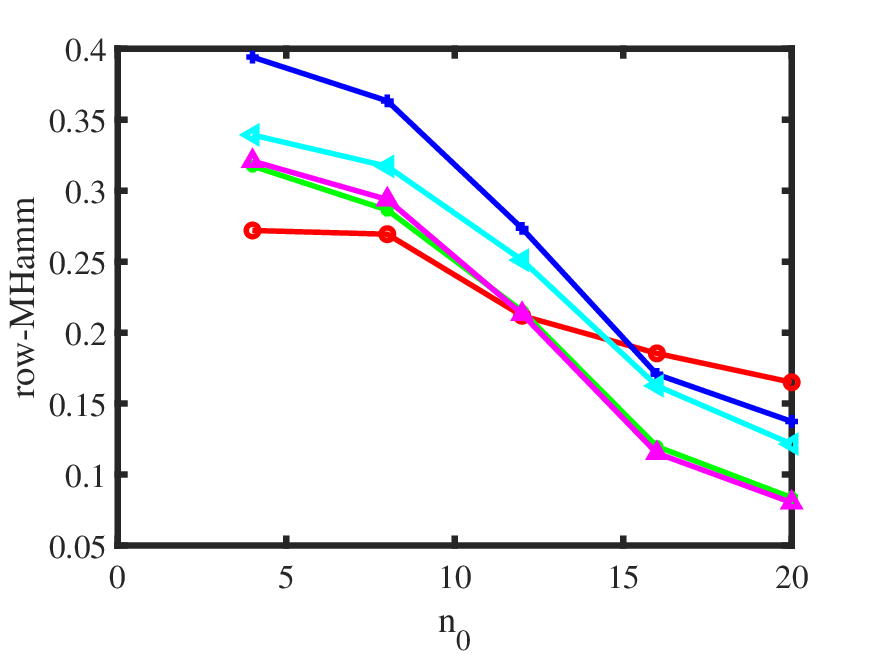}}
\subfigure[Changing $n_{0}$: column-MHamm]{\includegraphics[width=0.32\textwidth]{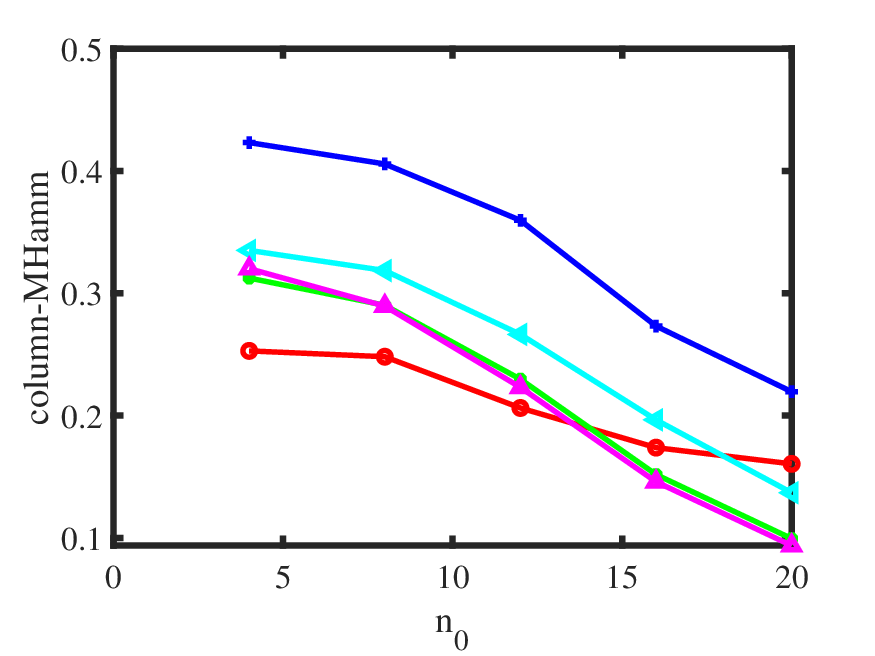}}
\subfigure[Changing $\rho$: DiMHamm]{\includegraphics[width=0.32\textwidth]{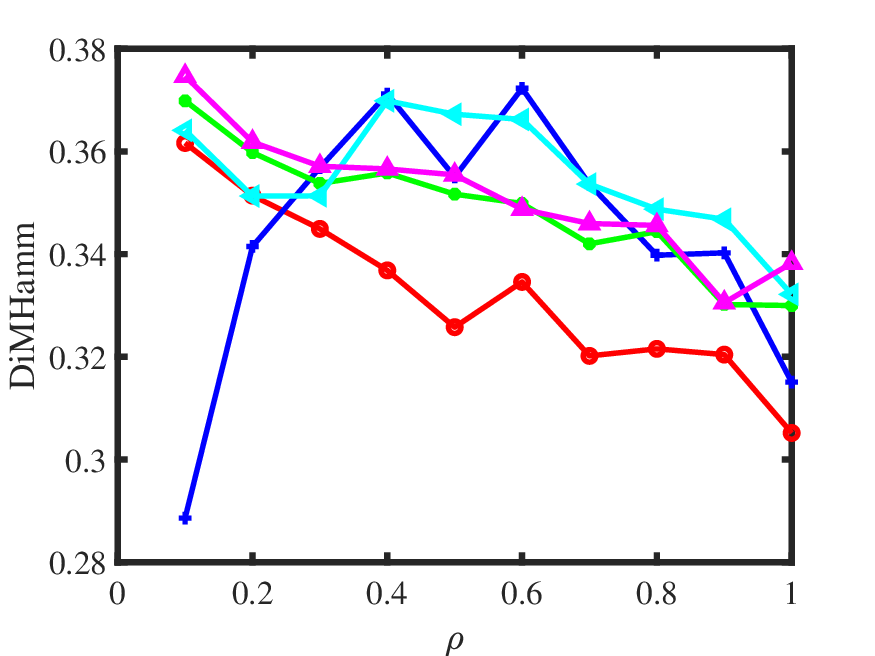}}
\subfigure[Changing $\rho$:
row-MHamm]{\includegraphics[width=0.32\textwidth]{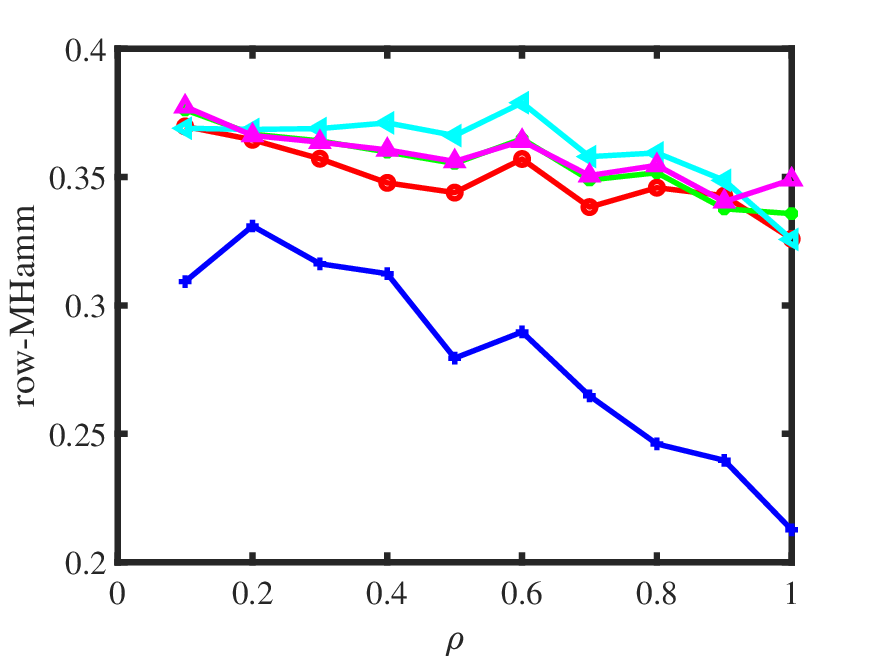}}
\subfigure[Changing $\rho$:
column-MHamm]{\includegraphics[width=0.32\textwidth]{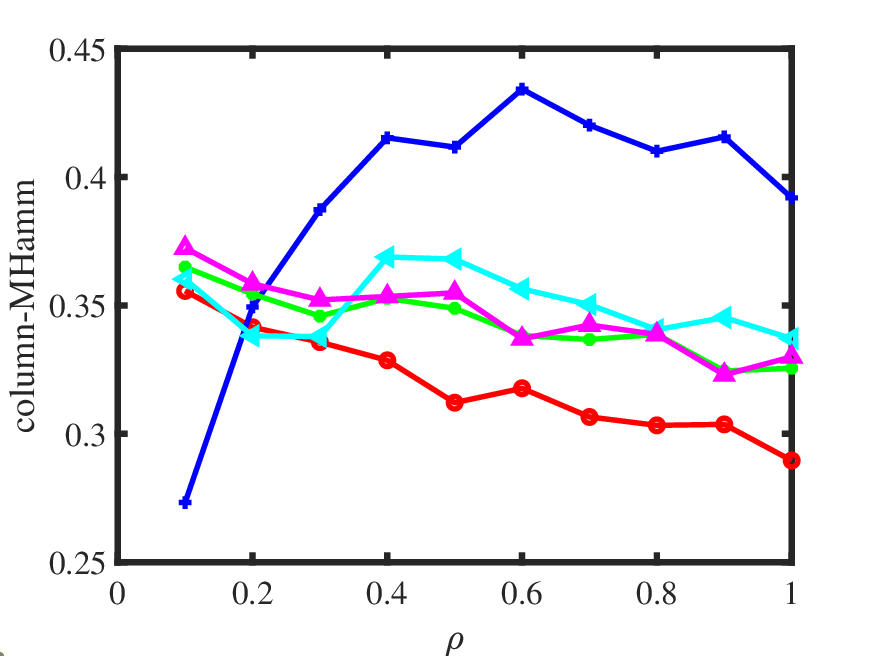}}
\subfigure[Changing $\beta$:
DiMHamm]{\includegraphics[width=0.32\textwidth]{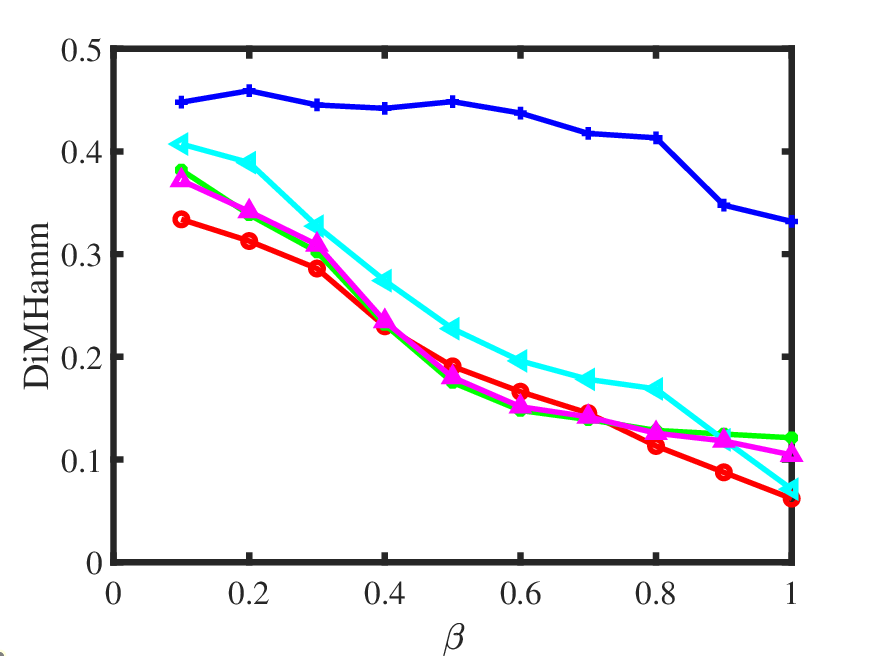}}
\subfigure[Changing $\beta$:
row-MHamm]{\includegraphics[width=0.32\textwidth]{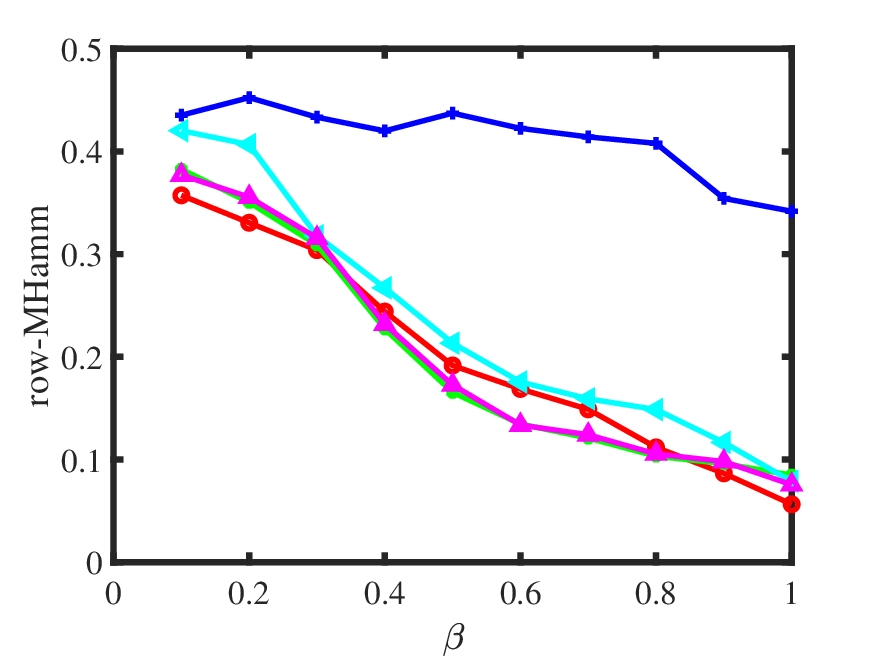}}
\subfigure[Changing $\beta$:
column-MHamm]{\includegraphics[width=0.32\textwidth]{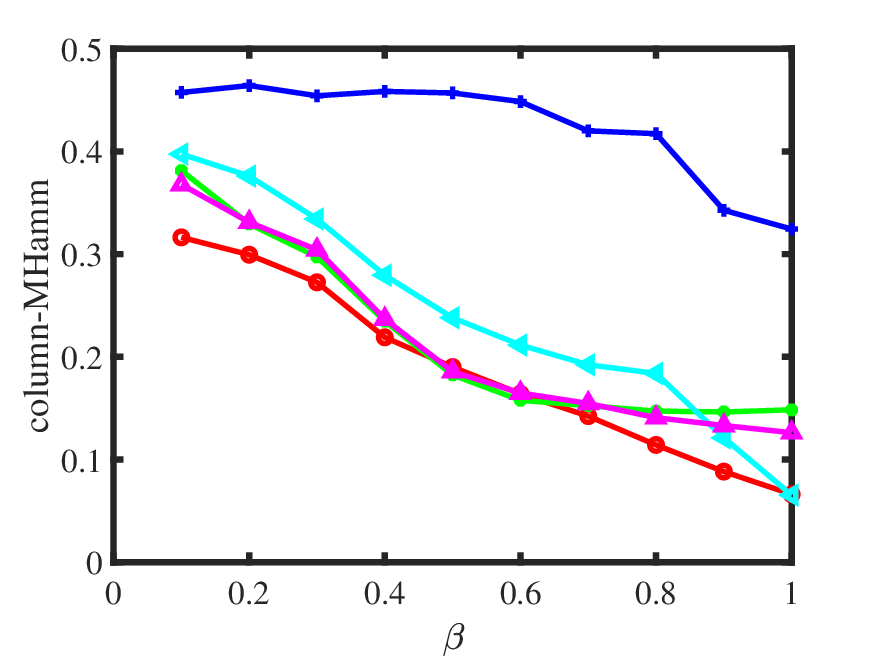}}
\subfigure[Changing $n_{0}$:
run-time]{\includegraphics[width=0.32\textwidth]{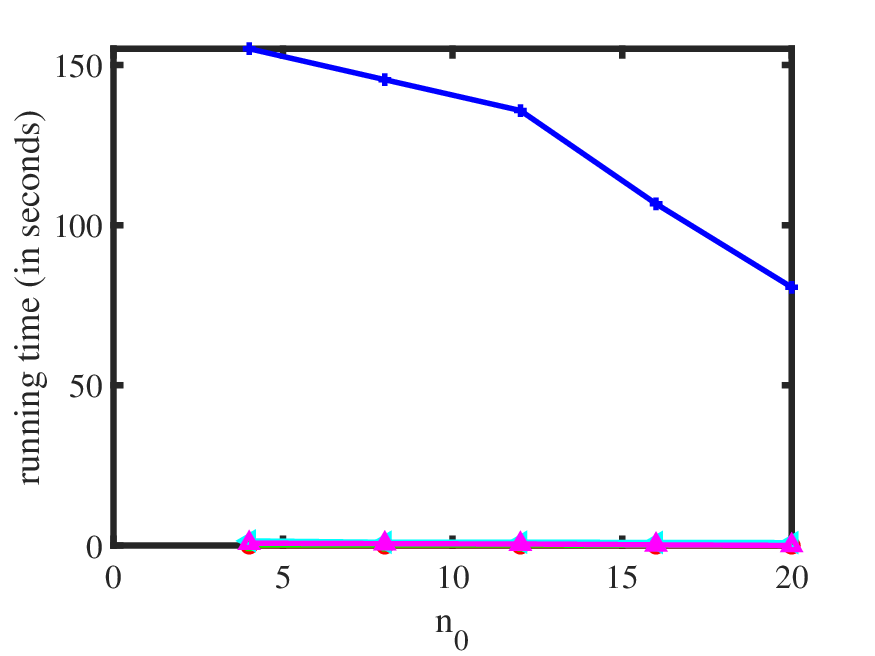}}
\subfigure[Changing $\rho$:
run-time]{\includegraphics[width=0.32\textwidth]{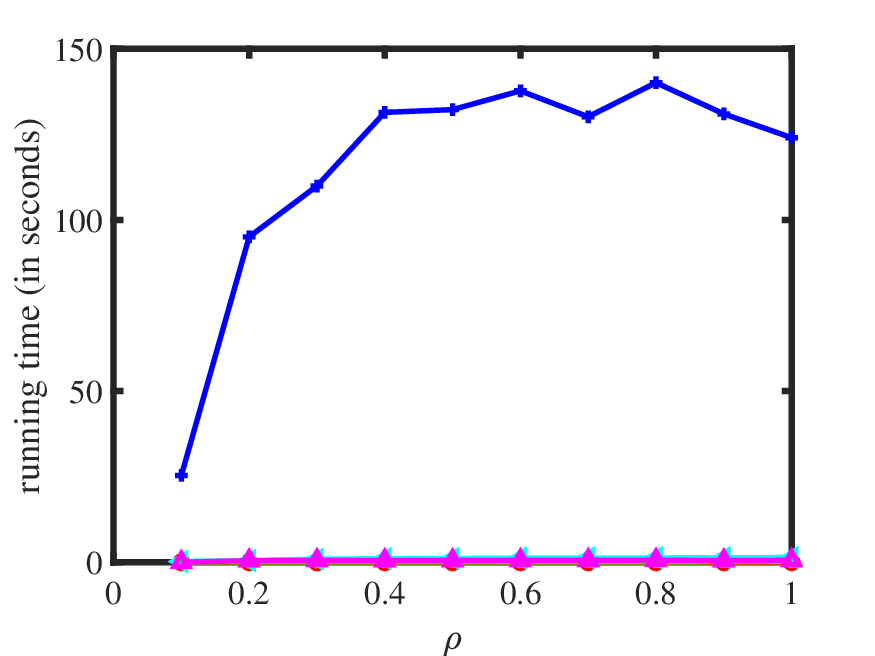}}
\subfigure[Changing $\beta$:
run-time]{\includegraphics[width=0.32\textwidth]{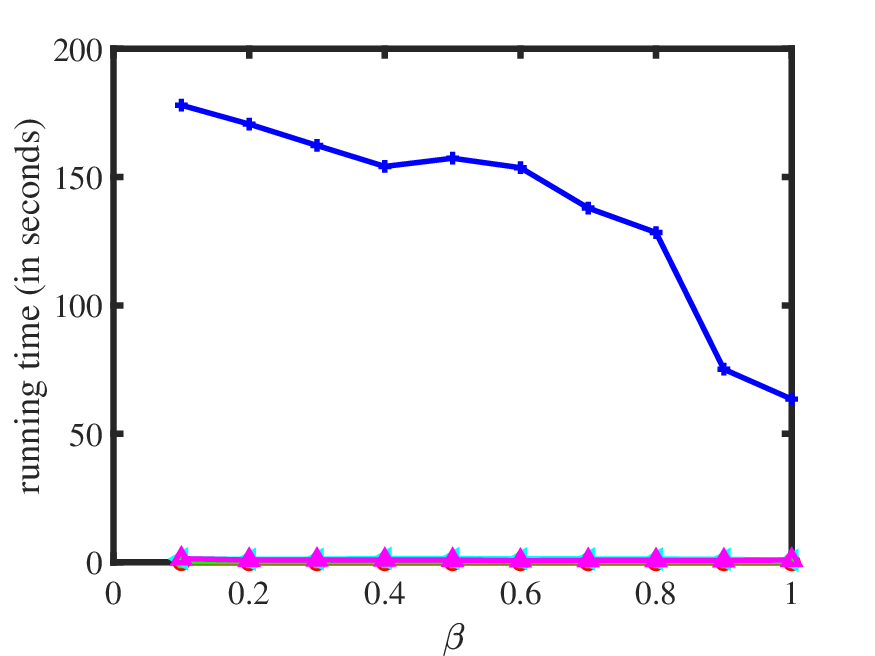}}
\caption{Numerical results of Experiments 1-3.}
\label{EX} 
\end{figure}

\texttt{Experiment 1: Changing $n_{0}$.} The probability matrix in this experiment is set as
 \[P=\begin{bmatrix}
 0.8&0.1&0.3\\
 0.2&0.9&0.4\\
 0.5&0.2&0.9\\
 \end{bmatrix}.
 \]

Let $n_{0}$ range in $\{4, 8, 12, 16,20\}$. 

Increasing $n_{0}$ corresponds to a higher proportion of pure nodes among both row and column communities. Panels (a)–(c) of Fig.~\ref{EX} present the retults. These three panels reveal that all three error metrics exhibit similar trends and the proportion of pure nodes affects how well these approaches perform: each approach yields lower error rates for more pure nodes. The plots of run-time are shown in panel (j) of Fig.~\ref{EX}. Meanwhile, codes for all numerical results in this paper are written in MATLAB R2021b. The total run-time of Experiment 1 for vEM is roughly 8 hours, and it is roughly 1.5 seconds for DiSP, Mixed-SCORE, GeoNMF, and SVM-cD. DiSP, Mixed-SCORE, GeoNMF, and SVM-cD outperform  vEM on both error rates and run-time. Furthermore, when $n_{0}\leq12$, our DiSP performs best, while it performs slightly poorer than Mixed-SCORE, GeoNMF, and SVM-cD when $n_{0}>12$ for this experiment.

\texttt{Experiment 2: Changing $\rho$.}
Let the sparsity parameter $\rho\in\{0.1,0.2,\ldots,1\}$. Set the probability matrix to be
 \[P=\rho\begin{bmatrix}
 1&0.4&0.4\\
 0.6&1&1\\
 0.2&0.2&0.4\\
 \end{bmatrix}.
 \]

A larger $\rho$ indicates a denser simulated network. Here, $P$ is set much differently from that in Experiment 1, because we aim to emphasize that DiMMSB doesn't have strict constraints on $P$ as long as $\mathrm{rank}(P)=K$ and all elements of $P$ are in $[0,1]$. This experiment's results can be found in panels (d), (e), and (f) of Fig.~\ref{EX} with panel (k) recording run-time. Meanwhile, the total run-time of Experiment 2 for vEM is roughly 16 hours, and it is roughly 3.44 seconds for DiSP. From these results, we see that DiSP, Mixed-SCORE, and SVM-cD outperform vEM on DiMHamm, column-MHamm, and run-time, while vEM performs best on row-MHamm.

\texttt{Experiment 3: Changing $\beta$.}
Let $\beta\in\{0.1,0.2,\ldots,1\}$. The probability matrix in this experiment is set as
 \[P=\begin{bmatrix}
 1&1-\beta&1-\beta\\
 1-\beta&1&1-\beta\\
 1-\beta&1-\beta&1\\
 \end{bmatrix}.
 \]

Since $\sigma_{K}(P)=\beta$, increasing $\beta$ reduces error rates in the analysis for the balanced directed mixed membership network.  Panels (g), (h), and (i) in Fig.~\ref{EX} display simulation results of this experiment, and panel (l) records run-time. In this experiment, the three error measures exhibit very little difference from one another. Meanwhile, the total run-time of Experiment 3 for vEM is roughly 16 hours, and it is roughly 2.9 seconds for DiSP. We see that DiSP, Mixed-SCORE, GeoNMF, and SVM-cD outperform vEM on both error rates and run-time. Furthermore, DiSP, Mixed-SCORE, and SVM-cD perform similarly and outperform GeoNMF slightly in this experiment.
\begin{rem}\label{SimulatedAvisual}
For visualization purposes, the bi-adjacency matrix $A$ generated under DiMMSB is displayed. Set $n_{r}=24, n_{c}=30, K=2$, and the probability matrix as
\(\begin{bmatrix}
    0.8&0.05\\
    0.1&0.7\\
\end{bmatrix}.\) For $1\leq i\leq 8$, we set $\Pi_{r}(i,1)=1$; for $9\leq i\leq16$, we set $\Pi_{r}(i,2)=1$; for $17\leq i\leq 24$, we set $\Pi_{r}(i,:)=[0.7~~0.3]$ (i.e., there are 16 pure row nodes and 8 mixed row nodes). For column nodes, when $1\leq i\leq 8$, let $\Pi_{c}(i,1)=1$; when $9\leq i\leq16$, let $\Pi_{c}(i,2)=1$; when $17\leq i\leq 30$ , let $\Pi_{c}(i,:)=[0.7~~0.3]$ (i.e., there are 16 pure column nodes and 14 mixed column nodes). Based on the aforementioned parameter choices, Fig.~\ref{DiMMSBReamrk3} depicts two randomly generated adjacency matrices, along with the corresponding error rates and computational time of DiSP. Here, because $A$ is provided in Fig.~\ref{DiMMSBReamrk3}, and $\Pi_{r}, \Pi_{c}$ and $K$ are known. readers can apply DiSP to $A$ in Fig.~\ref{DiMMSBReamrk3} to check the effectiveness of the proposed algorithm.
\begin{figure}
\centering
\subfigure[]{\includegraphics[width=0.49\textwidth]{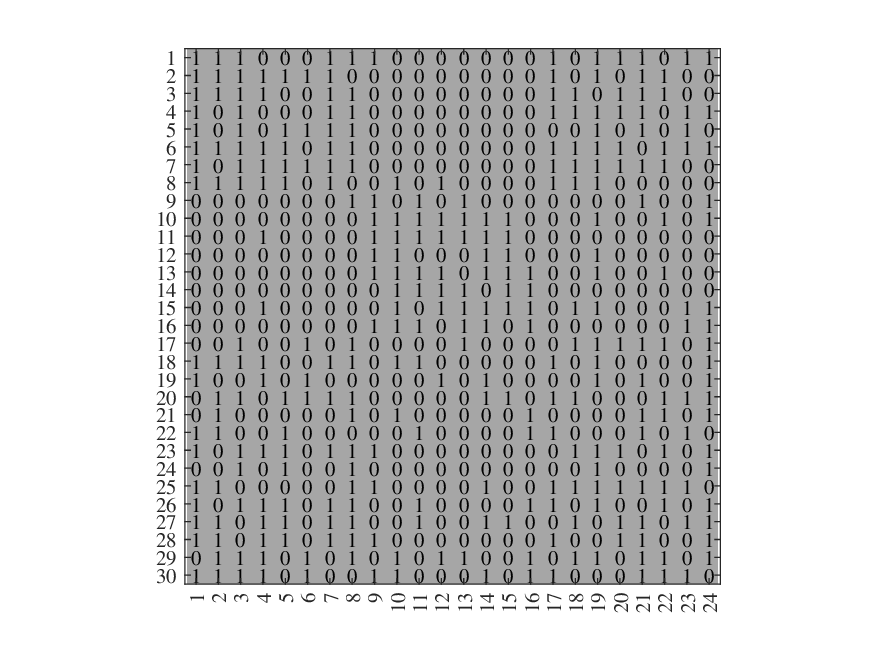}}
\subfigure[]{\includegraphics[width=0.49\textwidth]{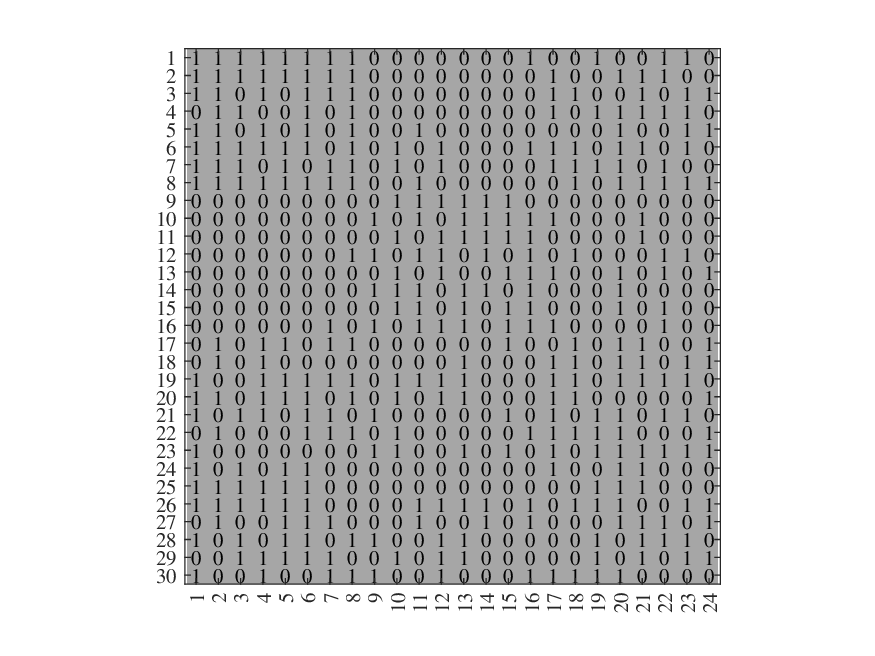}}
\caption{For $A$ in panel (a),$\mathrm{~DiMHamm},
 \mathrm{~row-MHamm}$,$\mathrm{~column-MHamm}$, and run-time for DiSP are 0.0948, 0.1070, 0.0849, and 0.0021, respectively. For $A$ in panel (b),$\mathrm{~DiMHamm},
 \mathrm{~row-MHamm}$,$\mathrm{~column-MHamm}$, and run-time for DiSP are 0.0778, 0.0643, 0.0886, and 0.0020 seconds, respectively. x-axis: row nodes; y-axis: column nodes.}
\label{DiMMSBReamrk3} 
\end{figure}
\end{rem}

\begin{rem}\label{SimulatedNetvisual}
To illustrate a concrete instance of a directed network generated from the DiMMSB model, we plot one realization in Figure~\ref{NetSimulated}. The simulation parameters are as follows: $n_r = n_c = 24$, $K = 2$, and the probability matrix is set to
\(P = 0.8 \begin{bmatrix} 1 & 0.1 \\ 0.2 & 0.6 \end{bmatrix}.\) For the row nodes, we assign pure memberships to the first 16 indices: $\Pi_r(i,1)=1$ for $1 \le i \le 8$ and $\Pi_r(i,2)=1$ for $9 \le i \le 16$. The remaining eight row nodes ($17 \le i \le 24$) are mixed, with membership vector $[0.7, 0.3]$. Hence, the row side contains 16 pure nodes and 8 mixed nodes. For the column nodes, the first 20 nodes are pure: $\Pi_c(i,1)=1$ for $1 \le i \le 10$ and $\Pi_c(i,2)=1$ for $11 \le i \le 20$. The last four column nodes ($21 \le i \le 24$) have the mixed membership $[0.7, 0.3]$. Thus, there are 20 pure and 4 mixed column nodes. From this configuration, a single random adjacency matrix $A$ is generated and displayed in panel (a) of Figure~\ref{NetSimulated}. Panels (b) and (c) of the same figure respectively illustrate the sending patterns (rows) and the receiving patterns (columns) of the simulated directed network.
\begin{figure}[htbp]
\centering
\subfigure[Adjacency matrix]{\includegraphics[width=0.325\textwidth]{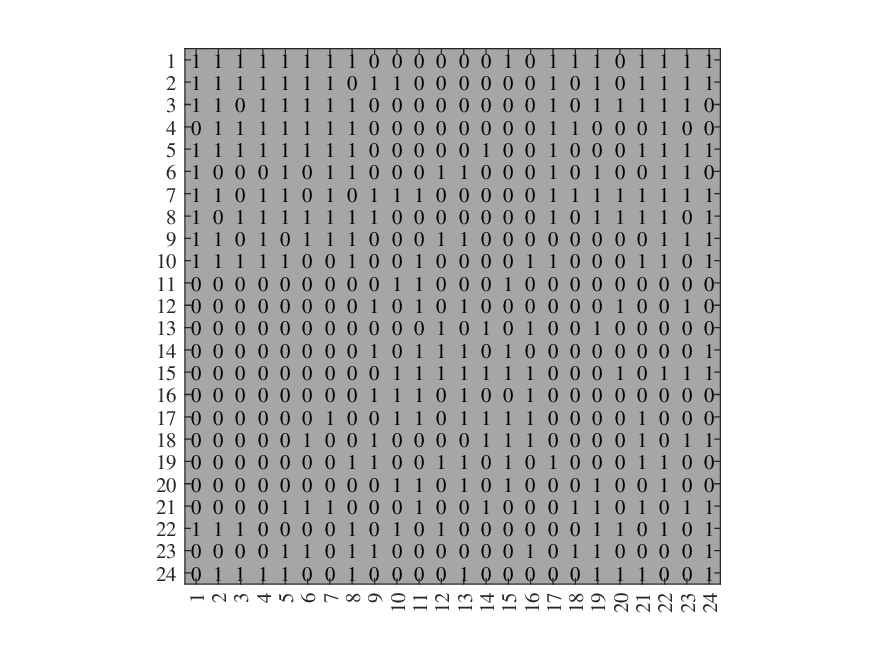}}
\subfigure[Sending clusters]{\includegraphics[width=0.325\textwidth]{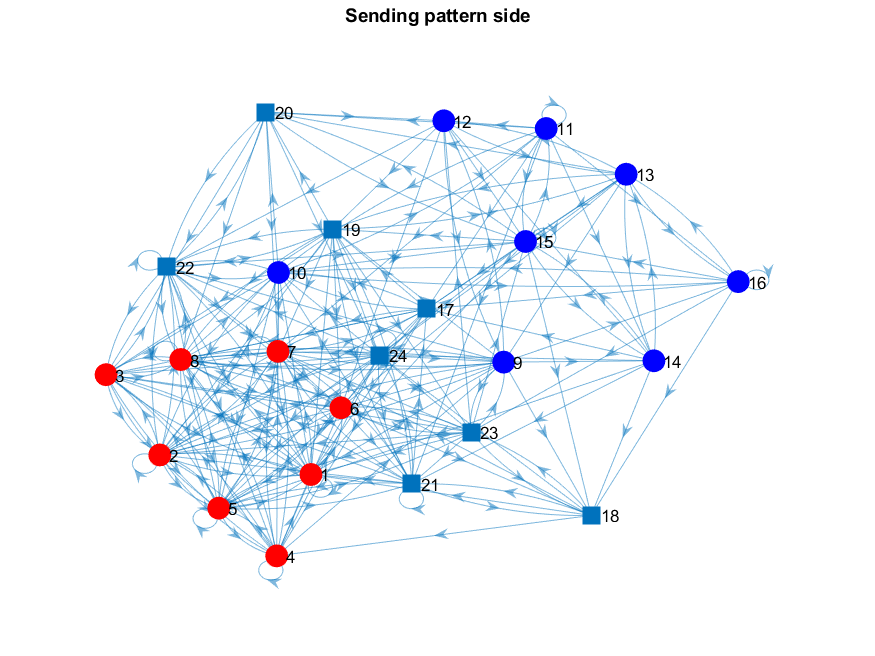}}
\subfigure[Receiving clusters]{\includegraphics[width=0.325\textwidth]{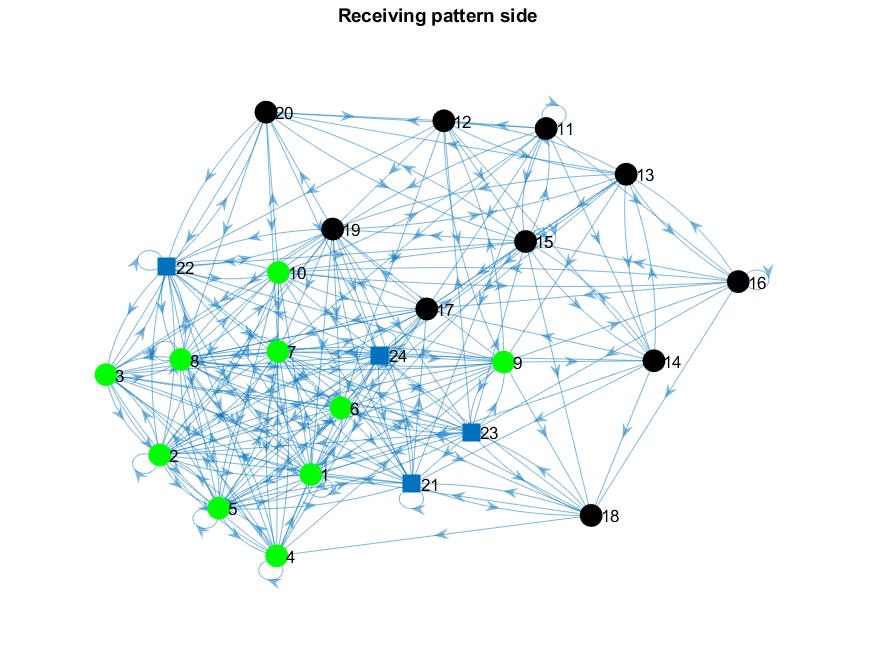}}
\caption{Illustration of a simulated directed network generated under DiMMSB. Panels (a), (b), and (c) show $A$, the sending clusters, and the receiving clusters of this simulated directed network, respectively. Here, the DiMHamm, row-MHamm, column-MHamm, and DiSP runtime are 0.0796, 0.0786, 0.0806, and 0.0021 seconds, respectively. Turning to the last two panels, colors represent different clusters, squares highlight mixed nodes, and the sending/receiving cluster assignments are derived from $\Pi_r$ and $\Pi_c$ as introduced in Remark~\ref{SimulatedNetvisual}. The horizontal axis corresponds to row nodes, the vertical axis to column nodes.}
\label{NetSimulated}
\end{figure}
\end{rem}

In Experiments 1-3, we mainly investigate the performance of DiSP by comparing it with vEM, Mixed-SCORE, GeoNMF, and SVM-cD on small directed mixed membership networks. The numerical results show that DiSP, Mixed-SCORE, GeoNMF, and SVM-cD perform much better than vEM on error rates, and they are much faster than vEM. However, the error rates are always quite large in Experiments 1-3 because the directed mixed membership network with 60 row nodes and 80 column nodes is too small, and a few edges can be generated for such small directed mixed membership networks under the settings in Experiments 1-3. In the next four experiments, we investigate the performances of DiSP, Mixed-SCORE, GeoNMF, and SVM-cD on some larger (compared with those under Experiments 1-3) directed mixed membership networks. Because the run-time for vEM is too large for large networks, we do not compare them with vEM in the next four experiments.
\begin{figure}
\centering
\subfigure[Changing $n_{0}$:
DiMHamm]{\includegraphics[width=0.32\textwidth]{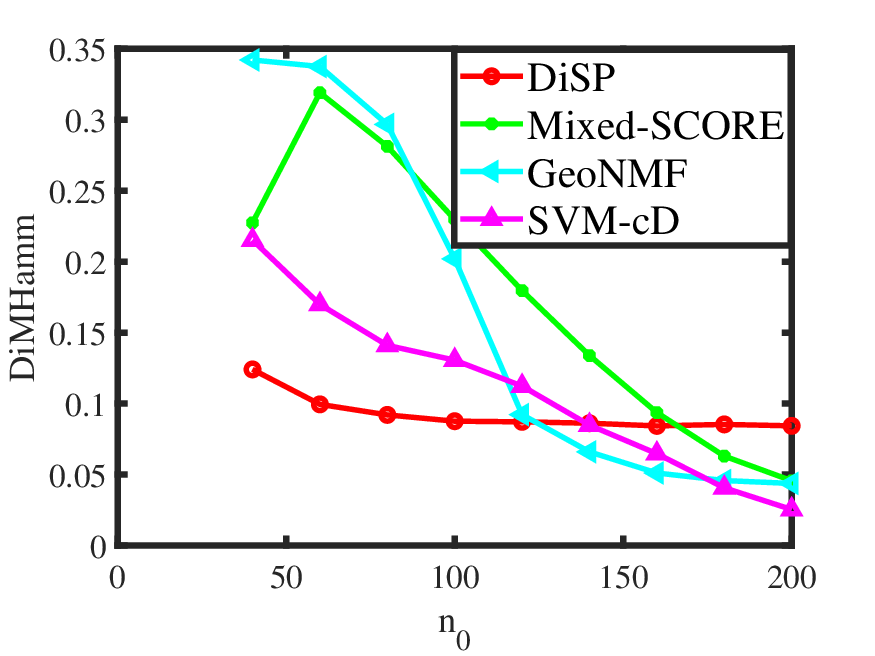}}
\subfigure[Changing $n_{0}$:
row-MHamm]{\includegraphics[width=0.32\textwidth]{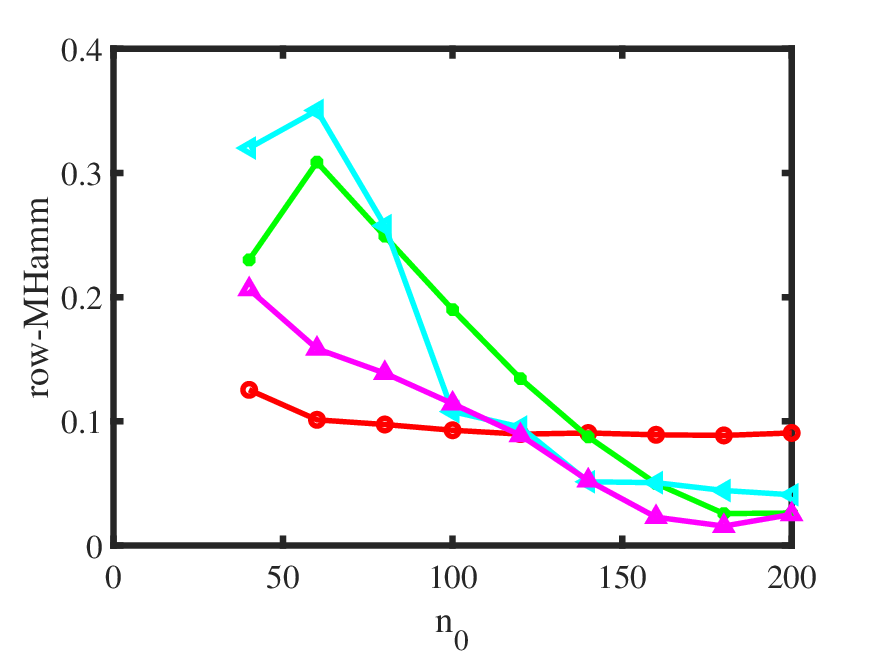}}
\subfigure[Changing $n_{0}$: column-MHamm]{\includegraphics[width=0.32\textwidth]{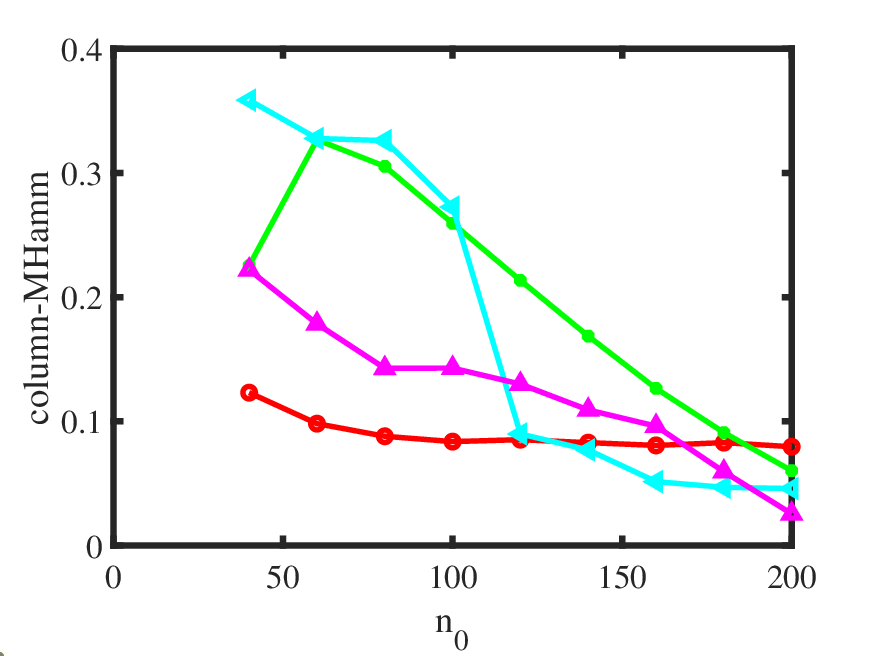}}
\subfigure[Changing $\rho$: DiMHamm]{\includegraphics[width=0.32\textwidth]{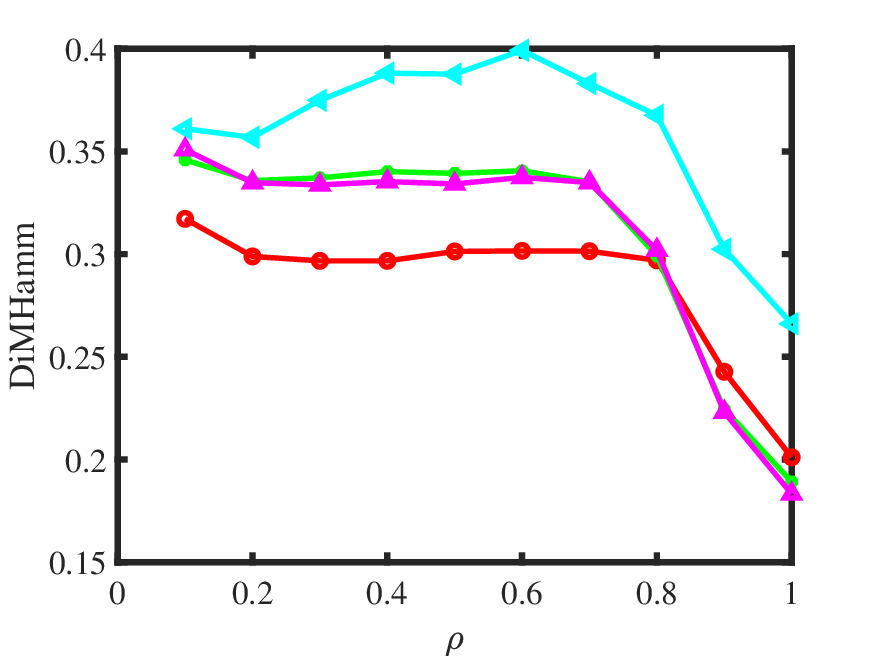}}
\subfigure[Changing $\rho$:
row-MHamm]{\includegraphics[width=0.32\textwidth]{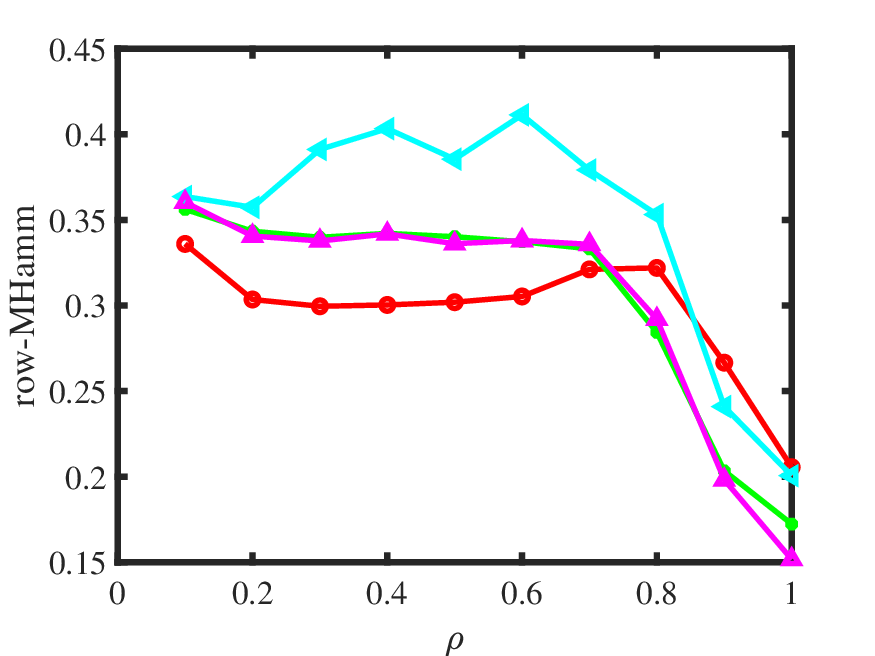}}
\subfigure[Changing $\rho$:
column-MHamm]{\includegraphics[width=0.32\textwidth]{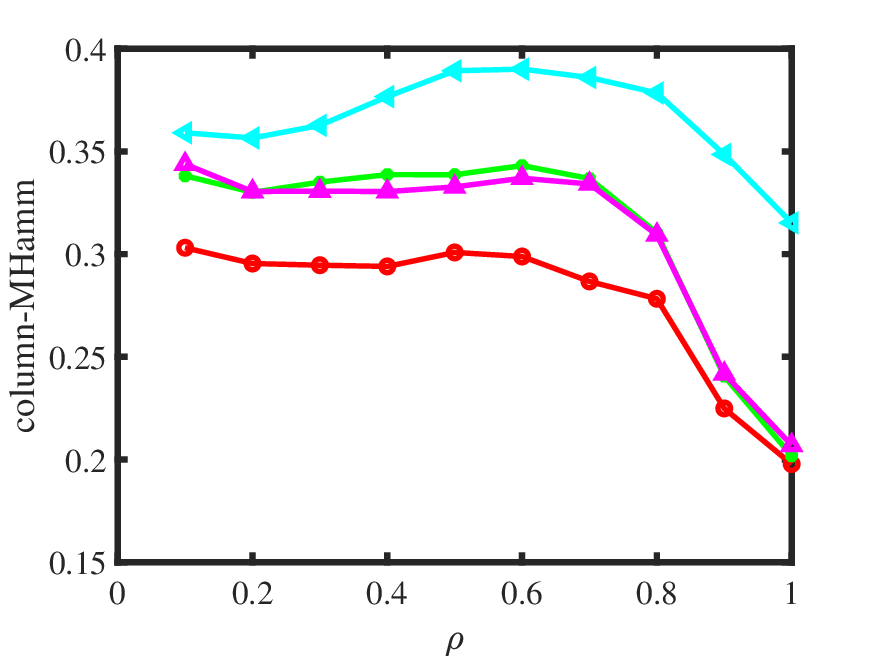}}
\subfigure[Changing $\beta$:
DiMHamm]{\includegraphics[width=0.32\textwidth]{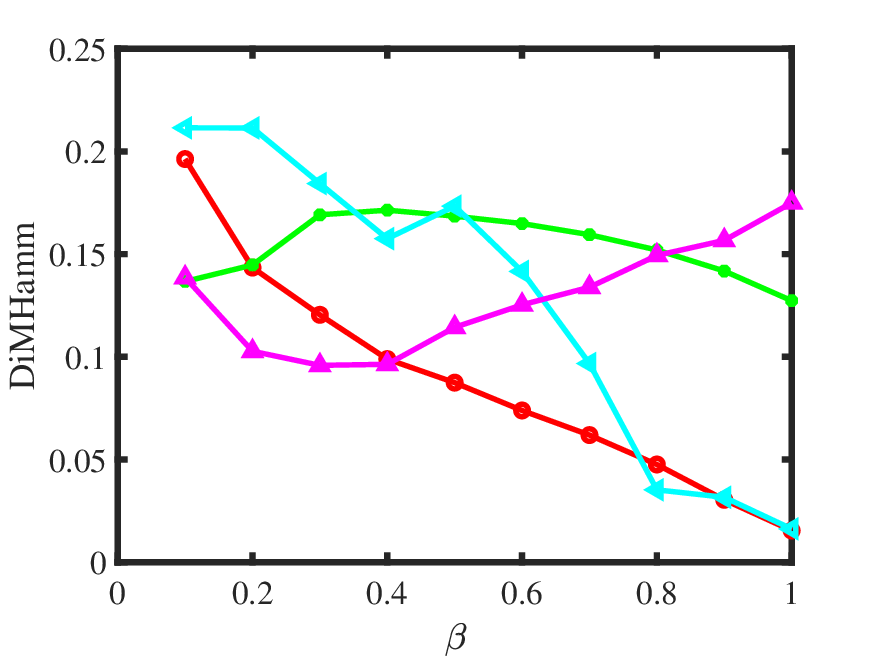}}
\subfigure[Changing $\beta$:
row-MHamm]{\includegraphics[width=0.32\textwidth]{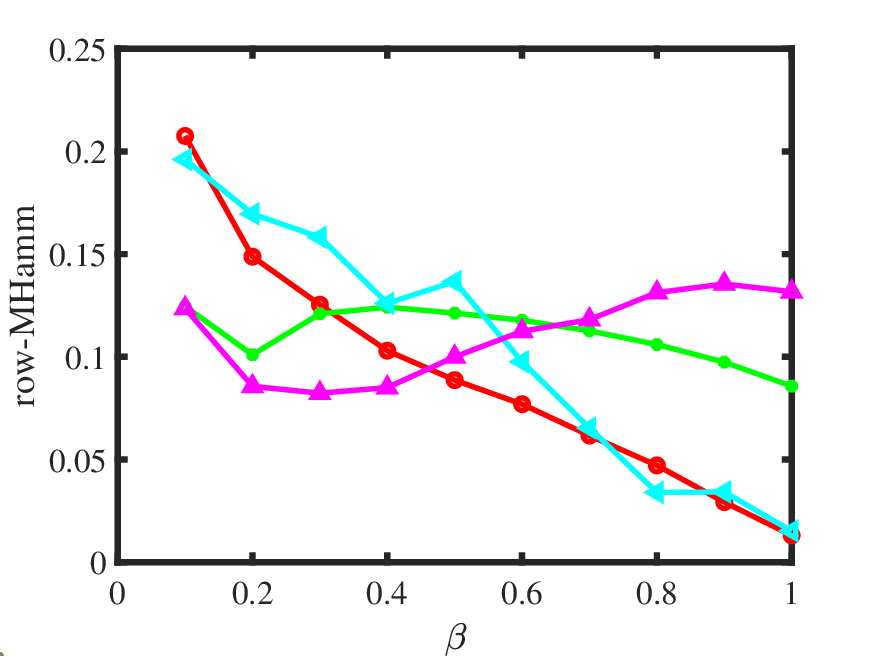}}
\subfigure[Changing $\beta$:
column-MHamm]{\includegraphics[width=0.32\textwidth]{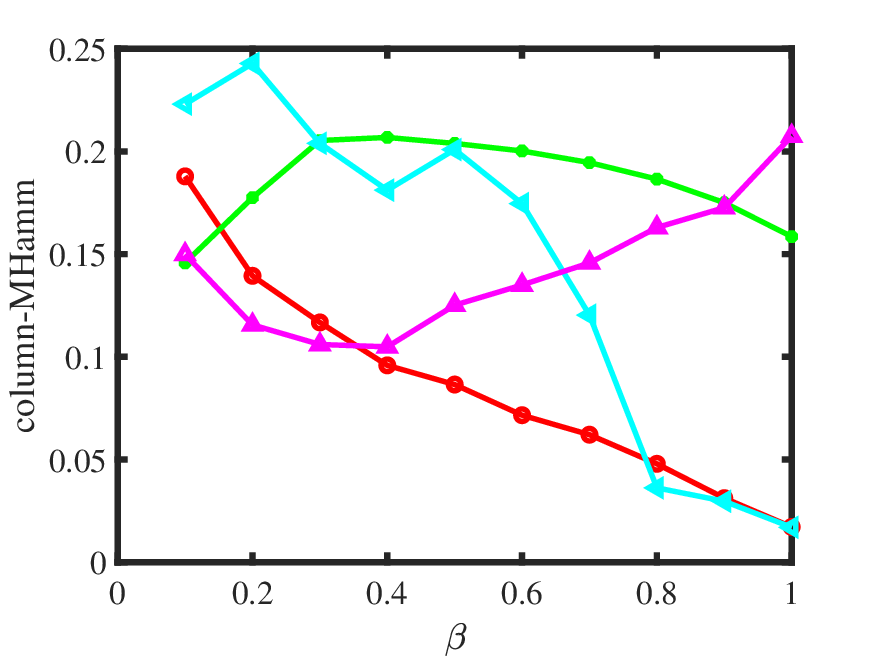}}
\subfigure[Changing $K$:
DiMHamm]{\includegraphics[width=0.32\textwidth]{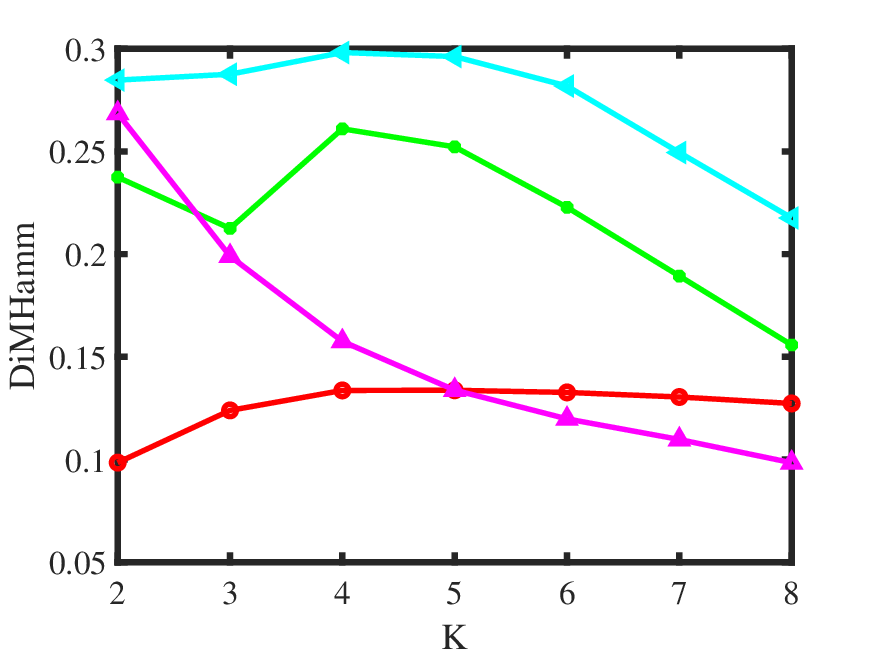}}
\subfigure[Changing $K$:
row-MHamm]{\includegraphics[width=0.32\textwidth]{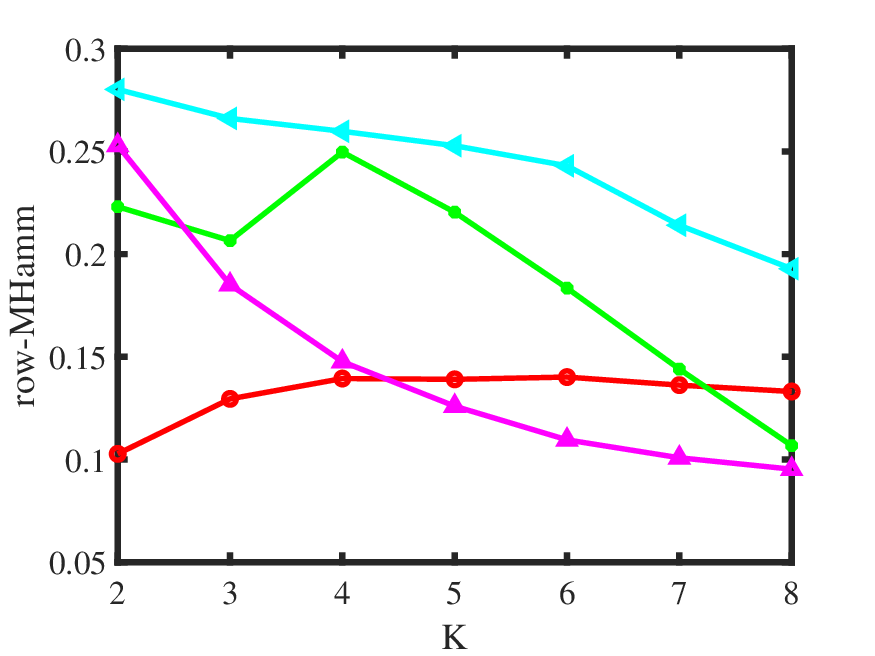}}
\subfigure[Changing $K$:
column-MHamm]{\includegraphics[width=0.32\textwidth]{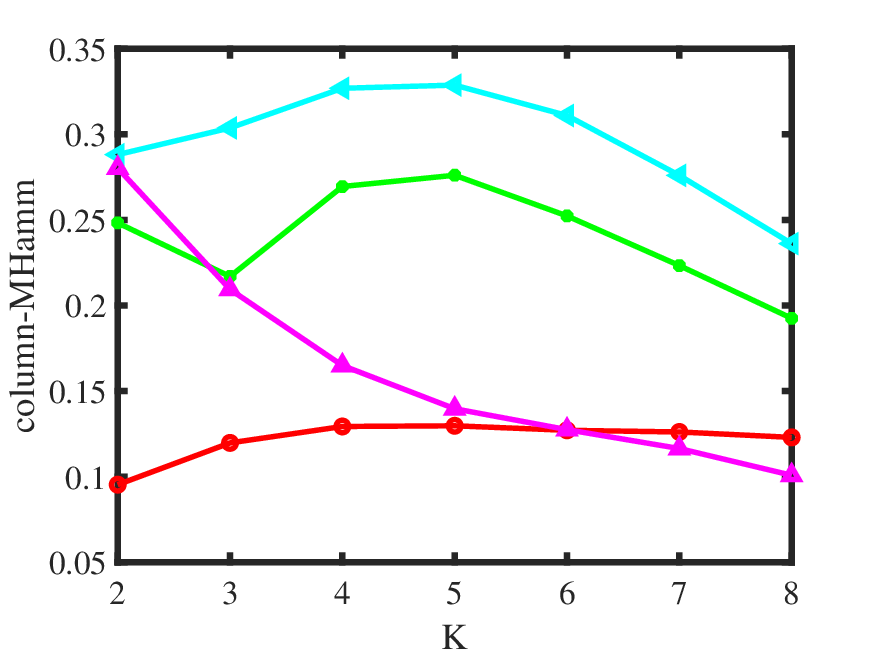}}
\subfigure[Changing $n_{0}$:
run-time]{\includegraphics[width=0.242\textwidth]{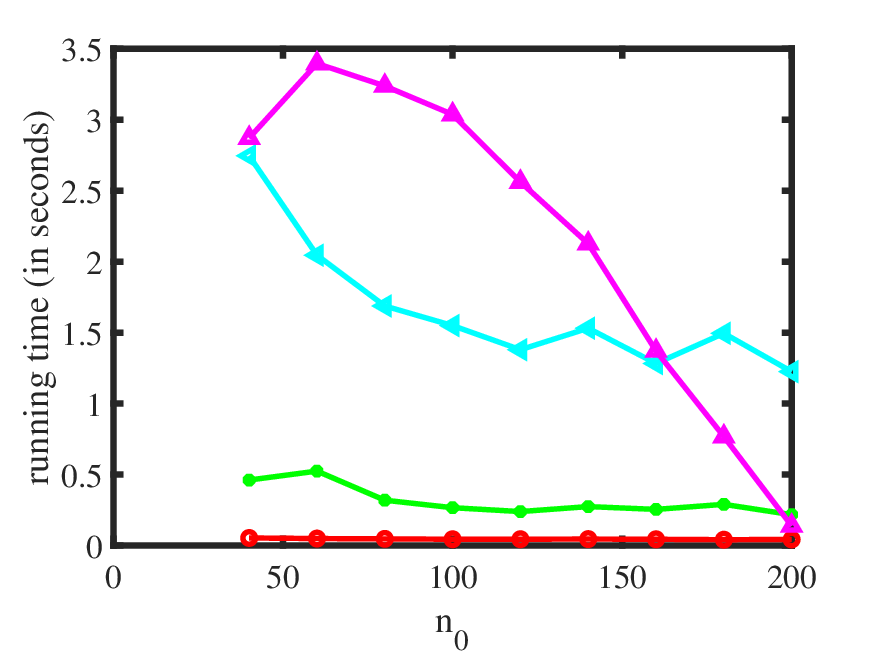}}
\subfigure[Changing $\rho$:
run-time]{\includegraphics[width=0.242\textwidth]{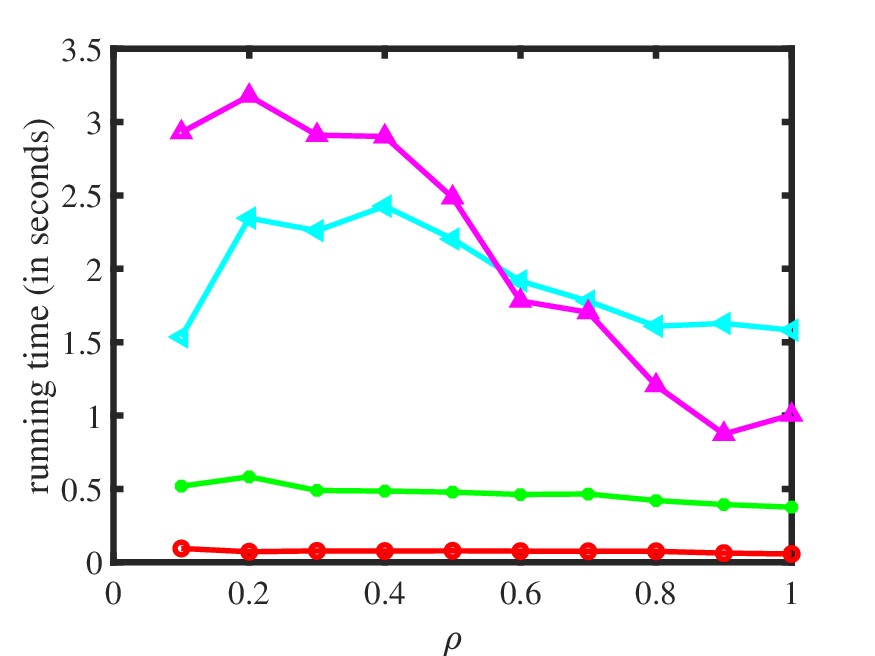}}
\subfigure[Changing $\beta$:
run-time]{\includegraphics[width=0.242\textwidth]{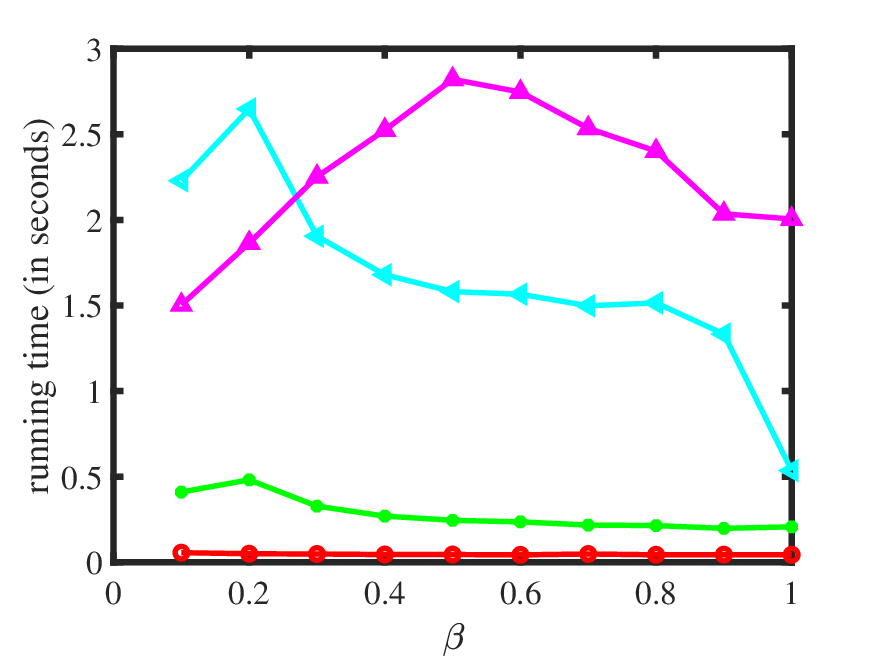}}
\subfigure[Changing $K$:
run-time]{\includegraphics[width=0.242\textwidth]{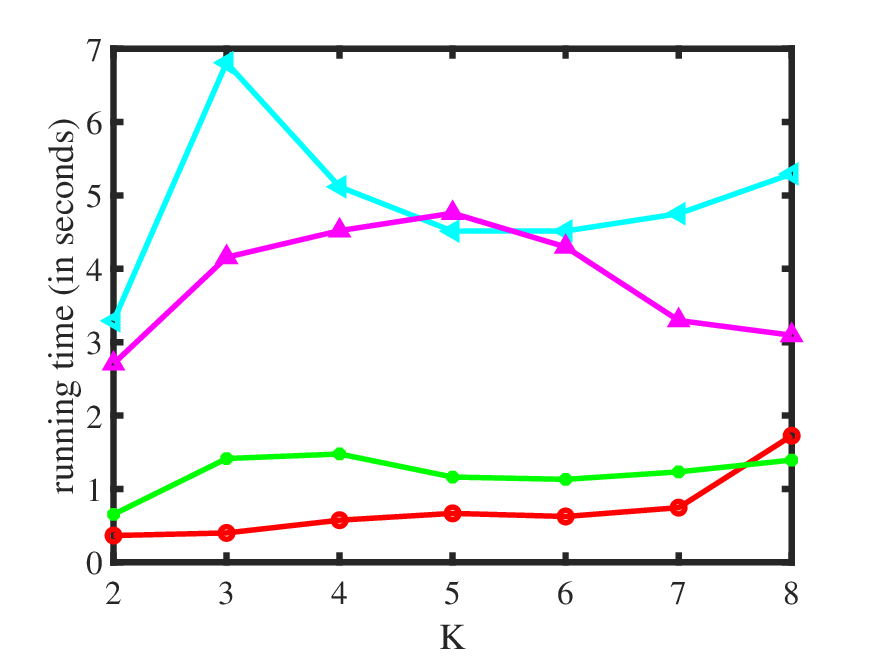}}
\caption{Numerical results of Experiments 4-7.}
\label{EXBi} 
\end{figure}

\texttt{Experiment 4: Changing $n_{0}$.} Let $n_{r}=600, n_{c}=800$, $n_{0}$ range in $\{40, 60,\ldots, 200\}$, and all other parameters are set the same as Experiment 1. Panels (a), (b), and (c) of Fig.~\ref{EXBi} record the error rates of these approaches in Experiment 4, and panel (m) records the run-time. The total run-time of DiSP for Experiment 4 is roughly 36 seconds. We see that as the fraction of pure nodes increases, error rates decrease. Meanwhile, since the size of the network is much larger than the network in Experiment 1, error rates in Experiment 4 are much lower than those of Experiment 1 (similar conclusions hold for Experiments 5-6). Furthermore, similar to the results of Experiment 1, DiSP performs best when $n_{0}$ is small, while it performs slightly poorer than the other three methods for a larger $n_{0}$. We also observe that our DiSP runs faster than Mixed-SCORE, GeoNMF, and SVM-cD (similar conclusions hold for Experiments 5-6).

\texttt{Experiment 5: Changing $\rho$.} Let $n_{r}=600, n_{c}=800, n_{0}=120$, and all other parameters are set the same as in Experiment 2. Error rates are shown in panels (d)-(f) of Fig.~\ref{EXBi}, and panel (n) records the run-time. The total run-time of DiSP for Experiment 5 is roughly 55 seconds. We see that as $\rho$ increases, error rates tend to decrease. Meanwhile, our DiSP performs best for this experiment.

\texttt{Experiment 6: Changing $\beta$.} Let $n_{r}=600, n_{c}=800, n_{0}=120$, and all other parameters are set the same as in Experiment 3. Panels (g), (h), and (i) of Fig.~\ref{EXBi} record the error rates of these approaches in Experiment 6, and panel (o) records the run-time. The total run-time of DiSP for Experiment 6 is roughly 40.6 seconds. We see that as $\beta$ increases, error rates of DiSP decrease, and this is consistent with the theoretical results in the last paragraph of Section \ref{sec4}. For comparison, the error rates of Mixed-SCORE, GeoNMF, and SVM-cD do not decrease as $\beta$ grows for this experiment.

\texttt{Experiment 7: Changing $K$.} Let $n_{r}=1200, n_{c}=1600,$ and $ n_{0}=120$. Set the diagonal elements, upper triangular elements, and lower triangular elements of $P$ as 0.5, 0.2, and 0.3, respectively. $K$ ranges in $\{2, 3, \ldots, 8\}$. Each mixed node is assigned to any of the $K$ blocks uniformly at random, i.e., with probability $1/K$ per block. Panels (j), (k), and (l) of Fig.~\ref{EXBi} record the error rates of DiSP in Experiment 7, and panel (p) records the run-time. The total run-time for Experiment 7 is roughly 407 seconds. From the numerical results, we see that our DiSP performs best when $K\leq4$ and it performs slightly poorer than SVM-cD when $K\geq5$.

\section{Applications to real-world data sets}\label{sec6}
In all real‑world directed networks analyzed here, the sets of row nodes and column nodes coincide. Consequently, we have $n_{r}=n_{c}=n$. Set $d_{r}(i)=\sum_{j=1}^{n}A(i,j)$ as the sending side degree of node $i$, and $d_{c}(i)=\sum_{j=1}^{n}A(j,i)$ as the receiving side degree of node $i$. We find that there exist many nodes with zero degree in real-world directed networks. Before applying our DiSP on the adjacency matrix of real-world directed networks, we need to pre-process the original directed network using the following procedure. 
\begin{algorithm}
\caption{Pre-processing}
\label{alg:prepro}
\begin{algorithmic}[1]
\Require Real-world directed network $\mathcal{N}$.
\State Let $A_{0}$ be the adjacency matrix of the original directed network $\mathcal{N}$.
\State Find the row nodes set in which row nodes have zero degree by setting $S_{r,0}=\{i:\sum_{j=1}^{n}A_{0}(i,j)=0\}$. For column nodes, set $S_{c,0}=\{i:\sum_{j=1}^{n}A_{0}(j,i)=0\}$.
\State Set $S_{0}=S_{r,0}\bigcup S_{c,0}$
\State Update $A_{0}$ by setting $A_{0}=A_{0}(S_{0},S_{0})$.
\State Repeat step 1 and step 2 until $S_{0}$ is a null set.
\State Set $A$ as the largest connected component of $A_{0}$.
\end{algorithmic}
\end{algorithm}

After pre-processing, we let $\hat{\Pi}_{r}$ and $\hat{\Pi}_{c}$ obtained from applying DiSP on $A$ with $n$ nodes and $K$ row (column) communities. Let $\hat{\ell}_{r}$ be an $n\times1$ vector such that $\hat{\ell}_{r}(i)=\mathrm{argmax}_{1\leq k\leq K}\hat{\Pi}_{r}(i,k)$, where $\hat{\ell}_{r}(i)$ is called the home base row community of node $i$. $\hat{\ell}_{c}$ is defined similarly by setting $\hat{\ell}_{c}(i)=\mathrm{argmax}_{1\leq k\leq K}\hat{\Pi}_{c}(i,k)$. We also need the following statistics to investigate the directed network.
\begin{itemize}
\item \texttt{Fraction of estimated highly mixed row (column) nodes:}
Treat row node $i$ as highly mixed if $\mathrm{max}_{1\leq k\leq K}\hat{\Pi}(i,k)\leq0.8$. Let $\tau_{r}$ represent the proportion of highly mixed row nodes such that $\tau_{r}=\frac{|\{i:\mathrm{max}_{1\leq k\leq K}\hat{\Pi}_{r}(i,k)\leq0.8\}|}{n}$. Similarly, define $\tau_{c}=\frac{|\{i:\mathrm{max}_{1\leq k\leq K}\hat{\Pi}_{c}(i,k)\leq0.8\}|}{n}$.
\item \texttt{The measurement of asymmetric structure between row and communities:} Since row nodes and column nodes are the same, to see whether the structure of row clusters differs from the structure of column clusters, we use the following metric:
\begin{align*}
\mathrm{MHamm}=\frac{\mathrm{min}_{O\in S}\|\hat{\Pi}_{r}O-\hat{\Pi}_{c}\|_{1}}{n}.
\end{align*}
We see that a larger (or smaller) $\mathrm{MHamm}$ indicates a heavy (or slight) asymmetry between row communities and column communities.
\item \texttt{Sparsity:} Define $\xi=\frac{\mathrm{Number~of~zeros~in~}A}{n^{2}}$ to quantify the sparsity of a directed network. A higher value of $\xi$ indicates a sparser network.
\end{itemize}

We are now ready to describe some real-world directed networks as follows:

\textbf{Political blogs}: The data originate from the 2004 U.S. presidential election \cite{adamic2005the}. This political blog network is naturally modeled as a directed graph, where every node represents a blog and is labeled either as liberal or conservative, so that the number of communities is $K=2$. A directed edge from node $i$ to node $j$  means blog $i$ hyperlinks to blog $j$; such links are not necessarily reciprocated. This data can be downloaded from \url{http://www-personal.umich.edu/~mejn/netdata/}. The original data has 1490 nodes, after pre-processing by Algorithm \ref{alg:prepro}, $A\in\{0,1\}^{813,813}$.

\begin{figure}
\centering
\subfigure[Political blogs]{\includegraphics[width=0.24\textwidth]{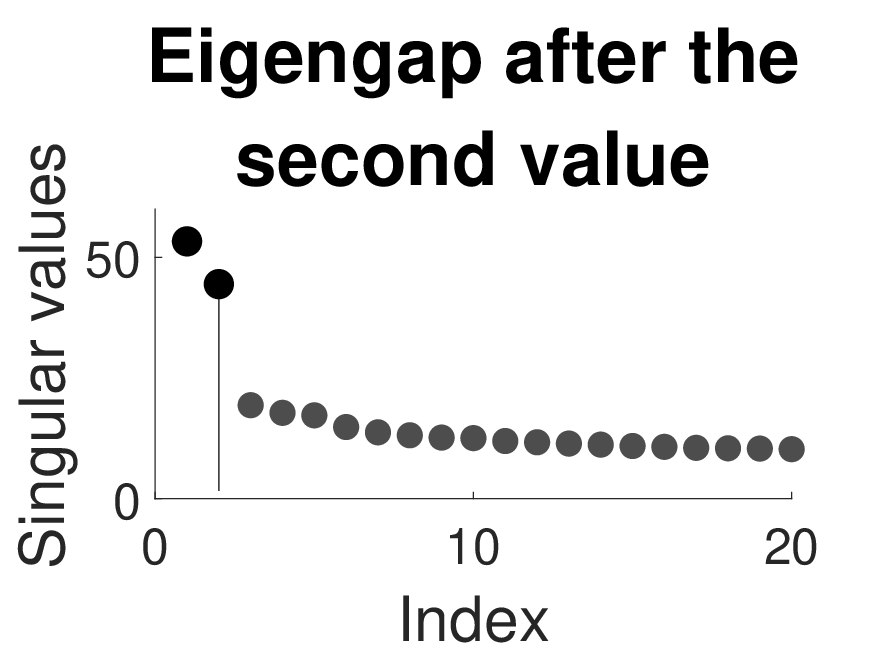}}
\subfigure[Human proteins (Stelzl)]{\includegraphics[width=0.24\textwidth]{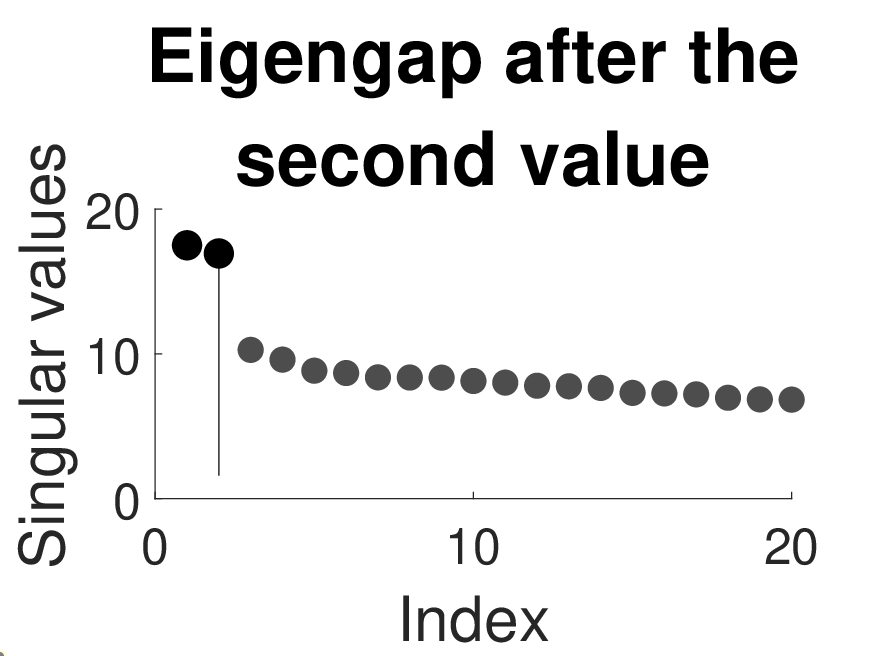}}
\subfigure[Wikipedia links (crh)]{\includegraphics[width=0.24\textwidth]{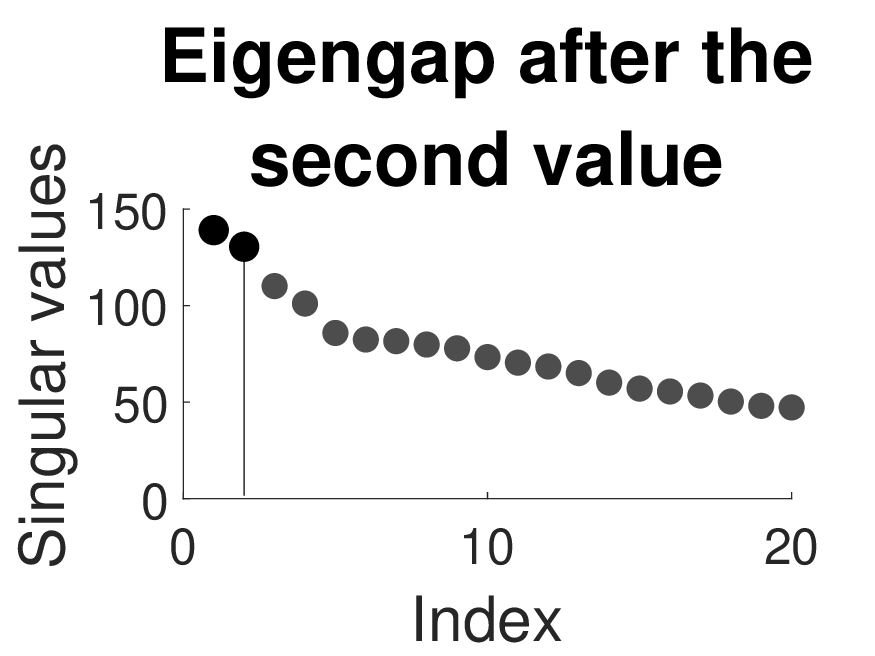}}
\subfigure[Wikipedia links (dv)]{\includegraphics[width=0.24\textwidth]{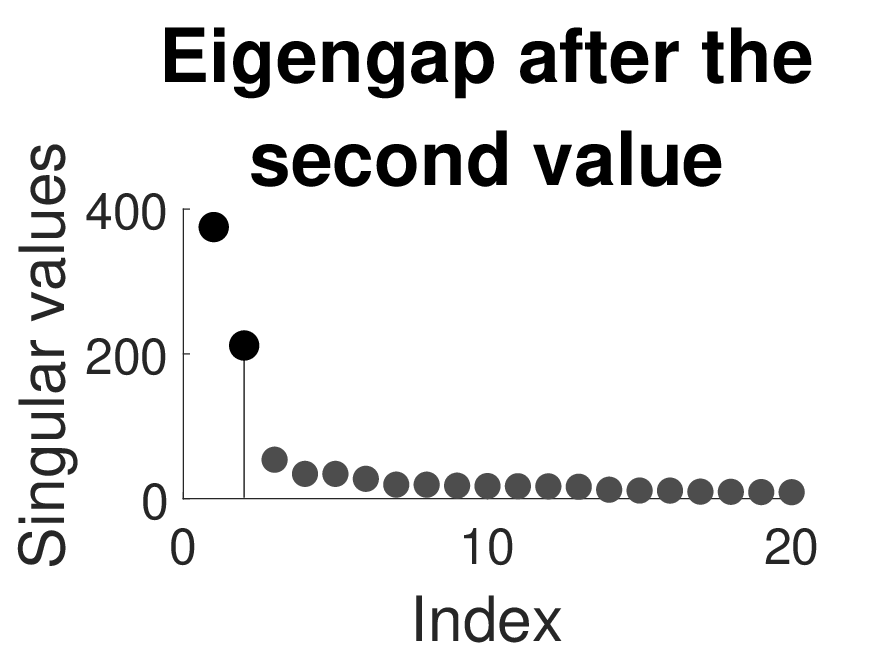}}
\caption{The first twenty singular values of $A$s corresponding to the four real networks.}
\label{Leading20}
\end{figure}

\textbf{Human proteins (Stelzl)}: This network can be downloaded from \url{http://konect.cc/networks/maayan-Stelzl}, and it represents interacting pairs of proteins in Humans (Homo sapiens) \cite{stelzl2005human}. In this data, node means protein and edge means interaction. The raw dataset contains 1706 nodes. After pre-processing, we obtain $A\in\{0,1\}^{1507\times1507}$. The number of row (column) clusters is unknown. For its determination, $A$'s first twenty singular values are displayed in panel (b) of Fig.~\ref{Leading20} and we find that the eigengap suggests $K=2$. Meanwhile, \cite{DISIM} also uses the idea of eigengap to choose $K$ for directed networks.

\textbf{Wikipedia links (crh)}: This dataset comprises the wikilink network of Wikipedia for the Crimean Turkish language (crh). The data is available for download at \url{http://konect.cc/networks/wikipedia_link_crh/}. In this network, a node denotes an article, and an edge denotes a wikilink \cite{kunegis2013konect}. After pre-processing, there are 3555 nodes. For this network, panel (c) of Fig.~\ref{Leading20} implies two communities.

\textbf{Wikipedia links (dv)}: In this data, nodes are Wikipedia articles and directed edges are wikilinks \cite{kunegis2013konect}. This network is available at \url{http://konect.cc/networks/wikipedia_link_dv/}. After pre-processing, $A\in\{0,1\}^{2394\times 2394}$. $K=2$ for this data. Panel (d) of Fig.~\ref{Leading20} suggests $K=2$ for this data.

\begin{figure}
\centering
\subfigure[Political blogs]{\includegraphics[width=0.36\textwidth]{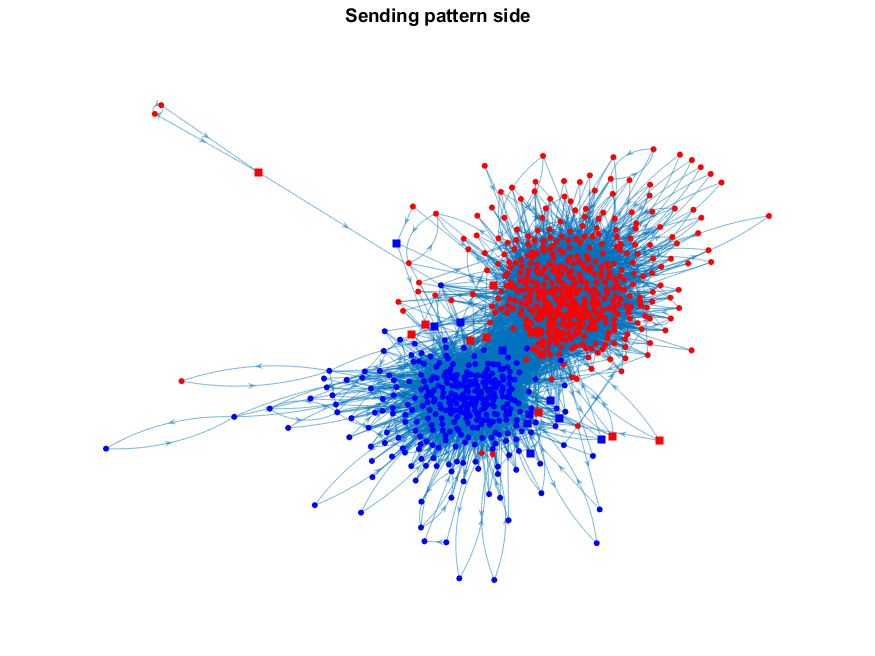}}
\subfigure[Political blogs]{\includegraphics[width=0.36\textwidth]{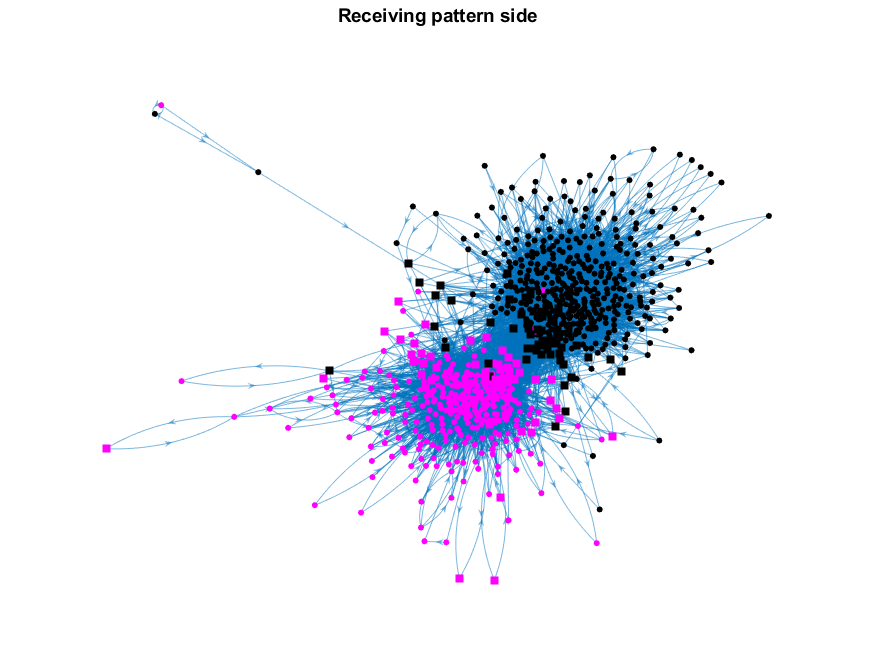}}
\subfigure[Human proteins (Stelzl)]{\includegraphics[width=0.36\textwidth]{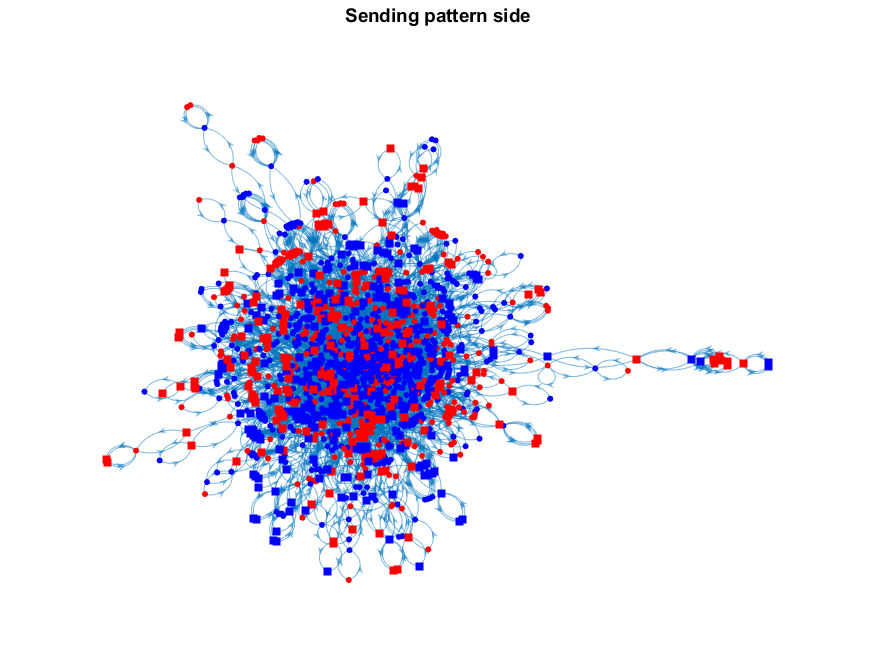}}
\subfigure[Human proteins (Stelzl)]{\includegraphics[width=0.36\textwidth]{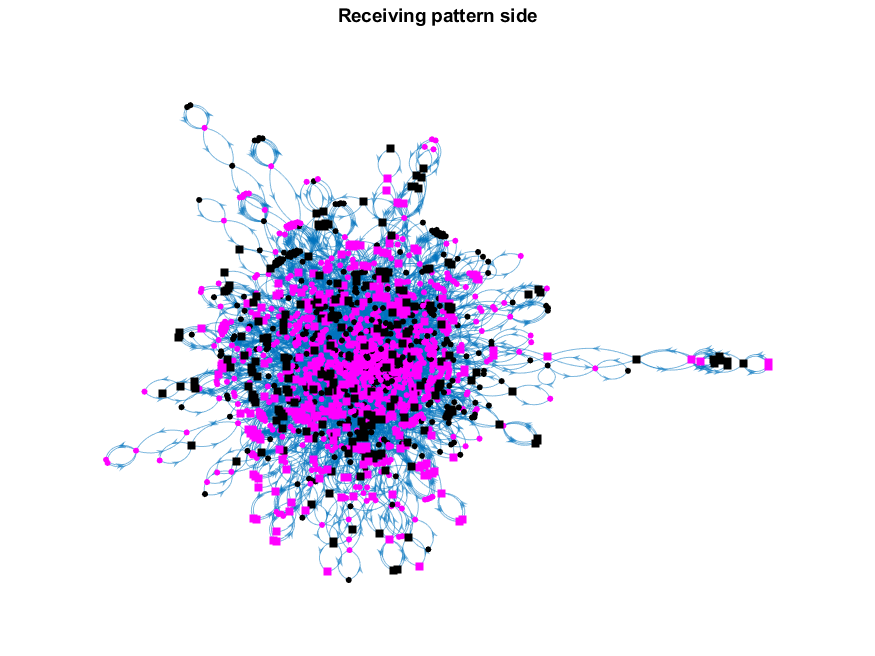}}
\subfigure[Wikipedia links (crh)]{\includegraphics[width=0.36\textwidth]{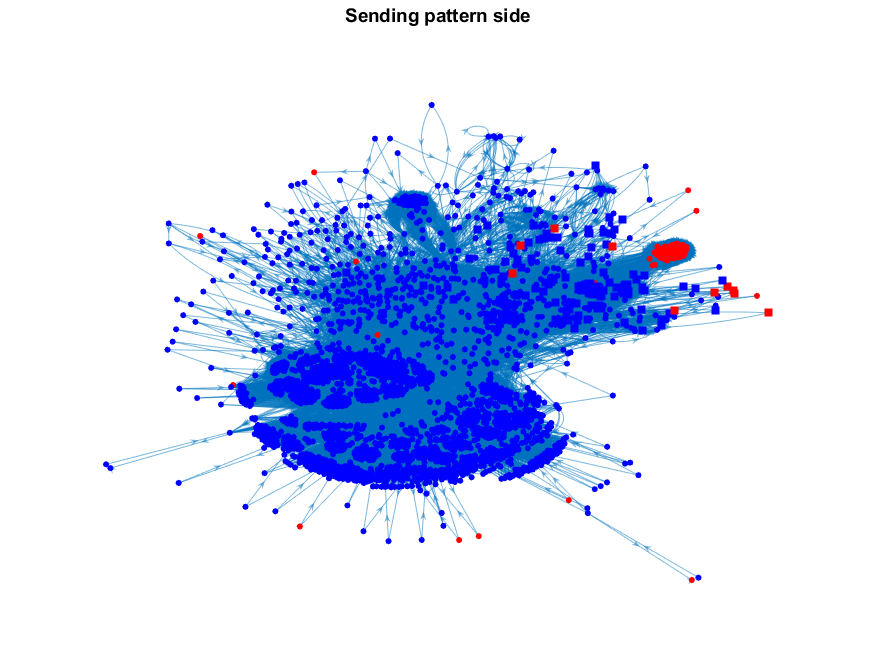}}
\subfigure[Wikipedia links (crh)]{\includegraphics[width=0.36\textwidth]{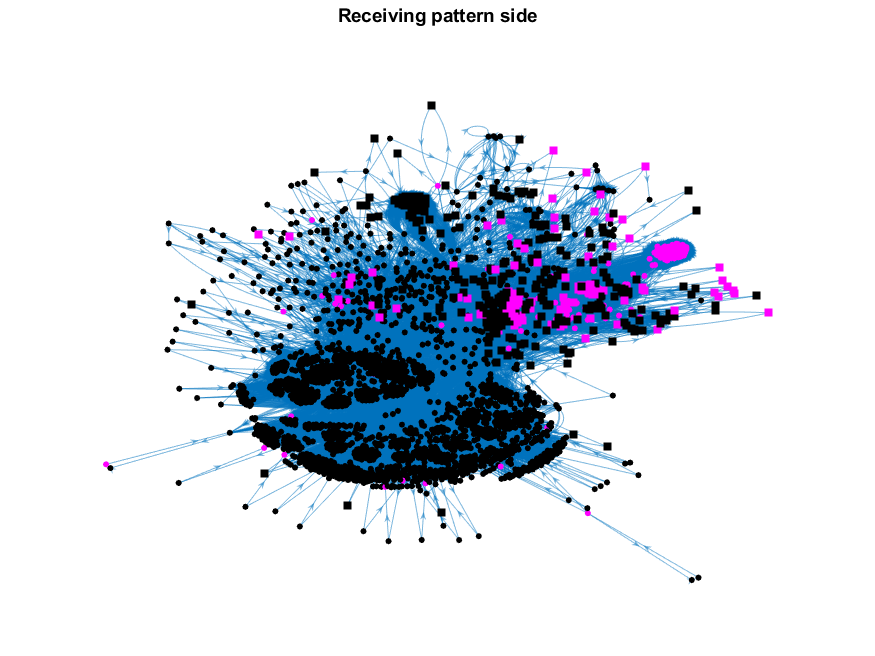}}
\subfigure[Wikipedia links (dv)]{\includegraphics[width=0.36\textwidth]{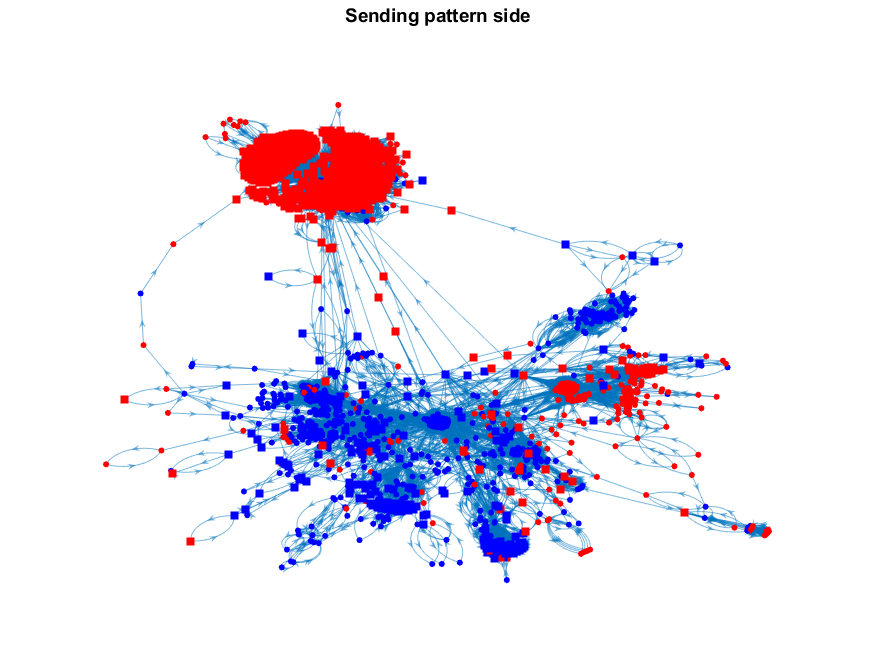}}
\subfigure[Wikipedia links (dv)]{\includegraphics[width=0.36\textwidth]{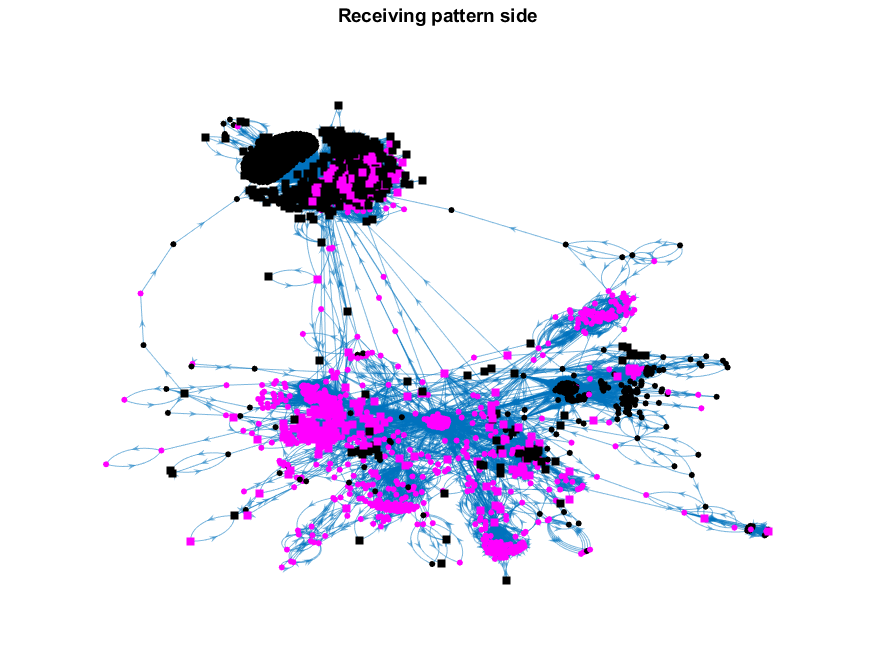}}
\caption{Sending and receiving clusters detected by DiSP for directed networks considered in this paper. Colors indicate clusters, and squares indicate highly mixed nodes, where sending and receiving clusters are obtained by $\hat{\ell}_{r}$ and $\hat{\ell}_{c}$, respectively.}
\label{NetSR}
\end{figure}

After obtaining $A$ and $K$ for real-world directed networks analyzed in this paper, we apply our DiSP to $A$ and report $\tau_{r},\tau_{c}$, $\mathrm{MHamm}$, $\xi$, and running time in Table \ref{realdata}. The results show that all of these networks are highly sparse, with over 96\% of the entries in the adjacency matrix $A$ being zeros. Meanwhile, Political blogs, Human proteins (Stelzl), and Wikipedia links (crh)  exhibit only a weak asymmetry between their row communities and column communities, because their $\mathrm{MHamm}$ is small, while row clusters differ a lot from column clusters for Wikipedia links (dv) due to its large $\mathrm{MHamm}$. Political blogs network has $0.0246\times813\approx20$ highly mixed nodes in the sending pattern side, while it has $813\times0.1353\approx110$ highly mixed nodes in the receiving pattern side. Similar computations hold for the other three datasets. In particular, Wikipedia links (dv) owns a large proportion of highly mixed nodes in both the sending and receiving pattern sides. Additionally, DiSP is computationally efficient, with running time ranging from 0.0120 seconds for the Political blogs dataset to 0.6384 seconds for the Wikipedia links (crh) dataset, indicating its scalability to larger networks. To provide a visual illustration, Fig.~\ref{NetSR} displays both the sending clusters and the receiving clusters inferred by DiSP from the real‑world directed networks studied in this work, where we also mark the highly mixed nodes by squares. Generally, we see that DiSP is useful in finding highly mixed nodes and studying the asymmetric structure between row and column clusters of a directed network.
\begin{table}[h!]
\small
	\centering
	\caption{The proportion of highly mixed nodes, the asymmetric structure measured by $\mathrm{MHamm}$, the sparsity index $\xi$, and the running time (in seconds) for real-world directed networks used in this paper by applying DiSP to their adjacency matrices assuming that there are $K=2$ row (column) communities. Here, DiSP's running time represents the average of 100 independent repetitions for each network.}
	\label{realdata}
\begin{tabular}{cccccccccc}
\hline\hline
data&$\tau_{r}$&$\tau_{c}$&$\mathrm{MHamm}$&$\xi$&Running time\\
\hline
Political blogs&0.0246&0.1353&0.0901&97.59\%&0.0120s\\
Human proteins (Stelzl)&0.2986&0.2999&0.0115&99.74\%&0.0760s\\
Wikipedia links (crh)&0.0444&0.1308&0.0643&98.67\%&0.6384s\\
Wikipedia links (dv)&0.4089&0.3008&0.1804&96.52\%&0.1175s\\
\hline\hline
\end{tabular}
\end{table}
\section{Discussions}\label{sec7}
In this paper, we introduce a directed mixed membership stochastic blockmodel to model directed networks with mixed memberships. DiMMSB allows both row and column nodes to have mixed memberships, and the numbers of row nodes and column nodes could be different. To fit this model, we develop an efficient approach DiSP and built its estimation consistency. Meanwhile, we also obtain the separation conditions of a standard directed mixed membership network. Our theoretical results are consistent with Theorem 2.2 in \cite{MixedSCORE}  when both models simplify to MMSB under mild conditions. Through the applications on some real-world directed networks, DiSP finds highly mixed nodes, and it also provides a deeper understanding of the asymmetries embedded in the connectivity patterns of these directed systems. The model DiMMSB developed in this paper is useful to model directed networks and generate directed mixed membership networks with true background membership matrices. When analyzing directed networks, our DiSP algorithm severs as a valuable tool for studying how the row‑side (sending) communities differ structurally from the column‑side (receiving) communities. We expect that the DiMMSB model and the DiSP algorithm will have applications beyond this paper and can be widely applied to study the properties of directed networks in network science.

While DiSP shows promising performance in synthetic and real-world networks, several open questions and research directions remain. Firstly, DiSP may be sensitive to data noise, especially at low signal-to-noise ratios \citep{jin2023optimal}. Exploring DiSP's theoretical guarantees across various signal-to-noise ratio levels is a challenging yet promising direction. Secondly, determining the number of communities in directed networks, whether generated by our DiMMSB model or other models like MMSB and ScBM, remains challenging. Extending the stepwise goodness-of-fit approach \citep{jin2023optimal} to DiMMSB could aid in determining $K$. Thirdly, enhancing DiSP's efficiency for large-scale directed networks is a meaningful research area. Random-projection and random-sampling techniques developed in \citep{guo2023randomized} may offer valuable solutions. Lastly, this study focused on static-directed networks. Extending these findings to dynamic-directed networks is a promising future direction.
\section*{CRediT authorship contribution statement}
\textbf{Huan Qing:} Conceptualization, Methodology, Investigation, Software, Formal analysis, Data curation, Writing – original draft, Writing – review \& editing, Funding acquisition. \textbf{Jingli Wang:} Data curation, Validation, Visualization, Writing – review \& editing, Funding acquisition.
\section*{Declaration of competing interest}
The authors declare no competing interests.
\section*{Data availability}
The data used in this study are publicly available.

\section*{Acknowledgements}
The authors would like to thank Dr. Edoardo M. Airoldi and  Dr. Xiaopei Wang for sharing codes of vEM  \cite{airoldi2013multi} with us. Qing's work was sponsored by the Scientific Research Foundation of Chongqing University of Technology (Grant No. 2024ZDR003), and the Science and Technology Research Program of Chongqing Municipal Education Commission (Grant No. KJQN202401168). Wang's work was supported by the National Natural Science Foundation of China (Grant 12001295, 12271272).

\bibliographystyle{elsarticle-num}
\bibliography{refDiMMSB}

\appendix

\section{Equivalence algorithm}\label{sec31}
For the convenience of theoretical analysis, we introduce an equivalent algorithm DiSP-equivalence which returns the same estimations as Algorithm \ref{alg:DiSP} (see Remark \ref{BenefitEquivalence} for details).
Denote $U_{2}=UU'\in\mathbb{R}^{n_{r}\times n_{r}}, \hat{U}_{2}=\hat{U}\hat{U}'\in\mathbb{R}^{n_{r}\times n_{r}}, V_{2}=VV'\in\mathbb{R}^{n_{c}\times n_{c}}, \hat{V}_{2}=\hat{V}\hat{V}'\in\mathbb{R}^{n_{c}\times n_{c}}$. The next lemma guarantees that $U_{2}$ and $V_{2}$ have simplex structures.
\begin{lem}\label{ExistBs2}
Under $DiMMSB(n_{r}, n_{c}, K, P,\Pi_{r}, \Pi_{c})$, we have $U_{2}=\Pi_{r}U_{2}(\mathcal{I}_{r},:)$ and $V_{2}=\Pi_{c}V_{2}(\mathcal{I}_{c},:)$.
\end{lem}

Since $U_{2}(\mathcal{I}_{r},:)\in \mathbb{R}^{K\times n_{r}}$ and $V_{2}(\mathcal{I}_{c},:)\in \mathbb{R}^{K\times n_{c}}$, $U_{2}(\mathcal{I}_{r},:)$ and $V_{2}(\mathcal{I}_{c},:)$ are singular matrices with rank $K$ by condition (I1). Lemma \ref{ExistBs2} gives that
\begin{align*}
\Pi_{r}=U_{2}U'_{2}(\mathcal{I}_{r},:)(U_{2}(\mathcal{I}_{r},:)U'_{2}(\mathcal{I}_{r},:))^{-1}, \Pi_{c}=V_{2}V'_{2}(\mathcal{I}_{c},:)(V_{2}(\mathcal{I}_{c},:)V'_{2}(\mathcal{I}_{c},:))^{-1}.
\end{align*}

Based on the above analysis, we are now ready to give the Ideal DiSP-equivalence. Input $\Omega$ and $K$. Output: $\Pi_{r}$ and $\Pi_{c}$.
\begin{itemize}
  \item \texttt{PCA step.} Obtain $U_{2}$ and $V_{2}$.
  \item \texttt{VH step.} Apply the SP algorithm on rows of $U_{2}$ to obtain $U_{2}(\mathcal{I}_{r},:)$ and on rows of $V_{2}$ to obtain $V_{2}(\mathcal{I}_{c},:)$ assuming there are $K$ row (column) communities.
  \item \texttt{MR step.} Recover $\Pi_{r}=U_{2}U'_{2}(\mathcal{I}_{r},:)(U_{2}(\mathcal{I}_{r},:)U'_{2}(\mathcal{I}_{r},:))^{-1}, \Pi_{c}=V_{2}V'_{2}(\mathcal{I}_{c},:)(V_{2}(\mathcal{I}_{c},:)V'_{2}(\mathcal{I}_{c},:))^{-1}$.
\end{itemize}

We now extend the ideal case to the real one as follows.
\begin{algorithm}
\caption{\textbf{DiSP-equivalence}}
\label{alg:DiSPequivalence}
\begin{algorithmic}[1]
\Require The adjacency matrix $A\in \mathbb{R}^{n_{r}\times n_{c}}$, the number of row (column)  communities $K$.
\Ensure The estimated $n_{r}\times K$ row membership matrix $\hat{\Pi}_{r,2}$ and the estimated $n_{c}\times K$ column membership matrix $\hat{\Pi}_{c,2}$.
\State \texttt{PCA step.} Compute $\hat{U}_{2}\in\mathbb{R}^{n_{r}\times n_{r}}$ and  $\hat{V}_{2}\in \mathbb{R}^{n_{c}\times n_{c}}$ of $A$.
\State \texttt{VH step.} Apply SP algorithm on $\hat{U}_{2}$ with $K$ row clusters to obtain $\hat{U}_{2}(\hat{\mathcal{I}}_{r,2},:)\in\mathbb{R}^{K\times n_{r}}$ where $\mathcal{\hat{I}}_{r,2}$ is the index set returned by SP algorithm.  Similarly, apply SP algorithm on $\hat{V}_{2}$ with $K$ column clusters to obtain $\hat{V}_{2}(\hat{\mathcal{I}}_{c,2},:)\in\mathbb{R}^{K\times n_{c}}$ where $\mathcal{\hat{I}}_{c,2}$ is the index set returned by SP algorithm.
\State \texttt{Membership Reconstruction (MR) step.} Compute the $n_{r}\times K$ matrix $\hat{Y}_{r,2}$ such that $\hat{Y}_{r,2}=\hat{U}_{2}\hat{U}'_{2}(\hat{\mathcal{I}}_{r,2},:)(\hat{U}_{2}(\hat{\mathcal{I}}_{r,2},:)\hat{U}'_{2}(\hat{\mathcal{I}}_{r,2},:))^{-1}$. Set $\hat{Y}_{r,2}=\mathrm{max}(0, \hat{Y}_{r,2})$ and estimate $\Pi_{r,2}(i,:)$ by $\hat{\Pi}_{r,2}(i,:)=\hat{Y}_{r,2}(i,:)/\|\hat{Y}_{r,2}(i,:)\|_{1}, 1\leq i\leq n_{r}$.  Similarly, compute the $n_{c}\times K$ matrix $\hat{Y}_{c,2}$ such that $\hat{Y}_{c,2}=\hat{V}_{2}\hat{V}'_{2}(\hat{V}_{2}(\hat{\mathcal{I}}_{c,2},:)\hat{V}'_{2}(\hat{\mathcal{I}}_{c,2},:))^{-1}$. Set $\hat{Y}_{c,2}=\mathrm{max}(0, \hat{Y}_{c,2})$ and estimate $\Pi_{c,2}(j,:)$ by $\hat{\Pi}_{c,2}(j,:)=\hat{Y}_{c,2}(j,:)/\|\hat{Y}_{c,2}(j,:)\|_{1}, 1\leq j\leq n_{c}$.
\end{algorithmic}
\end{algorithm}
Lemma 3.2 in \cite{mao2020estimating} gives $\hat{\mathcal{I}}_{r}=\hat{\mathcal{I}}_{r,2}$ and $\hat{\mathcal{I}}_{c}=\hat{\mathcal{I}}_{c,2}$ (i.e., SP algorithm will return the same indices in both $\hat{U}$ and $\hat{U}_{2}$ as well as $\hat{V}$ and $\hat{V}_{2}$), which gives that $\hat{U}_{2}\hat{U}'_{2}(\hat{\mathcal{I}}_{r,2},:)=\hat{U}_{2}\hat{U}'_{2}(\hat{\mathcal{I}}_{r},:)=\hat{U}\hat{U}'((\hat{U}\hat{U}')(\hat{\mathcal{I}}_{r},:))'=\hat{U}\hat{U}'(\hat{U}(\hat{\mathcal{I}}_{r},:)\hat{U}')'=\hat{U}\hat{U}'\hat{U}\hat{U}'(\hat{\mathcal{I}}_{r},:)=\hat{U}\hat{U}'(\hat{\mathcal{I}}_{r},:)$, and $\hat{U}_{2}(\hat{\mathcal{I}}_{r,2},:)\hat{U}'_{2}(\hat{\mathcal{I}}_{r,2},:)=\hat{U}_{2}(\hat{\mathcal{I}}_{r},:)\hat{U}'_{2}(\hat{\mathcal{I}}_{r},:)=\hat{U}(\hat{\mathcal{I}}_{r},:)\hat{U}'(\hat{U}(\hat{\mathcal{I}}_{r},:)\hat{U}')'=\hat{U}(\hat{\mathcal{I}}_{r},:)\hat{U}'(\hat{\mathcal{I}}_{r},:)$. Therefore, $\hat{Y}_{r,2}=\hat{Y}_{r}, \hat{\Pi}_{r,2}=\hat{\Pi}_{r}$. Following a similar analysis, we also have $\hat{Y}_{c,2}=\hat{Y}_{c},$ and $\hat{\Pi}_{c,2}=\hat{\Pi}_{c}$. Hence, the above analysis guarantees that the two algorithms \ref{alg:DiSP} and \ref{alg:DiSPequivalence} return the same estimations for both row and column nodes' memberships.

\section{Proof for identifiability}
\subsection{Proof of Proposition \ref{id}}
\begin{proof}
To prove the identifiability, we follow a similar idea as the proof of (a) in Theorem 2.1 \cite{mao2020estimating}, which provides the proof of the identifiability of MMSB. Let $\Omega=U\Lambda V'$ be the compact singular value decomposition of $\Omega$. By Lemma \ref{RISCIS}, $U=\Pi_{r}B_{r}, V=\Pi_{c}B_{c}$. Thus, for any node $i$, $U(i,:)$ lies in the convex hull of the $K$ rows of $B_{r}$, i.e., $U(i,:)\subseteq \mathrm{Conv}(B_{r})$ for $1\leq i\leq n_{r}$. Similarly, we have $V(j,:)\subseteq \mathrm{Conv}(B_{c})$ for $1\leq j\leq n_{c}$, where we use $\mathrm{Conv}(M)$ denote the convex hull of the rows of the matrix $M$.

Now, if $\Omega$ can be generated by another set of parameters $(\tilde{\Pi}_{r}, \tilde{P}, \tilde{\Pi}_{c})$ (i.e.,  $\Omega=\Pi_{r}P\Pi_{c}'=\tilde{\Pi}_{r}\tilde{P}\tilde{\Pi}_{c}'$), where $\tilde{\Pi}_{r}$ and $\tilde{\Pi}_{c}$ have different pure node sets, with indices $\tilde{\mathcal{I}}_{r}\neq 1:K, \tilde{\mathcal{I}}_{c}\neq 1:K$. By the previous argument, we  have $U(\tilde{\mathcal{I}}_{r},:)\subseteq \mathrm{Conv}(B_{r})$ and $V(\tilde{\mathcal{I}}_{c},:)\subseteq \mathrm{Conv}(B_{c})$. Since $(\Pi_{r}, P, \Pi_{c})$ and $(\tilde{\Pi}_{r}, \tilde{P}, \tilde{\Pi}_{c})$ generate the same $\Omega$, they have the same compact singular value decomposition up to a permutation of communities. Thus, swapping the roles of $\Pi_{r}$ and $\tilde{\Pi}_{r}$, and reapplying the above argument, we have $B_{r}\subseteq \mathrm{Conv}(U(\tilde{\mathcal{I}}_{r},:))$. Then $\mathrm{Conv}(B_{r})\subseteq \mathrm{Conv}(U(\tilde{\mathcal{I}}_{r},:))\subseteq\mathrm{Conv}(B_{r})$, therefore we must have $\mathrm{Conv}(B_{r})=\mathrm{Conv}(U(\tilde{\mathcal{I}}_{r},:))$. This means that pure nodes in $\Pi_{r}$ and $\tilde{\Pi}_{r}$ are aligned up to a permutation, i.e., $U(\tilde{\mathcal{I}}_{r},:)=M_{r}B_{r}$, where $M_{r}\in \mathbb{R}^{K\times K}$ is a permutation matrix. Similarly, we have $V(\tilde{\mathcal{I}}_{c},:)=M_{c}B_{c}$, where $M_{c}\in \mathbb{R}^{K\times K}$ is a permutation matrix.

By Lemma \ref{RISCIS}, we have $U=\Pi_{r}B_{r}$ and $U=\tilde{\Pi}_{r}U(\tilde{\mathcal{I}}_{r},:)$, combining with $U(\tilde{\mathcal{I}}_{r},:)=M_{r}B_{r}$, we have
\begin{align*}
(\Pi_{r}-\tilde{\Pi}_{r}M_{r})B_{r}=0.
\end{align*}
Since $\mathrm{rank}(P)=K$ based on Condition (I1), we have $\mathrm{rank}(B_{r})=K$, i.e., $B_{r}$ is  full rank. So we have $\Pi_{r}=\tilde{\Pi}_{r}M_{r}$. Thus, $\Pi_{r}$ and $\tilde{\Pi}_{r}$ are identical up to a permutation. Similarly, $\Pi_{c}=\tilde{\Pi}_{c}M_{c}$, i.e., $\Pi_{c}$ and $\tilde{\Pi}_{c}$ are identical up to a permutation. To have the same $\Omega$, we have
\begin{align*}
\Pi_{r}P\Pi_{c}'&=\tilde{\Pi}_{r}\tilde{P}\tilde{\Pi}_{c}'\\
&\Downarrow\\
\tilde{\Pi}_{r}M_{r}P(\tilde{\Pi}_{c}M_{c})'&=\tilde{\Pi}_{r}\tilde{P}\tilde{\Pi}_{c}'\\
&\Downarrow\\
M_{r}PM_{c}'&=\tilde{P},
\end{align*}
where the last equality holds by Lemma \ref{PiX} and condition (I2). $M_{r}PM_{c}'=\tilde{P}$ gives that $P$ and $\tilde{P}$ are identical up to a row permutation and a column permutation.

\begin{lem}\label{PiX}
For any membership matrix $\Pi\in\mathbb{R}^{n\times K}$ whose $i$-th row $[\Pi(i,1), \Pi(i,2), \ldots, \Pi(i,K)]$ is the PMF of node $i$ for $1\leq i\leq n$, such that each community has at least one pure node, then for any $X,\tilde{X}\in\mathbb{R}^{K\times K}$, if $\Pi X=\Pi\tilde{X}$, we have $X=\tilde{X}$.
\end{lem}
\begin{proof}	
Assume that node $i$ is a pure node such that $\Pi(i,k)=1$, then the $i$-th row of $\Pi X$ is $[X(k,1), X(k,2), \ldots, X(k,K)]$ (i.e., the $i$-th row of $\Pi X$ is the $k$-th row of $X$ if $\Pi(i,k)=1$); similarly, the $i$-th row of $\Pi \tilde{X}$ is the $k$-th row of $\tilde{X}$. Since $\Pi X=\Pi\tilde{X}$, we have $[X(k,1), X(k,2), \ldots, X(k,K)]=[\tilde{X}(k,1), \tilde{X}(k,2), \ldots, \tilde{X}(k,K)]$ for $1\leq k\leq K$, hence $X=\tilde{X}$.
\end{proof}
\begin{rem}
Here, we propose an alternative proof of DiMMSB's identifiability. As in the main text, we always set $\Pi_{r}(\mathcal{I}_{r},:)=I_{K}$ and $\Pi_{c}(\mathcal{I}_{c},:)=I_{K}$. By Lemma \ref{RISCIS}, we have $U=\Pi_{r}U(\mathcal{I}_{r},:)=\tilde{\Pi}_{r}U(\mathcal{I}_{r},:)$ and $U(\mathcal{I}_{r},:)$ is invertible based on Conditions (I1) and (I2), which gives $\Pi_{r}=\tilde{\Pi}_{r}$. Similarly, we have $\Pi_{c}=\tilde{\Pi}_{c}$. By Lemma \ref{PiX}, we have $P=\tilde{P}$. Here, there is no need to consider permutation since we set $\Pi_{r}(\mathcal{I}_{r},:)=I_{K}$ and $\Pi_{c}(\mathcal{I}_{c},:)=I_{K}$. Note that, in this proof, the invertibility of $U(\mathcal{I}_{r},:)$ and $V(\mathcal{I}_{c},:)$ requires the number of row communities equals that of column communities, and this is the reason we do not model a directed mixed membership network whose number of row communities does not equal that of column communities in the definition of DiMMSB.
\end{rem}
\end{proof}
\section{Ideal simplex}
\subsection{Proof of Lemma \ref{RISCIS}}
\begin{proof}
Since $\Omega=U\Lambda V'$ and $V'V=I_{K}$, we have $U=\Omega V\Lambda^{-1}$. Recall that $\Omega=\Pi_{r}P\Pi'_{c}$, we have $U=\Pi_{r}P\Pi'_{c}V\Lambda^{-1}=\Pi_{r}B_{r}$, where we set $B_{r}=P\Pi'_{c}V\Lambda^{-1}$. Since $U(\mathcal{I}_{r},:)=\Pi_{r}(\mathcal{I}_{r},:)B_{r}=B_{r}$, we have $B_{r}=U(\mathcal{I}_{r},:)$. For $1\leq i\leq n_{r}$, $U(i,:)=e'_{i}\Pi_{r}B_{r}=\Pi_{r}(i,:)B_{r}$, so sure we have $U(i,:)=U(\bar{i},:)$ when $\Pi_{r}(i,:)=\Pi_{r}(\bar{i},:)$. Follow a similar analysis for $V$, and this lemma holds surely.
\end{proof}
\subsection{Proof of Theorem \ref{IdealDiSP}}
\begin{proof}
For column nodes, Remark \ref{inputUinSP} guarantees that the SP algorithm returns $\mathcal{I}_{r}$ when the input is $U$ with $K$ row communities, hence Ideal DiSP recovers $\Pi_{r}$ exactly. Similar, for recovering $\Pi_{c}$ from $V$, this theorem follows.
\end{proof}
\subsection{Proof of Lemma \ref{ExistBs2}}
\begin{proof}
By Lemma \ref{RISCIS}, we know that $U=\Pi_{r}U(\mathcal{I}_{r},:)$, which gives that $U_{2}=UU'=\Pi_{r}U(\mathcal{I}_{r},:)U'=\Pi_{r}(UU')(I_{r},:)=\Pi_{r}U_{2}(\mathcal{I}_{r},:)$. Similar to $V_{2}$, thus this lemma holds.
\end{proof}
\section{Basic properties of $\Omega$}
\begin{lem}\label{P2}
Under $DiMMSB(n_{r}, n_{c}, K, P,\Pi_{r}, \Pi_{c})$, we have
\begin{align*}
&\sqrt{\frac{1}{K\lambda_{1}(\Pi'_{r}\Pi_{r})}}\leq\|U(i,:)\|_{F}\leq \sqrt{\frac{1}{\lambda_{K}(\Pi'_{r}\Pi_{r})}}, \qquad 1\leq i\leq n_{r},
\\
&\sqrt{\frac{1}{K\lambda_{1}(\Pi'_{c}\Pi_{c})}}\leq\|V(j,:)\|_{F}\leq \sqrt{\frac{1}{\lambda_{K}(\Pi'_{c}\Pi_{c})}}, \qquad 1\leq j\leq n_{c}.
\end{align*}
\end{lem}
\begin{proof}
For $\|U(i,:)\|_{F}$, since $U=\Pi_{r}B_{r}$, we have
\begin{align*}
\mathrm{min}_{i}\|e'_{i}U\|^{2}_{F}&=\mathrm{min}_{i}e'_{i}UU'e_{i}=\mathrm{min}_{i}\Pi_{r}(i,:)B_{r}B'_{r}\Pi'_{r}(i,:)=\mathrm{min}_{i}\|\Pi_{r}(i,:)\|^{2}_{F}\frac{\Pi_{r}(i,:)}{\|\Pi_{r}(i,:)\|_{F}}B_{r}B'_{r}\frac{\Pi'_{r}(i,:)}{\|\Pi_{r}(i,:)\|_{F}}\\
&\geq \mathrm{min}_{i}\|\Pi_{r}(i,:)\|^{2}_{F}\mathrm{min}_{\|x\|_{F}=1}x'B_{r}B'_{r}x=\mathrm{min}_{i}\|\Pi_{r}(i,:)\|^{2}_{F}\lambda_{K}(B_{r}B'_{r})\overset{\mathrm{By~Lemma~}\ref{P3}}{=}\frac{\mathrm{min}_{i}\|\Pi_{r}(i,:)\|^{2}_{F}}{\lambda_{1}(\Pi'_{r}\Pi_{r})}\\
&\geq \frac{1}{K\lambda_{1}(\Pi'_{r}\Pi_{r})},
\end{align*}
where the last inequality holds because $\|\Pi_{r}(i,:)\|_{F}\geq \|\Pi_{r}(i,:)\|_{1}/\sqrt{K}=1/\sqrt{K}$ for $1\leq i\leq n_{r}$. Meanwhile,
\begin{align*}
\mathrm{max}_{i}\|e'_{i}U\|^{2}_{F}&=\mathrm{max}_{i}\|\Pi_{r}(i,:)\|^{2}_{F}\frac{\Pi_{r}(i,:)}{\|\Pi_{r}(i,:)\|_{F}}B_{r}B'_{r}\frac{\Pi'_{r}(i,:)}{\|\Pi_{r}(i,:)\|_{F}}\leq \mathrm{max}_{i}\|\Pi_{r}(i,:)\|^{2}_{F}\mathrm{max}_{\|x\|_{F}=1}x'B_{r}B'_{r}x\\
&=\mathrm{max}_{i}\|\Pi_{r}(i,:)\|^{2}_{F}\lambda_{1}(B_{r}B'_{r})\overset{\mathrm{By~Lemma~}\ref{P3}}{=}\frac{\mathrm{max}_{i}\|\Pi_{r}(i,:)\|^{2}_{F}}{\lambda_{K}(\Pi'_{r}\Pi_{r})}\leq \frac{1}{\lambda_{K}(\Pi'_{r}\Pi_{r})}.
\end{align*}
This lemma holds by following a similar proof for $\|V(j,:)\|_{F}$.
\end{proof}
\begin{lem}\label{P3}
Under $DiMMSB(n_{r}, n_{c}, K, P,\Pi_{r}, \Pi_{c})$, we have
\begin{align*}	
\lambda_{1}(B_{r}B'_{r})=\frac{1}{\lambda_{K}(\Pi'_{r}\Pi_{r})},\lambda_{K}(B_{r}B'_{r})=\frac{1}{\lambda_{1}(\Pi'_{r}\Pi_{r})},\mathrm{and~}\lambda_{1}(B_{c}B'_{c})=\frac{1}{\lambda_{K}(\Pi'_{c}\Pi_{c})},\lambda_{K}(B_{c}B'_{c})=\frac{1}{\lambda_{1}(\Pi'_{c}\Pi_{c})}.
\end{align*}
\end{lem}
\begin{proof}
Recall that $U=\Pi_{r}B_{r}$ and $U'U=I$, we have $I=B'_{r}\Pi'_{r}\Pi_{r}B_{r}$. As $B_{r}$ is full rank, we have $\Pi'_{r}\Pi_{r}=(B_{r}B'_{r})^{-1}$, which gives
\begin{align*}
\lambda_{1}(B_{r}B'_{r})=\frac{1}{\lambda_{K}(\Pi'_{r}\Pi_{r})},\lambda_{K}(B_{r}B'_{r})=\frac{1}{\lambda_{1}(\Pi'_{r}\Pi_{r})}.
\end{align*}
Follow similar proof for $B_{c}B'_{c}$, this lemma follows.
\end{proof}
\begin{lem}\label{P4}
Under $DiMMSB(n_{r}, n_{c}, K, P,\Pi_{r}, \Pi_{c})$, we have
\begin{align*} \sigma_{K}(\Omega)\geq\rho\sigma_{K}(\tilde{P})\sigma_{K}(\Pi_{r})\sigma_{K}(\Pi_{c})\mathrm{~and~} \sigma_{1}(\Omega)\leq\rho\sigma_{1}(\tilde{P})\sigma_{1}(\Pi_{r})\sigma_{1}(\Pi_{c}).
\end{align*}
\end{lem}
\begin{proof}
For $\sigma_{K}(\Omega)$, we have
\begin{align*}
\sigma^{2}_{K}(\Omega)=\lambda_{K}(\Omega\Omega')&=\lambda_{K}(\Pi_{r}P\Pi'_{c}\Pi_{c}P'\Pi'_{r})=\lambda_{K}(\Pi'_{r}\Pi_{r}P\Pi'_{c}\Pi_{c}P')\\
&\geq \lambda_{K}(\Pi'_{r}\Pi_{r})\lambda_{K}(P\Pi'_{c}\Pi_{c}P')=\lambda_{K}(\Pi'_{r}\Pi_{r})\lambda_{K}(\Pi'_{c}\Pi_{c}P'P)\\
&\geq\lambda_{K}(\Pi'_{r}\Pi_{r})\lambda_{K}(\Pi'_{c}\Pi_{c})\lambda_{K}(PP')=\rho^{2}\sigma^{2}_{K}(\Pi_{r})\sigma^{2}_{K}(\Pi_{c})\sigma^{2}_{K}(\tilde{P}),
\end{align*}
where we have used the fact that for any matrices $X, Y$, the nonzero eigenvalues of $XY$ are the same as the nonzero eigenvalues of $YX$.

For $\sigma_{1}(\Omega)$, since $\Omega=\Pi_{r}P\Pi'_{c}=\rho \Pi_{r}\tilde{P}\Pi'_{c}$, we have
\begin{align*}
\sigma_{1}(\Omega)=\|\Omega\|=\rho\|\Pi_{r}\tilde{P}\Pi'_{c}\|\leq \rho \|\Pi_{r}\|\|\tilde{P}\|\|\Pi_{c}\|=\rho\sigma_{1}(P)\sigma_{1}(\Pi_{r})\sigma_{1}(\Pi_{c}).
\end{align*}
\end{proof}
\section{Proof of consistency of DiSP}
\subsection{Proof of Lemma \ref{BoundAOmega}}
\begin{proof}
We use the rectangular version of the Bernstein inequality in \cite{tropp2012user} to bound $\|A-\Omega\|$. First, we write the rectangular version of the Bernstein inequality as follows:
\begin{thm}\label{Bernstein}
Consider a sequence $\{X_{k}\}$ of $d_{1}\times d_{1}$ random matrices that satisfy the assumptions
\begin{align*}
\mathbb{E}(X_{k})=0~~~\mathrm{and}~~~\|X_{k}\|\leq R~~~\mathrm{almost~surely},
\end{align*}
then
\begin{align*}
\mathbb{P}(\|\sum_{k}X_{k}\|\geq t)\leq(d_{1}+d_{2})\cdot\mathrm{exp}(\frac{-t^{2}/2}{\sigma^{2}+Rt/3}),
\end{align*}
where the variance parameter
\begin{align*}
\sigma^{2}:=\mathrm{max}(\|\sum_{k}\mathbb{E}(X_{k}X'_{k})\|, \|\sum_{k}\mathbb{E}(X'_{k}X_{k})\|).
\end{align*}
\end{thm}

Let $e_{i}$ be an $n_{r}\times 1$ vector, where $e_{i}(i)=1$ and 0 elsewhere, for row nodes $1\leq i\leq n_{r}$, and $\tilde{e}_{j}$ be an $n_{c}\times 1$ vector, where $\tilde{e}_{j}(j)=1$ and 0 elsewhere, for column nodes $1\leq j\leq n_{c}$. Then we can write $W$ as $W=\sum_{i=1}^{n^{r}}\sum_{j=1}^{n_{c}}W(i,j)e_{i}\tilde{e}'_{j}$, where $W=A-\Omega$. Set $W^{(i,j)}$ as the $n_{r}\times n_{c}$ matrix such that $W^{(i,j)}=W(i,j)e_{i}\tilde{e}'_{j}$, for $1\leq i\leq n_{r}, 1\leq j\leq n_{c}$. Surely, we have $\mathbb{E}(W^{(i,j)})=0$. By the definition of the matrix spectral norm, for $1\leq i\leq n_{r}, 1\leq j\leq n_{c}$, we have
\begin{align*}
\|W^{(i,j)}\|&=\|W(i,j)e_{i}\tilde{e}'_{j}\|=|W(i,j)|\|e_{i}\tilde{e}'_{j}\|=|W(i,j)|=|A(i,j)-\Omega(i,j)|\leq 1.
\end{align*}
Next, we consider the variance parameter
\begin{align*}
\sigma^{2}:=\mathrm{max}(\|\sum_{i=1}^{n_{r}}\sum_{j=1}^{n_{c}}\mathbb{E}(W^{(i,j)}(W^{(i,j)})')\|,\|\sum_{i=1}^{n_{r}}\sum_{j=1}^{n_{c}}\mathbb{E}((W^{(i,j)})'W^{(i,j)})\|).
\end{align*}
Since $\Omega(i,j)=\mathbb{E}(A(i,j))$, we can obtain the bound of $\mathbb{E}(W^{2}(i,j))$ first. We have
\begin{align*}
\mathbb{E}(W^{2}(i,j))&=\mathbb{E}((A(i,j)-\Omega(i,j))^{2})=\mathbb{E}((A(i,j)-\mathbb{E}(A(i,j)))^{2})=\mathrm{Var}(A(i,j)),
\end{align*}
where $\mathrm{Var}(A(i,j))$ denotes the variance of Bernoulli random variable $A(i,j)$. Then we have
\begin{align*}
\mathbb{E}(W^{2}(i,j))=\mathrm{Var}(A(i,j))=\mathbb{P}(A(i,j)=1)(1-\mathbb{P}(A(i,j)=1))\leq \mathbb{P}(A(i,j)=1)=\Omega(i,j)=e'_{i}\Pi_{r}\rho\tilde{P}\Pi'_{c}\tilde{e}_{j}\leq \rho.
\end{align*}
Since $e_{i}e'_{i}$ is an $n_{r}\times n_{r}$ diagonal matrix with $(i,i)$-th entry being 1 and others entries being 0, then we bound $\|\sum_{i=1}^{n_{r}}\sum_{j=1}^{n_{c}}\mathbb{E}(W^{(i,j)}(W^{(i,j)})')\|$ as
\begin{align*}
\|\sum_{i=1}^{n_{r}}\sum_{j=1}^{n_{c}}\mathbb{E}(W^{(i,j)}(W^{(i,j)})')\|&=\|\sum_{i=1}^{n_{r}}\sum_{j=1}^{n_{c}}\mathbb{E}(W^{2}(i,j))e_{i}\tilde{e}'_{j}\tilde{e}_{j}e'_{i}\|=\|\sum_{i=1}^{n_{r}}\sum_{j=1}^{n_{c}}\mathbb{E}(W^{2}(i,j))e_{i}e'_{i}\|\\
&=\underset{1\leq i\leq n_{r}}{\mathrm{max}}|\sum_{j=1}^{n_{c}}\mathbb{E}(W^{2}(i,j))|\leq \rho n_{c}.
\end{align*}
Similarly, we have $\|\sum_{i=1}^{n_{r}}\sum_{j=1}^{n_{c}}\mathbb{E}((W^{(i,j)})'W^{(i,j)})\|\leq\rho n_{r}$. Thus, we have
\begin{align*}
\sigma^{2}=\mathrm{max}(\|\sum_{i=1}^{n_{r}}\sum_{j=1}^{n_{c}}\mathbb{E}(W^{(i,j)}(W^{(i,j)})')\|,\|\sum_{i=1}^{n_{r}}\sum_{j=1}^{n_{c}}\mathbb{E}((W^{(i,j)})'W^{(i,j)})\|)\leq\rho\mathrm{max}(n_{r}, n_{c}).
\end{align*}
By  the rectangular version of Bernstein inequality, combining with $\sigma^{2}\leq \rho\mathrm{max}(n_{r},n_{c}), R=1, d_{1}+d_{2}=n_{r}+n_{c}$, set
$t=\frac{\alpha+1+\sqrt{\alpha^{2}+20\alpha+19}}{3}\sqrt{\rho\mathrm{max}(n_{r},n_{c})\mathrm{log}(n_{r}+n_{c})}$, we have
\begin{align*}
\mathbb{P}(\|W\|\geq t)&=\mathbb{P}(\|\sum_{i=1}^{n^{r}}\sum_{j=1}^{n_{c}}W^{(i,j)}\|\geq t)=(n_{r}+n_{c})\mathrm{exp}(-\frac{t^{2}/2}{\sigma^{2}+\frac{Rt}{3}})\leq (n_{r}+n_{c})\mathrm{exp}(-\frac{t^{2}/2}{\rho\mathrm{max}(n_{r}, n_{c})+t/3})\\
&=(n_{r}+n_{c})\mathrm{exp}(-(\alpha+1)\mathrm{log}(n_{r}+n_{c})\cdot \frac{1}{\frac{2(\alpha+1)\rho\mathrm{max}(n_{r}, n_{c})\mathrm{log}(n_{r}+n_{c})}{t^{2}}+\frac{2(\alpha+1)}{3}\frac{\mathrm{log}(n_{r}+n_{c})}{t}})\\
&=(n_{r}+n_{c})\mathrm{exp}(-(\alpha+1)\mathrm{log}(n_{r}+n_{c})\cdot \frac{1}{\frac{18}{(\sqrt{\alpha+19}+\sqrt{\alpha+1})^{2}}+\frac{2\sqrt{\alpha+1}}{\sqrt{\alpha+19}+\sqrt{\alpha+1}}\sqrt{\frac{\mathrm{log}(n_{r}+n_{c})}{\rho\mathrm{max}(n_{r}, n_{c})}}})\\
&\leq (n_{r}+n_{c})\mathrm{exp}(-(\alpha+1)\mathrm{log}(n_{r}+n_{c}))=\frac{1}{(n_{r}+n_{c})^{\alpha}},
\end{align*}
where we have used the assumption (\ref{a1}) and the fact that $\frac{18}{(\sqrt{\alpha+19}+\sqrt{\alpha+1})^{2}}+\frac{2\sqrt{\alpha+1}}{\sqrt{\alpha+19}+\sqrt{\alpha+1}}\sqrt{\frac{\mathrm{log}(n_{r}+n_{c})}{\rho\mathrm{max}(n_{r}, n_{c})}}\leq \frac{18}{(\sqrt{\alpha+19}+\sqrt{\alpha+1})^{2}}+\frac{2\sqrt{\alpha+1}}{\sqrt{\alpha+19}+\sqrt{\alpha+1}}=1$ in the last inequality. Thus, the claim follows.
\end{proof}
\subsection{Proof of Lemma \ref{rowwiseerror}}
\begin{proof}
We use Theorem 4.3.1 \cite{chen2021spectral} to bound $\|\hat{U}\mathrm{sgn}(H_{\hat{U}})-U\|_{2\rightarrow\infty}$ and  $\|\hat{V}\mathrm{sgn}(H_{\hat{V}})-V\|_{2\rightarrow\infty}$ where $\mathrm{sgn}(H_{\hat{U}})$ and $\mathrm{sgn}(H_{\hat{V}})$ are defined later. Let $H_{\hat{U}}=\hat{U}'U$, and $H_{\hat{U}}=U_{H_{\hat{U}}}\Sigma_{H_{\hat{U}}}V'_{H_{\hat{U}}}$ be the SVD decomposition of $H_{\hat{U}}$ with $U_{H_{\hat{U}}},V_{H_{\hat{U}}}\in \mathbb{R}^{n_{r}\times K}$, where $U_{H_{\hat{U}}}$ and $V_{H_{\hat{U}}}$ represent respectively the left and right singular matrices of $H_{\hat{U}}$. Define $\mathrm{sgn}(H_{\hat{U}})=U_{H_{\hat{U}}}V'_{H_{\hat{U}}}$. $\mathrm{sgn}(H_{\hat{V}})$ is defined similarly. Since $\mathbb{E}(A(i,j)-\Omega(i,j))=0$, $\mathbb{E}[(A(i,j)-\Omega(i,j))^{2}]\leq \rho$ by the proof of Lemma \ref{BoundAOmega}, $\frac{1}{\sqrt{\rho \mathrm{min}(n_{r},n_{c})/(\mu \mathrm{log}(n_{r}+n_{c}))}}\leq O(1)$ holds by assumption (\ref{a1}). Then by Theorem 4.3.1. \cite{chen2021spectral}, with high probability,
\begin{align*}
\mathrm{max}(\|\hat{U}\mathrm{sgn}(H_{\hat{U}})-U\|_{2\rightarrow\infty},\|\hat{V}\mathrm{sgn}(H_{\hat{V}})-V\|_{2\rightarrow\infty})&\leq C\frac{\sqrt{\rho K}(\kappa(\Omega)\sqrt{\frac{\mathrm{max}(n_{r},n_{c})\mu}{\mathrm{min}(n_{r},n_{c})}}+\sqrt{\mathrm{log}(n_{r}+n_{c})})}{\sigma_{K}(\Omega)},
\end{align*}
provided that $c_{1}\sigma_{K}(\Omega)\geq \sqrt{\rho(n_{r}+n_{c})\mathrm{log}(n_{r}+n_{c})}$ for some sufficiently small constant $c_{1}$.

Now we are ready to bound $\|\hat{U}\hat{U}'-UU'\|_{2\rightarrow\infty}$ and $\|\hat{V}\hat{V}'-VV'\|_{2\rightarrow\infty}$. Since $\hat{U}'\hat{U}=I$, by basic algebra, we have
\begin{align*}	
\|\hat{U}\hat{U}'-UU'\|_{2\rightarrow\infty}&\leq2\|U-\hat{U}\mathrm{sgn}(H_{\hat{U}})\|_{2\rightarrow\infty}\leq C\frac{\sqrt{K}\sqrt{\rho}(\kappa(\Omega)\sqrt{\frac{\mathrm{max}(n_{r},n_{c})\mu}{\mathrm{min}(n_{r},n_{c})}}+\sqrt{\mathrm{log}(n_{r}+n_{c})})}{\sigma_{K}(\Omega)}\\
&\overset{\mathrm{By~Lemma~}\ref{P4}}{\leq}C\frac{\sqrt{K}(\kappa(\Omega)\sqrt{\frac{\mathrm{max}(n_{r},n_{c})\mu}{\mathrm{min}(n_{r},n_{c})}}+\sqrt{\mathrm{log}(n_{r}+n_{c})})}{\sigma_{K}(\tilde{P})\sigma_{K}(\Pi_{r})\sigma_{K}(\Pi_{c})\sqrt{\rho}}.	
\end{align*}
The lemma holds by following similar proof for $\|\hat{V}\hat{V}'-VV'\|_{2\rightarrow\infty}$.
\end{proof}
\subsection{Proof of Lemma \ref{boundC}}
\begin{proof}
First, we write down the SP algorithm as follows.
	\begin{algorithm}
		\caption{\textbf{Successive Projection (SP)} \cite{gillis2015semidefinite}}
		\label{alg:SP}
		\begin{algorithmic}[1]
			\Require Near-separable matrix $Y_{sp}=S_{sp}M_{sp}+Z_{sp}\in\mathbb{R}^{m\times n}_{+}$ , where $S_{sp}, M_{sp}$ should satisfy Assumption 1 \cite{gillis2015semidefinite}, the number $r$ of columns to be extracted.
			\Ensure Set of indices $\mathcal{K}$ such that $Y_{sp}(\mathcal{K},:)\approx S$ (up to permutation)
			\State Let $R=Y_{sp}, \mathcal{K}=\{\}, k=1$.
			\State \textbf{While} $R\neq 0$ and $k\leq r$ \textbf{do}
			\State ~~~~~~~$k_{*}=\mathrm{argmax}_{k}\|R(k,:)\|_{F}$.
			\State ~~~~~~$u_{k}=R(k_{*},:)$.
			\State ~~~~~~$R\leftarrow (I-\frac{u_{k}u'_{k}}{\|u_{k}\|^{2}_{F}})R$.
			\State ~~~~~~$\mathcal{K}=\mathcal{K}\cup \{k_{*}\}$.
			\State ~~~~~~k=k+1.
			\State \textbf{end while}
		\end{algorithmic}
	\end{algorithm}
Based on Algorithm \ref{alg:SP}, the following theorem is Theorem 1.1 in \cite{gillis2015semidefinite}.
\begin{thm}\label{gillis2015siamSP}
Fix $m\geq r$ and $n\geq r$. Consider a matrix $Y_{sp}=S_{sp}M_{sp}+Z_{sp}$, where $S_{sp}\in\mathbb{R}^{m\times r}$ has a full column rank, $M_{sp}\in \mathbb{R}^{r\times n}$ is a nonnegative matrix such that the sum of each column is at most 1, and $Z_{sp}=[Z_{sp,1},\ldots, Z_{sp,n}]\in \mathbb{R}^{m\times n}$. Suppose $M_{sp}$ has a submatrix equal to $I_{r}$. Write $\epsilon\leq \mathrm{max}_{1\leq i\leq n}\|Z_{sp,i}\|_{F}$. Suppose $\epsilon=O(\frac{\sigma_{\mathrm{min}}(S_{sp})}{\sqrt{r}\kappa^{2}(S_{sp})})$, where $\sigma_{\mathrm{min}}(S_{sp})$ and $\kappa(S_{sp})$ are the minimum singular value and condition number of $S_{sp}$, respectively. If we apply the SP algorithm to columns of $Y_{sp}$, then it outputs an index set $\mathcal{K}\subset \{1,2,\ldots, n\}$ such that $|\mathcal{K}|=r$ and $\mathrm{max}_{1\leq k\leq r}\mathrm{min}_{j\in\mathcal{K}}\|S_{sp}(:,k)-Y_{sp}(:,j)\|_{F}=O(\epsilon \kappa^{2}(S_{sp}))$, where $S_{sp}(:,k)$ is the $k$-th column of $S_{sp}$.
\end{thm}
First, we consider row nodes. Let $m=K, r=K, n=n_{r}, Y_{sp}=\hat{U}'_{2}, Z=\hat{U}'_{2}-U'_{2}, S_{sp}=U'_{2}(\mathcal{I}_{r},:),$ and $M_{sp}=\Pi_{r}'$. By condition (I2), $M_{sp}$ has an identity submatrix $I_{K}$. By Lemma \ref{rowwiseerror}, we have
\begin{align*}
\epsilon=\mathrm{max}_{1\leq i\leq n_{r}}\|\hat{U}_{2}(i,:)-U_{2}(i,:)\|_{F}=\|\hat{U}_{2}(i,:)-U_{2}(i,:)\|_{2\rightarrow\infty}\leq \varpi.
\end{align*}
By Theorem \ref{gillis2015siamSP}, there exists a permutation matrix $\mathcal{P}_{r}$ such that
\begin{align*} \mathrm{max}_{1\leq k\leq K}\|e'_{k}(\hat{U}_{2}(\mathcal{\hat{I}}_{r},:)-\mathcal{P}'_{r}U_{2}(\mathcal{I}_{r},:))\|_{F}=O(\epsilon\kappa^{2}(U_{2}(\mathcal{I}_{r},:))\sqrt{K})=O(\varpi\kappa^{2}(U_{2}(\mathcal{I}_{r},:))).
\end{align*}
Since $\kappa^{2}(U_{2}(\mathcal{I}_{r},:))=\kappa(U_{2}(\mathcal{I}_{r},:)U'_{2}(\mathcal{I}_{r},:))=\kappa(U(\mathcal{I}_{r},:)U'(\mathcal{I}_{r},:))=\kappa(\Pi'_{r}\Pi_{r})$ where the last equality holds by Lemma \ref{P3}, we have
\begin{align*}
\mathrm{max}_{1\leq k\leq K}\|e'_{k}(\hat{U}_{2}(\mathcal{\hat{I}}_{r},:)-\mathcal{P}'_{r}U_{2}(\mathcal{I}_{r},:))\|_{F}=O(\varpi\kappa(\Pi'_{r}\Pi_{r})).
\end{align*}
Follow similar analysis for column nodes, we have
\begin{align*} \mathrm{max}_{1\leq k\leq K}\|e'_{k}(\hat{V}_{2}(\mathcal{\hat{I}}_{c},:)-\mathcal{P}'_{c}V_{2}(\mathcal{I}_{c},:))\|_{F}=O(\varpi\kappa(\Pi'_{c}\Pi_{c})).
\end{align*}
\begin{rem}\label{inputUinSP}
For the ideal case, let $m=K, r=K, n=n_{c}, Y_{sp}=U', Z_{sp}=U'-U'\equiv0, S_{sp}=U'(\mathcal{I}_{r},:),$ and $M_{sp}=\Pi_{r}'$. Then, we have $\mathrm{max}_{1\leq i\leq n_{r}}\|U(i,:)-U(i,:)\|_{F}=0$. By Theorem \ref{gillis2015siamSP}, the SP algorithm returns $\mathcal{I}_{r}$ when the input is $U$, assuming there are $K$ row communities.
\end{rem}
\end{proof}
\subsection{Proof of Lemma \ref{boundY}}
\begin{proof}
First, we consider row nodes. Recall that $U(\mathcal{I}_{r},:)=B_{r}$. For convenience, set $\hat{U}(\mathcal{\hat{I}}_{r},:)=\hat{B}_{r}, U_{2}(\mathcal{I}_{r},:)=B_{2r}, \hat{U}_{2}(\mathcal{\hat{I}}_{r},:)=\hat{B}_{2r}$. We bound $\|e'_{i}(\hat{Y}_{r}-Y_{r}\mathcal{P}_{r})\|_{F}$ when the input is $\hat{U}$ in the SP algorithm.  Recall that $Y_{r}=\mathrm{max}(UU'(\mathcal{I}_{r},:)(U(\mathcal{I}_{r},:)U'(\mathcal{I}_{r},:))^{-1},0)\equiv \Pi_{r}$,  for $1\leq i\leq n_{r}$, we have
\begin{align*} &\|e'_{i}(\hat{Y}_{r}-Y_{r}\mathcal{P}_{r})\|_{F}=\|e'_{i}(\mathrm{max}(0,\hat{U}\hat{B}'_{r}(\hat{B}_{r}\hat{B}'_{r})^{-1})-UB'_{r}(B_{r}B'_{r})^{-1}\mathcal{P}_{r})\|_{F}\\
&\leq\|e'_{i}(\hat{U}\hat{B}'_{r}(\hat{B}_{r}\hat{B}'_{r})^{-1}-UB'_{r}(B_{r}B'_{r})^{-1}\mathcal{P}_{r})\|_{F}\\ &=\|e'_{i}(\hat{U}-U(U'\hat{U}))\hat{B}'_{r}(\hat{B}_{r}\hat{B}'_{r})^{-1}+e'_{i}(U(U'\hat{U})\hat{B}'_{r}(\hat{B}_{r}\hat{B}'_{r})^{-1}-U(U'\hat{U})(\mathcal{P}'_{r}(B_{r}B'_{r})(B'_{r})^{-1}(U'\hat{U}))^{-1})\|_{F}\\ &\leq\|e'_{i}(\hat{U}-U(U'\hat{U}))\hat{B}'_{r}(\hat{B}_{r}\hat{B}'_{r})^{-1}\|_{F}+\|e'_{i}U(U'\hat{U})(\hat{B}'_{r}(\hat{B}_{r}\hat{B}'_{r})^{-1}-(\mathcal{P}'_{r}(B_{r}B'_{r})(B'_{r})^{-1}(U'\hat{U}))^{-1})\|_{F}\\
&\leq\|e'_{i}(\hat{U}-U(U'\hat{U}))\|_{F}\|\hat{B}^{-1}_{r}\|_{F}+\|e'_{i}U(U'\hat{U})(\hat{B}'_{r}(\hat{B}_{r}\hat{B}'_{r})^{-1}-(\mathcal{P}'_{r}(U_{r}B'_{r})(B'_{r})^{-1}(U'\hat{U}))^{-1})\|_{F}\\	&\leq \sqrt{K}\|e'_{i}(\hat{U}-U(U'\hat{U}))\|_{F}/\sqrt{\lambda_{K}(\hat{B}_{r}\hat{B}'_{r})}+\|e'_{i}U(U'\hat{U})(\hat{B}^{-1}_{r}-(\mathcal{P}_{r}'B_{r}(U'\hat{U}))^{-1})\|_{F}\\
&=\sqrt{K}\|e'_{i}(\hat{U}\hat{U}'-UU')\hat{U}\|_{F}O(\sqrt{\lambda_{1}(\Pi'_{r}\Pi_{r})})+\|e'_{i}U(U'\hat{U})(\hat{B}^{-1}_{r}-(\mathcal{P}_{r}'B_{r}(U'\hat{U}))^{-1})\|_{F}\\
&\leq \sqrt{K}\|e'_{i}(\hat{U}\hat{U}'-UU')\|_{F}O(\sqrt{\lambda_{1}(\Pi'_{r}\Pi_{r})})+\|e'_{i}U(U'\hat{U})(\hat{B}^{-1}_{r}-(\mathcal{P}'_{r}B_{r}(U'\hat{U}))^{-1})\|_{F}\\
&\leq \sqrt{K}\varpi O(\sqrt{\lambda_{1}(\Pi'_{r}\Pi_{r})})+\|e'_{i}U(U'\hat{U})(\hat{B}^{-1}_{r}-(\mathcal{P}'_{r}B_{r}(U'\hat{U}))^{-1})\|_{F}\\	&=O(\varpi\sqrt{K\lambda_{1}(\Pi'_{r}\Pi_{r})})+\|e'_{i}U(U'\hat{U})(\hat{B}^{-1}_{r}-(\mathcal{P}'_{r}B_{r}(U'\hat{U}))^{-1})\|_{F},
\end{align*}
where we have used similar idea in the proof of Lemma VII.3 in \cite{mao2020estimating} such that apply $O(\frac{1}{\lambda_{K}(B_{r}B'_{r})})$ to estimate $\frac{1}{\lambda_{K}(\hat{B}_{r}\hat{B}'_{r})}$, then by Lemma \ref{P3}, we have $\frac{1}{\lambda_{K}(\hat{B}_{r}\hat{B}'_{r})}=O(\lambda_{1}(\Pi'_{r}\Pi_{r}))$.

Now we aim to bound $\|e'_{i}U(U'\hat{U})(\hat{B}^{-1}_{r}-(\mathcal{P}_{r}'B_{r}(U'\hat{U}))^{-1})\|_{F}$. For convenience, set $T=U'\hat{U}, S=\mathcal{P}_{r}'B_{r}T$. We have
\begin{align}	&\|e'_{i}U(U'\hat{U})(\hat{B}^{-1}_{r}-(\mathcal{P}'_{r}B_{r}(U'\hat{U}))^{-1})\|_{F}=\|e'_{i}UTS^{-1}(S-\hat{B}_{r})\hat{B}^{-1}_{r}\|_{F}\notag\\	&\leq\|e'_{i}UTS^{-1}(S-\hat{B}_{r})\|_{F}\|\hat{B}^{-1}_{r}\|_{F}\leq\|e'_{i}UTS^{-1}(S-\hat{B}_{r})\|_{F}\frac{\sqrt{K}}{|\lambda_{K}(\hat{B}_{r})|}\notag\\	&=\|e'_{i}UTS^{-1}(S-\hat{B}_{r})\|_{F}\frac{\sqrt{K}}{\sqrt{\lambda_{K}(\hat{B}_{r}\hat{B}'_{r})}}\leq\|e'_{i}UTS^{-1}(S-\hat{B}_{r})\|_{F}O(\sqrt{K\lambda_{1}(\Pi'_{r}\Pi_{r})})\notag\\	&=\|e'_{i}UTT^{-1}B'_{r}(B_{r}B'_{r})^{-1}\mathcal{P}_{r}(S-\hat{B}_{r})\|_{F}O(\sqrt{K\lambda_{1}(\Pi'_{r}\Pi_{r})})\notag\\	&=\|e'_{i}UB'_{r}(B_{r}B'_{r})^{-1}\mathcal{P}_{r}(S-\hat{B}_{r})\|_{F}O(\sqrt{K\lambda_{1}(\Pi'_{r}\Pi_{r})})\notag\\ &=\|e'_{i}Y_{r}\mathcal{P}_{r}(S-\hat{B}_{r})\|_{F}O(\sqrt{K\lambda_{1}(\Pi'_{r}\Pi_{r})})\overset{\mathrm{By~}Y_{r}=\Pi_{r}}{\leq}\mathrm{max}_{1\leq k\leq K}\|e'_{k}(S-\hat{B}_{r})\|_{F}O(\sqrt{K\lambda_{1}(\Pi'_{r}\Pi_{r})})\notag\\ &=\mathrm{max}_{1\leq k\leq K}\|e'_{k}(\hat{B}_{r}-\mathcal{P}_{r}'B_{r}U'\hat{U})\|_{F}O(\sqrt{K\lambda_{1}(\Pi'_{r}\Pi_{r})})\notag\\
&=\mathrm{max}_{1\leq k\leq K}\|e'_{k}(\hat{B}_{r}\hat{U}'-\mathcal{P}'_{r}B_{r}U')\hat{U}\|_{F}O(\sqrt{K\lambda_{1}(\Pi'_{r}\Pi_{r})})\notag\\ &\leq\mathrm{max}_{1\leq k\leq K}\|e'_{k}(\hat{B}_{r}\hat{U}'-\mathcal{P}_{r}'B_{r}U')\|_{F}O(\sqrt{K\lambda_{1}(\Pi'_{r}\Pi_{r})})\notag\\
&=\mathrm{max}_{1\leq k\leq K}\|e'_{k}(\hat{B}_{2r}-\mathcal{P}_{r}'B_{2r})\|_{F}O(\sqrt{K\lambda_{1}(\Pi'_{r}\Pi_{r})})\label{Benefit}\\
&=O(\varpi\kappa(\Pi'_{r}\Pi_{r})\sqrt{K\lambda_{1}(\Pi'_{r}\Pi_{r})})\notag.
\end{align}
\begin{rem}\label{BenefitEquivalence}
Eq (\ref{Benefit}) supports our statement that building the theoretical framework of DiSP benefits a lot by introducing DiSP-equivalence algorithm since $\|\hat{B}_{2r}-\mathcal{P}'_{r}B_{2r}\|_{2\rightarrow\infty}$ is obtained from DiSP-equivalence (i.e., inputing $\hat{U}_{2}$ in the SP algorithm obtains $\|\hat{B}_{2r}-\mathcal{P}'_{r}B_{2r}\|_{2\rightarrow\infty}$. Similar benefits hold for column nodes.).
\end{rem}
Then, we have
\begin{align*}
\|e'_{i}(\hat{Y}_{r}-Y_{r}\mathcal{P}_{r})\|_{F}&\leq O(\varpi\sqrt{K\lambda_{1}(\Pi'_{r}\Pi_{r})})+\|e'_{i}U(U'\hat{U})(\hat{B}^{-1}_{r}-(\mathcal{P}'_{r}B_{r}(U'\hat{U}))^{-1})\|_{F}\\
&\leq O(\varpi\sqrt{K\lambda_{1}(\Pi'_{r}\Pi_{r})})+O(\varpi\kappa(\Pi'_{r}\Pi_{r})\sqrt{K\lambda_{1}(\Pi'_{r}\Pi_{r})})\\
&=O(\varpi\kappa(\Pi'_{r}\Pi_{r})\sqrt{K\lambda_{1}(\Pi'_{r}\Pi_{r})}).
\end{align*}
Follow similar proof for column nodes, we have, for $1\leq j\leq n_{c}$,
\begin{align*}	\|e'_{j}(\hat{Y}_{c}-Y_{c}\mathcal{P}_{c})\|_{F}=O(\varpi\kappa(\Pi'_{c}\Pi_{c})\sqrt{K\lambda_{1}(\Pi'_{c}\Pi_{c})}).
\end{align*}
\end{proof}
\subsection{Proof of Theorem \ref{Main}}
\begin{proof}
Since
\begin{align*}	\|e'_{i}(\hat{\Pi}_{r}-\Pi_{r}\mathcal{P}_{r})\|_{1}&=\|\frac{e'_{i}\hat{Y}_{r}}{\|e'_{i}\hat{Y}_{r}\|_{1}}-\frac{e'_{i}Y_{r}\mathcal{P}_{r}}{\|e'_{i}Y_{r}\mathcal{P}_{r}\|_{1}}\|_{1}=\|\frac{e'_{i}\hat{Y}_{r}\|e'_{i}Y_{r}\|_{1}-e'_{i}Y_{r}\mathcal{P}_{r}\|e'_{i}\hat{Y}_{r}\|_{1}}{\|e'_{i}\hat{Y}_{r}\|_{1}\|e'_{i}Y_{r}\|_{1}}\|_{1}\\	&=\|\frac{e'_{i}\hat{Y}_{r}\|e'_{i}Y_{r}\|_{1}-e'_{i}\hat{Y}_{r}\|e'_{i}\hat{Y}_{r}\|_{1}+e'_{i}\hat{Y}_{r}\|e'_{i}\hat{Y}_{r}\|_{1}-e'_{i}Y_{r}\mathcal{P}\|e'_{i}\hat{Y}_{r}\|_{1}}{\|e'_{i}\hat{Y}_{r}\|_{1}\|e'_{i}Y_{r}\|_{1}}\|_{1}\\
&\leq\frac{\|e'_{i}\hat{Y}_{r}\|e'_{i}Y_{r}\|_{1}-e'_{i}\hat{Y}_{r}\|e'_{i}\hat{Y}_{r}\|_{1}\|_{1}+\|e'_{i}\hat{Y}_{r}\|e'_{i}\hat{Y}_{r}\|_{1}-e'_{i}Y_{r}\mathcal{P}_{r}\|e'_{i}\hat{Y}_{r}\|_{1}\|_{1}}{\|e'_{i}\hat{Y}_{r}\|_{1}\|e'_{i}Y_{r}\|_{1}}\\	&=\frac{\|e'_{i}\hat{Y}_{r}\|_{1}|\|e'_{i}Y_{r}\|_{1}-\|e'_{i}\hat{Y}_{r}\|_{1}|+\|e'_{i}\hat{Y}_{r}\|_{1}\|e'_{i}\hat{Y}_{r}-e'_{i}Y_{r}\mathcal{P}_{r}\|_{1}}{\|e'_{i}\hat{Y}_{r}\|_{1}\|e'_{i}Y_{r}\|_{1}}\\ &=\frac{|\|e'_{i}Y_{r}\|_{1}-\|e'_{i}\hat{Y}_{r}\|_{1}|+\|e'_{i}\hat{Y}_{r}-e'_{i}Y_{r}\mathcal{P}_{r}\|_{1}}{\|e'_{i}Y_{r}\|_{1}}\leq\frac{2\|e'_{i}(\hat{Y}_{r}-Y_{r}\mathcal{P}_{r})\|_{1}}{\|e'_{i}Y_{r}\|_{1}}\\
&=\frac{2\|e'_{i}(\hat{Y}_{r}-Y_{r}\mathcal{P}_{r})\|_{1}}{\|e'_{i}\Pi_{r}\|_{1}}=2\|e'_{i}(\hat{Y}_{r}-Y_{r}\mathcal{P}_{r})\|_{1}\leq 2\sqrt{K}\|e'_{i}(\hat{Y}_{r}-Y_{r}\mathcal{P}_{r})\|_{F},
\end{align*}
we have
\begin{align*}	\|e'_{i}(\hat{\Pi}_{r}-\Pi_{r}\mathcal{P}_{r})\|_{1}=O(\varpi\kappa(\Pi'_{r}\Pi_{r})K\sqrt{\lambda_{1}(\Pi'_{r}\Pi_{r})}).
\end{align*}
Follow similar proof for column nodes, we have, for $1\leq j\leq n_{c}$,
\begin{align*}	\|e'_{j}(\hat{\Pi}_{c}-\Pi_{c}\mathcal{P}_{c})\|_{1}=O(\varpi\kappa(\Pi'_{c}\Pi_{c})K\sqrt{\lambda_{1}(\Pi'_{c}\Pi_{c})}).
\end{align*}
\end{proof}
\subsection{Proof of Corollary \ref{AddConditions}}
\begin{proof}
Under  conditions of Corollary \ref{AddConditions}, we have
\begin{align*}
&\|e'_{i}(\hat{\Pi}_{r}-\Pi_{r}\mathcal{P}_{r})\|_{1}=O(\varpi K\sqrt{\frac{n_{r}}{K}})=O(\varpi\sqrt{Kn_{r}}),\\ &\|e'_{j}(\hat{\Pi}_{c}-\Pi_{c}\mathcal{P}_{c})\|_{1}=O(\varpi
K\sqrt{\frac{n_{c}}{K}})=O(\varpi\sqrt{Kn_{c}}).
\end{align*}
Under  conditions of Corollary \ref{AddConditions}, $\kappa(\Omega)=O(1)$ by Lemma \ref{P4} and $\mu=O(1)=C$ by Lemma \ref{P2} for some constant $C>0$. Then, by Lemma \ref{rowwiseerror}, we have
\begin{align*}
\varpi&=O(\frac{\sqrt{K}(\kappa(\Omega)\sqrt{\frac{\mathrm{max}(n_{r},n_{c})\mu}{\mathrm{min}(n_{r},n_{c})}}+\sqrt{\mathrm{log}(n_{r}+n_{c})})}{\sqrt{\rho}\sigma_{K}(\tilde{P})\sigma_{K}(\Pi_{r})\sigma_{K}(\Pi_{c})})=O(\frac{\sqrt{K}(\sqrt{C\frac{\mathrm{max}(n_{r},n_{c})}{\mathrm{min}(n_{r},n_{c})}}+\sqrt{\mathrm{log}(n_{r}+n_{c})})}{\sqrt{\rho}\sigma_{K}(\tilde{P})\sigma_{K}(\Pi_{r})\sigma_{K}(\Pi_{c})})\\
&=O(\frac{\sqrt{K}(\sqrt{C\frac{\mathrm{max}(n_{r},n_{c})}{\mathrm{min}(n_{r},n_{c})}}+\sqrt{\mathrm{log}(n_{r}+n_{c})})}{\sqrt{\rho}\sigma_{K}(\tilde{P})\sqrt{n_{r}n_{c}}/K})=O(\frac{K^{1.5}(\sqrt{C\frac{\mathrm{max}(n_{r},n_{c})}{\mathrm{min}(n_{r},n_{c})}}+\sqrt{\mathrm{log}(n_{r}+n_{c})})}{\sigma_{K}(\tilde{P})\sqrt{\rho n_{r}n_{c}}}),
\end{align*}
which gives that
\begin{align*}
&\|e'_{i}(\hat{\Pi}_{r}-\Pi_{r}\mathcal{P}_{r})\|_{1}=O(\frac{K^{2}(\sqrt{C\frac{\mathrm{max}(n_{r},n_{c})}{\mathrm{min}(n_{r},n_{c})}}+\sqrt{\mathrm{log}(n_{r}+n_{c})})}{\sigma_{K}(\tilde{P})\sqrt{\rho
n_{c}}}),\\
&\|e'_{j}(\hat{\Pi}_{c}-\Pi_{c}\mathcal{P}_{c})\|_{1}=O(\frac{K^{2}(\sqrt{C\frac{\mathrm{max}(n_{r},n_{c})}{\mathrm{min}(n_{r},n_{c})}}+\sqrt{\mathrm{log}(n_{r}+n_{c})})}{\sigma_{K}(\tilde{P})\sqrt{\rho n_{r}}}).
\end{align*}
\end{proof}

\end{document}